\pdfoutput=1

\documentclass[acmtog]{acmart} 

\setcopyright{licensedothergov}\acmJournal{TOG}
\acmYear{2022}\acmVolume{41}\acmNumber{4}\acmArticle{38}\acmMonth{7} \acmDOI{10.1145/3528223.3530180}

\acmSubmissionID{757} 

\usepackage{booktabs} %
\usepackage[ruled]{algorithm2e} %
\usepackage{hyperref, url}

\newcommand{\ie}{\textit{i.e.,}~}
\newcommand{\eg}{\textit{e.g.,}~}

\newcommand{\Dt}{\Delta t}
\newcommand{\R}{\mathbb{R}}

\newcommand{\wx}{\widehat{x}}

\SetAlFnt{\small}
\SetAlCapFnt{\small}
\SetAlCapNameFnt{\small}
\SetAlCapHSkip{0pt}

\setcitestyle{square}

\acmPrice{15}
\setcopyright{rightsretained}
\acmJournal{TOG}
\acmYear{2022}
 
\usepackage{multirow}
\usepackage{booktabs}
\usepackage{bigstrut}
\usepackage{tabu}  
\usepackage{rotating}
\usepackage[export]{adjustbox}

\DeclareMathOperator*{\argmin}{arg\,min}
\newcommand{\red}[1]{\textcolor{black}{#1}}

\begin{document}

\title{High Dynamic Range and Super-Resolution from Raw Image Bursts}
\author{Bruno Lecouat} 
\affiliation{
\institution{Inria and DI/ENS (ENS-PSL, CNRS, Inria)}
\city{Paris}
\country{France}
}
\email{bruno.lecouat@inria.fr}

\author{Thomas Eboli}
\affiliation{
\institution{Université Paris-Saclay, ENS Paris-Saclay, Centre Borelli }
\city{Paris}
\country{France}
}
\email{thomas.eboli@ens-paris-saclay.fr}

\author{Jean Ponce} 
\affiliation{
\institution{Inria and DI/ENS (ENS-PSL, CNRS, Inria), France \& Center for Data Science, New York University}
\city{New York}
\country{USA}
}
\email{jean.ponce@inria.fr}

\author{Julien Mairal} 
\affiliation{
\institution{Univ. Grenoble Alpes, Inria, CNRS, Grenoble INP, LJK}
\city{Grenoble}
\country{France}
}
\email{julien.marial@inria.fr}

\begin{abstract}
    Photographs captured by smartphones and mid-range cameras have limited spatial resolution and dynamic range, with noisy response in underexposed regions and color artefacts in saturated areas. This paper introduces the first approach (to the best of our knowledge) to the reconstruction of high-resolution, high-dynamic range color images from raw photographic bursts captured by a handheld camera with exposure bracketing. This method uses a physically-accurate model of image formation to combine an iterative optimization algorithm for solving the corresponding inverse problem with a learned image representation for robust alignment and a learned natural image prior.  The proposed algorithm is fast, with low memory requirements compared to state-of-the-art learning-based approaches to image restoration, and features that are learned end to end from synthetic yet realistic data. Extensive experiments demonstrate its excellent performance with super-resolution factors of up to $\times 4$ on real photographs taken in the wild with hand-held cameras, and high robustness to low-light conditions, noise, camera shake, and moderate object motion.
\end{abstract}



\begin{teaserfigure}
\centering
\includegraphics[width=\textwidth]{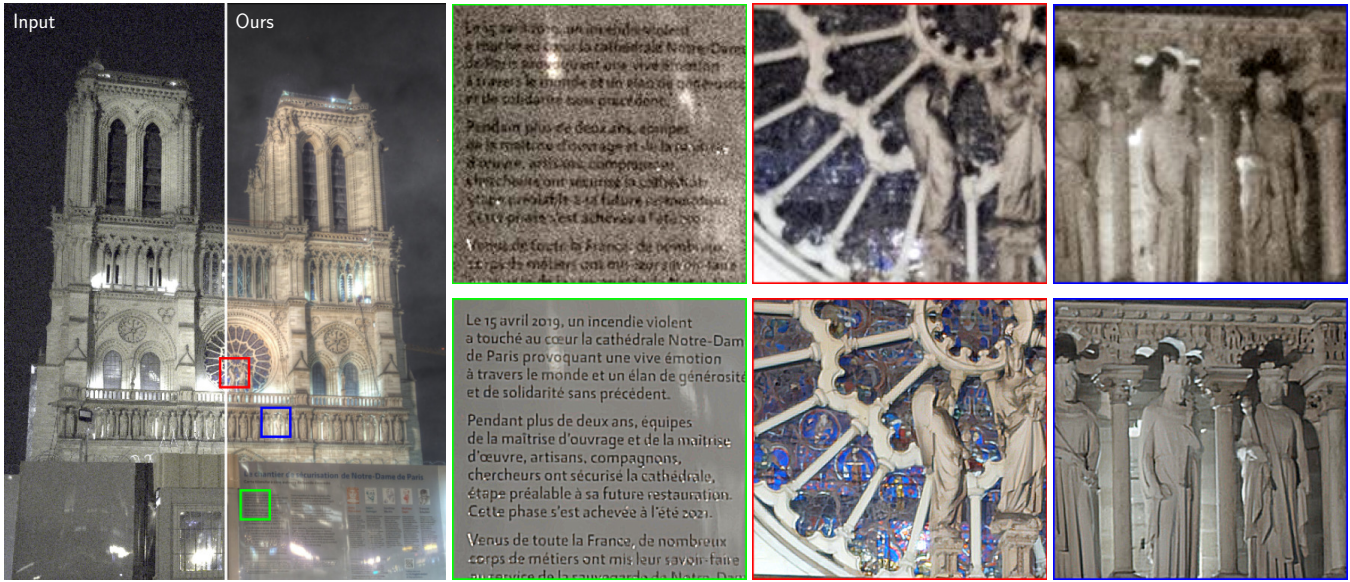}
\caption{An example of joint super-resolution (SR) and high-dynamic range (HDR) imaging. {\bf Left:}  An 18-photo burst was shot at night from a hand-held Pixel 4a smartphone at 12MP resolution with an exposure time varying from $1/340$s to $1/4$s. The left half of the central image from the burst is shown
along with the right half of the 192MP HDR image
reconstructed by our algorithm with a super-resolution factor of $\times 4$ (after tone mapping). {\bf Right:} Three small crops of the two images
corresponding to the colored square regions on the left. Crops from the central image of the burst 
are rendered using Adobe Camera Raw to convert raw files into jpg with
highest quality setting. The HDR/SR results are rendered using the
PhotoMatix tone mapper\url{https://www.hdrsoft.com/}. Note that the 192MP HDR image on the left is not
reproduced at full resolution because of the corresponding file's size.}
\label{fig:teaser}
\end{teaserfigure}

\maketitle

\def\comment#1{{}}
\section{Introduction}
Key factors limiting the level of detail of photographs captured by digital cameras are their spatial resolution and dynamic range: High resolution is necessary to zoom on small image regions, and high dynamic range is needed to reveal details hidden in dark areas (\eg shadows) and avoid color artefacts due to saturation in bright ones (\eg highlights). For a given sensor size, higher resolution also means smaller pixel size, with less light reaching each photoreceptor, resulting in lower dynamic range and increased noise in dark regions, an effect exacerbated in smartphones by their small sensor size. It is natural, and by now rather common, to use multiple photographs to reconstruct an image with higher spatial resolution, a process known as {\em super-resolution} (or {\em SR} for short in this presentation, see, for example~\cite{wronski19handheld}), or dynamic range, a process known as {\em high dynamic range} (or {\em HDR}) imaging (see, for example~\cite{debevec97recovering}).

We propose in this paper a novel method for {\em joint} SR and HDR imaging from the {\em raw} image bursts featuring a range of different exposures that can now be captured by most smartphones and mid-range cameras (Figure~\ref{fig:teaser}). A major challenge tackled by our algorithm is the automated alignment with sub-pixel accuracy of the burst elements required to compensate for camera shake and possibly (moderate) object motion, despite the variations in saturation and signal-to-noise ratio due to the different exposures used across the burst. Other notable difficulties include the high contrasts and noise levels encountered in night scenes for example, where a photo might feature both very dark and noisy regions and saturated ones near light sources, as well as the fact that a digital camera only captures one color channel at each pixel according to the corresponding {\em color filter array} (or {\em CFA}, often a Bayer pattern). Despite the latter challenge, it now seems clear that it is better to work directly with the raw image data than with the sRGB pictures produced by the {\em image signal processor} (or {\em ISP}) of the camera since their construction involves several steps, including white balance, denoising, demosaicking, gamma correction, compression of each color channel content to 8 bits, etc., that result in an unavoidable loss of information in high spatial frequencies and dynamic range.

The approach proposed in the rest of this presentation extends the algorithm for multi-frame super-resolution of~\cite{lecouat21aliasing} to jointly perform blind denoising, demosaicking, super-resolution and HDR image reconstruction from raw bursts. Its key features can be summarized as follows:
\begin{itemize}
\item   Our method uses a physically-accurate model of image formation that
accounts for the successive transformations applied to the original analog irradiance
image, including quantization of the signal, noise, exposure and spatial quantization. \\
\item We combine an iterative optimization algorithm for solving the corresponding inverse problem with a learned image representation for robust alignment and a learned natural image prior. 
This is the first main technical novelty of our paper, enabling us to address the joint reconstruction of high-resolution, high-dynamic range color images from raw photographics bursts captured by a handheld camera with exposure bracketing. \\
\item The proposed algorithm is fast, with low memory requirements compared to state-of-the-art learning-based approaches to image restoration, and features that are learned end to end from synthetic yet realistic data, generated using again our image formation model.  \\
\item We introduce an image alignment method to compensate for camera shake which is robust to (moderate) object motions and an image fusion technique which is itself tolerant to alignment errors. Together, these form the second main technical novelty of our paper, and they are key factors in the robustness of
our algorithm in both the SR and HDR imaging tasks with, notably, significant improvement over~\cite{lecouat21aliasing} in super-resolution.\\
\item Extensive experiments demonstrate the excellent performance of the proposed
approach with super-resolution factors of up to $\times 4$ on real photographs taken in the wild with hand-held cameras, and high robustness to low-light conditions, noise, camera shake, and moderate object motion. These results are confirmed by
quantitative and qualitative comparisons with the state of the art in super-resolution and HDR imaging tasks on synthetic and real image bursts.
\end{itemize}

\section{Background}

\subsection{High dynamic range imaging}

\paragraph{Bracketing techniques.} \cite{mann95undigital,debevec97recovering, granados10optimal,hasinoff10noiseoptimal} construct an HDR image by combining multiple photographs of the same scene with different exposures.  The darkest pictures are used to reconstruct areas prone to saturation and the brightest ones are needed for restoring dark regions that are likely to be noisy (we will come back to that point later).  They typically work on {\em linRGB} images, that is, demosaicked images {\em before}
they are transformed by the camera's ISP into {\em sRGB} images ready for display. 
A sequence of sRGB input photographs must therefore in general be ``linearized''
by inverting this mapping, also known as the camera response function (or {\em CRF}). 
The HDR image is then reconstructed as a weighted sum of the linearized bracket images, normalized by the corresponding shutter speed.  Its pixel values are
typically represent as single-precision floating-point numbers, with min and max  those of the image bracket. Bracketing-based approaches to HDR imaging face a number of classical issues, including choosing the optimal fusion weights, estimating the CRF~\cite{debevec97recovering}, leveraging accurate raw image noise models~\cite{granados10optimal, aguerrebere14best, hanji20noise},
selecting the best exposure parameters for a fixed
number of frames in the bracket~\cite{gallo12metering, hasinoff10noiseoptimal},
registering
images with different exposures~\cite{zimmer11freehand, gallo15nonrigid},
which is significantly more challenging than aligning same-exposure images~\cite{ma17robust}, and removing ghosting artefacts~\cite{sen12robust, tursun16deghosting} due to misalignment.

\paragraph{Using raw bursts with constant shutter speed.}
Unlike classical exposure bracketing techniques, HDR+~\cite{hasinoff16burst} takes as input a burst of raw underexposed images captured with the same exposure time.  These are mostly free of saturation but noisy in dark regions.  A 12-bit, denoised raw image is obtained by aggregating the 10-bit photos of the burst. It is then demosaicked and tone mapped.  
Recent updates of HDR+ use a couple of well-exposed frames to achieve better denoising and
deghosting~\cite{ernst21hdrplus}, or leverage the metering
technique of \citet{hasinoff10noiseoptimal} to adapt
the original algorithm to low-light 
situations~\cite{liba19lowlight}.

\paragraph{Using pixelwise ISO sensitivities.} Instead of relying on classical  imaging devices, \citet{nayar2000high} reconstruct a single HDR image from a sensor with spatially-varying pixel exposures. 
This approach can be further combined with learning-based methods \cite{serrano2016convolutional, martel2020neural}. 
Even though our work focuses on standard sensors, we believe it to be flexible enough to be adapted to pixelwise ISO sensitive sensors under simple modification of the image formation model. This is an interesting research direction for future work, but beyond the scope of our paper.

\begin{figure*}[ht]
    \setlength\tabcolsep{0.5pt}
    \renewcommand{\arraystretch}{0.5}
    \centering 
    \begin{tabular}{ccccc}
        \includegraphics[trim=0 0 0 0,clip,height=0.199\textwidth]{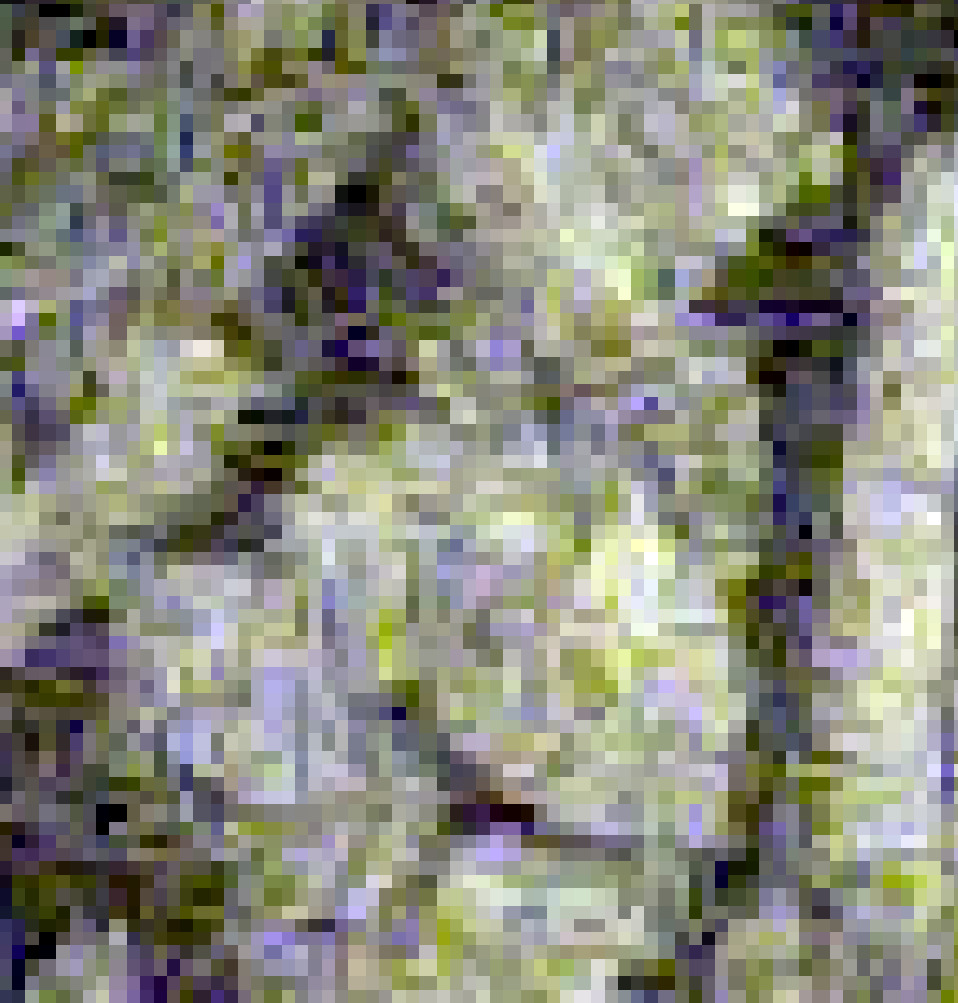}        &  
        \includegraphics[trim=0 0 0 0,clip,height=0.199\textwidth]{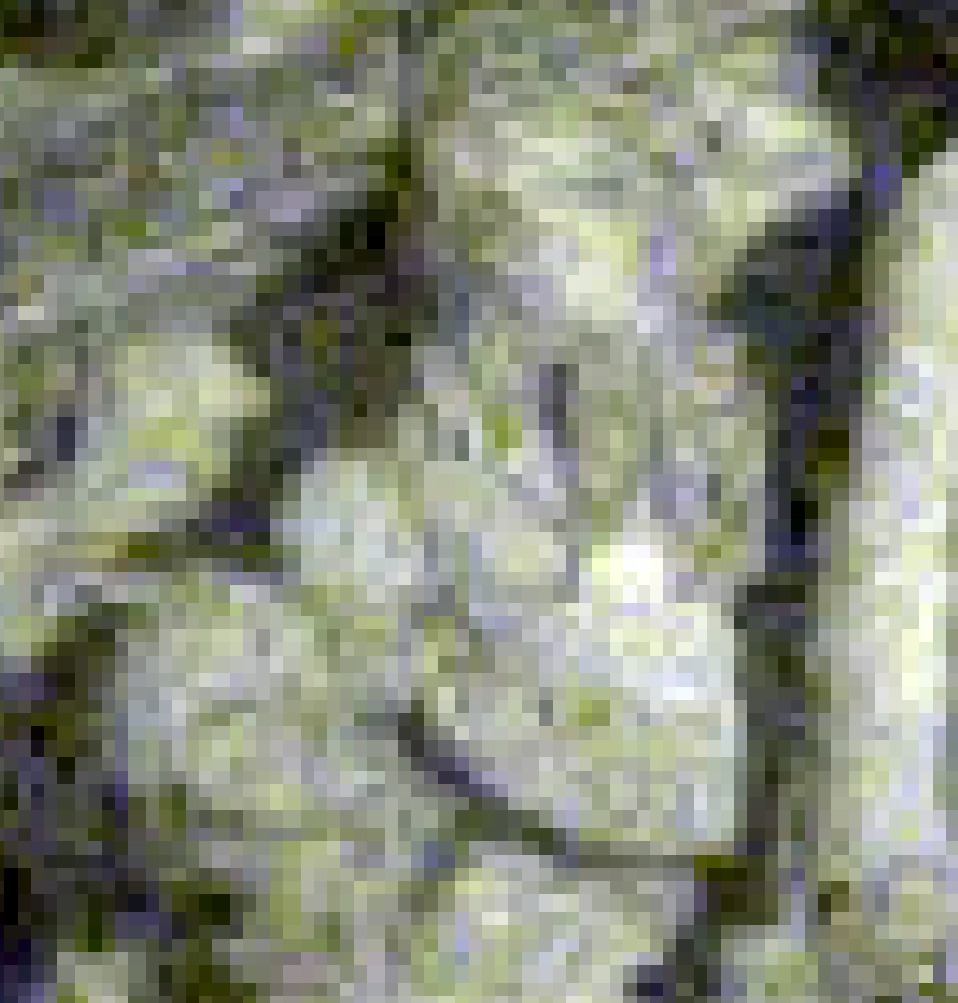}        &  
        \includegraphics[trim=0 0 0 0,clip,height=0.199\textwidth]{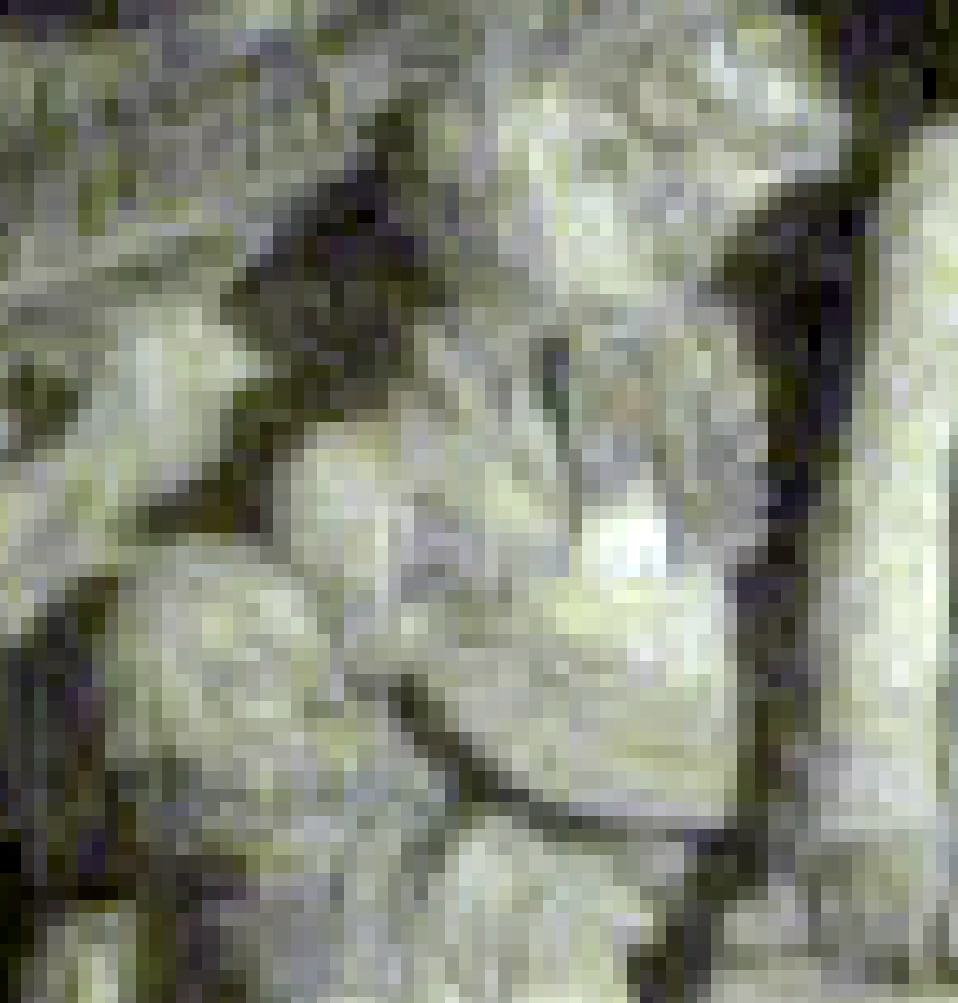}        &  
        \includegraphics[trim=0 0 0 0,clip,height=0.199\textwidth]{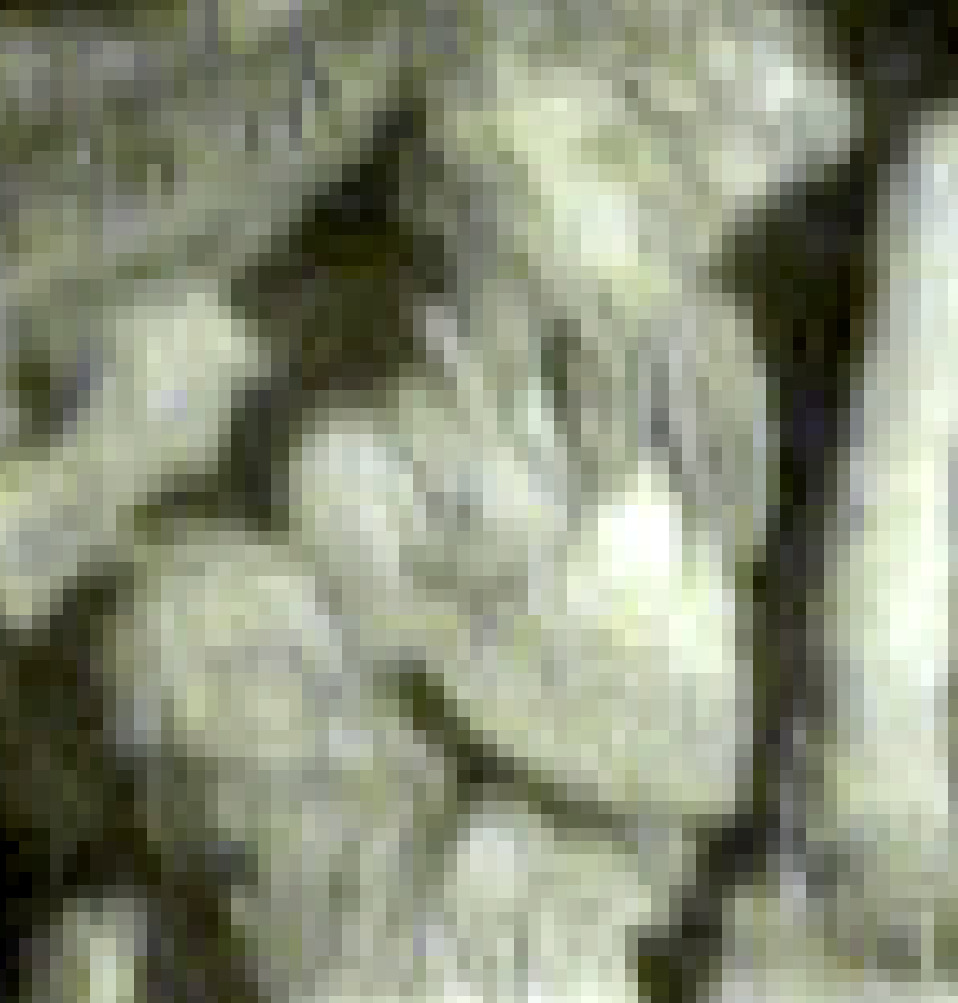}        &  
        \includegraphics[trim=0 0 0 0,clip,height=0.199\textwidth]{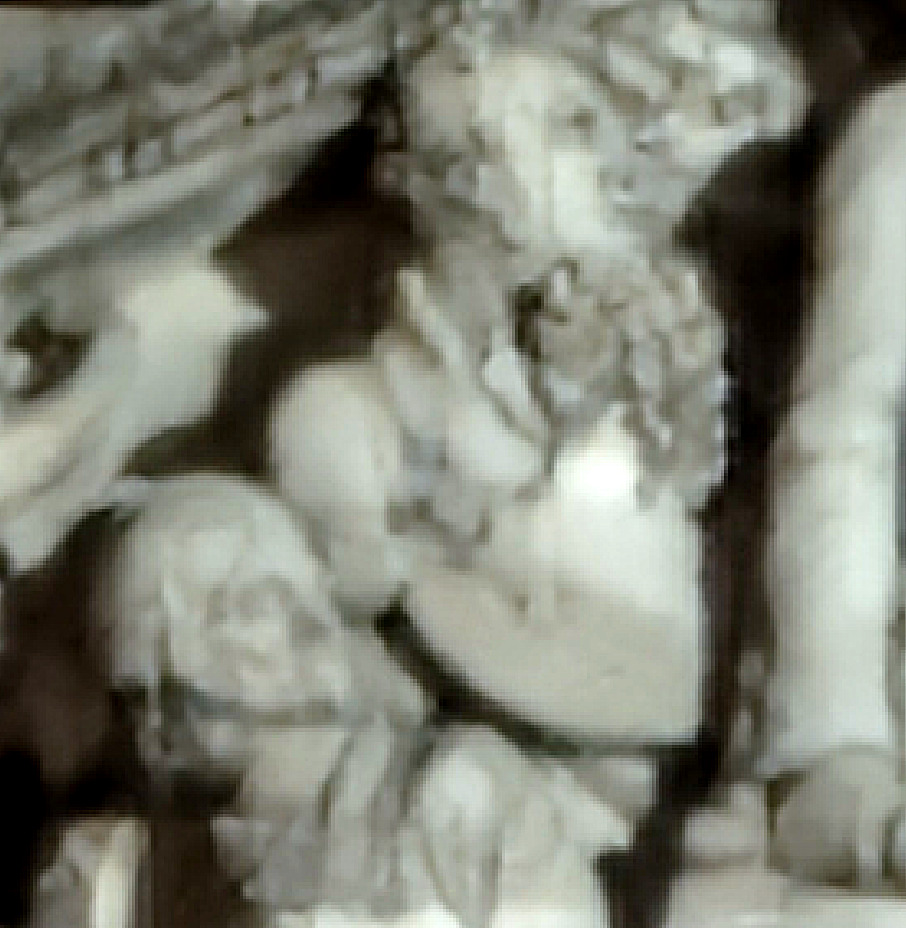}    \\
    \includegraphics[trim=0 0 0 0,clip,height=0.199\textwidth]{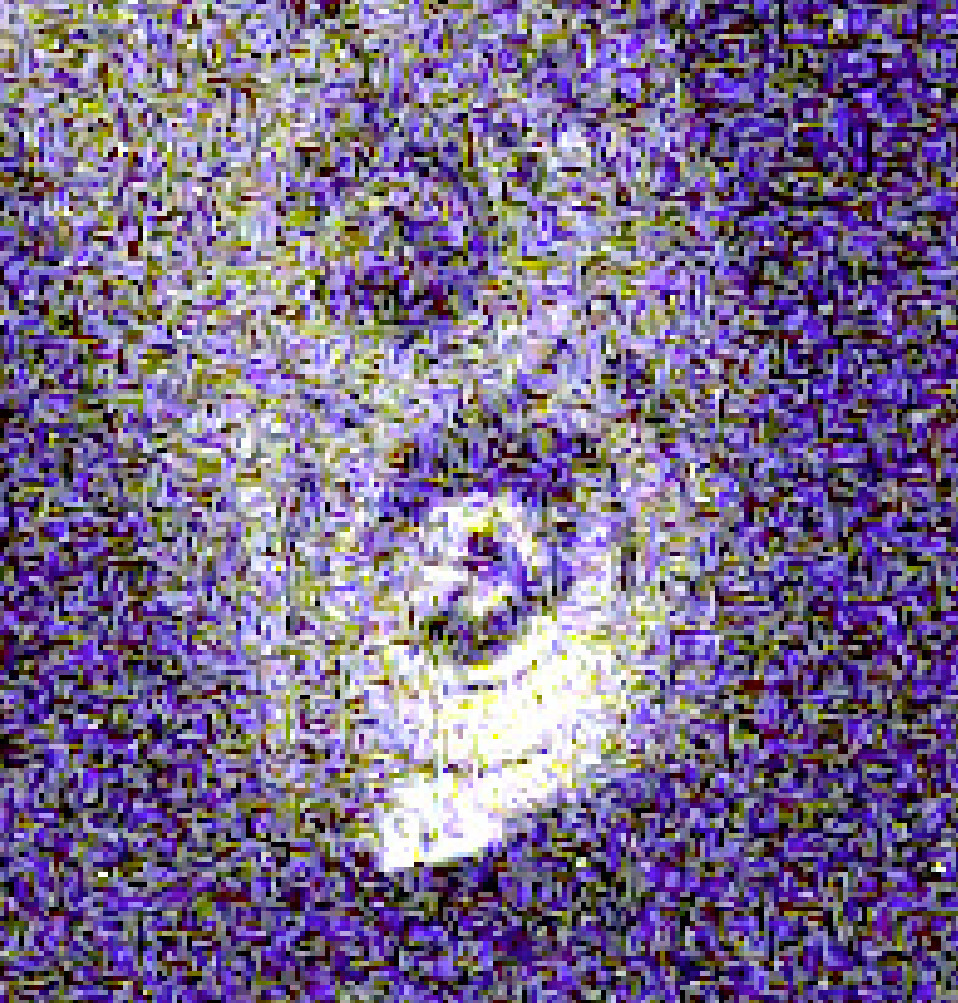}        &  
    \includegraphics[trim=0 0 0 0,clip,height=0.199\textwidth]{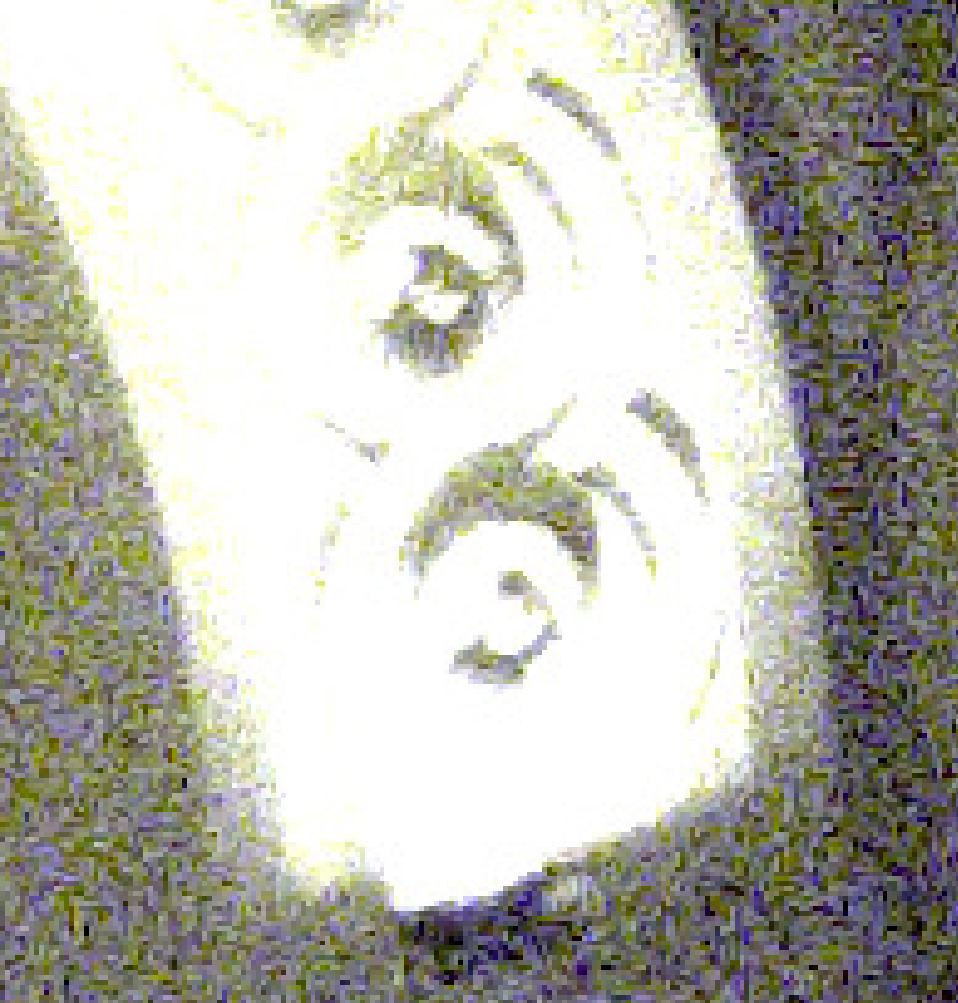}        &  
    \includegraphics[trim=0 0 0 0,clip,height=0.199\textwidth]{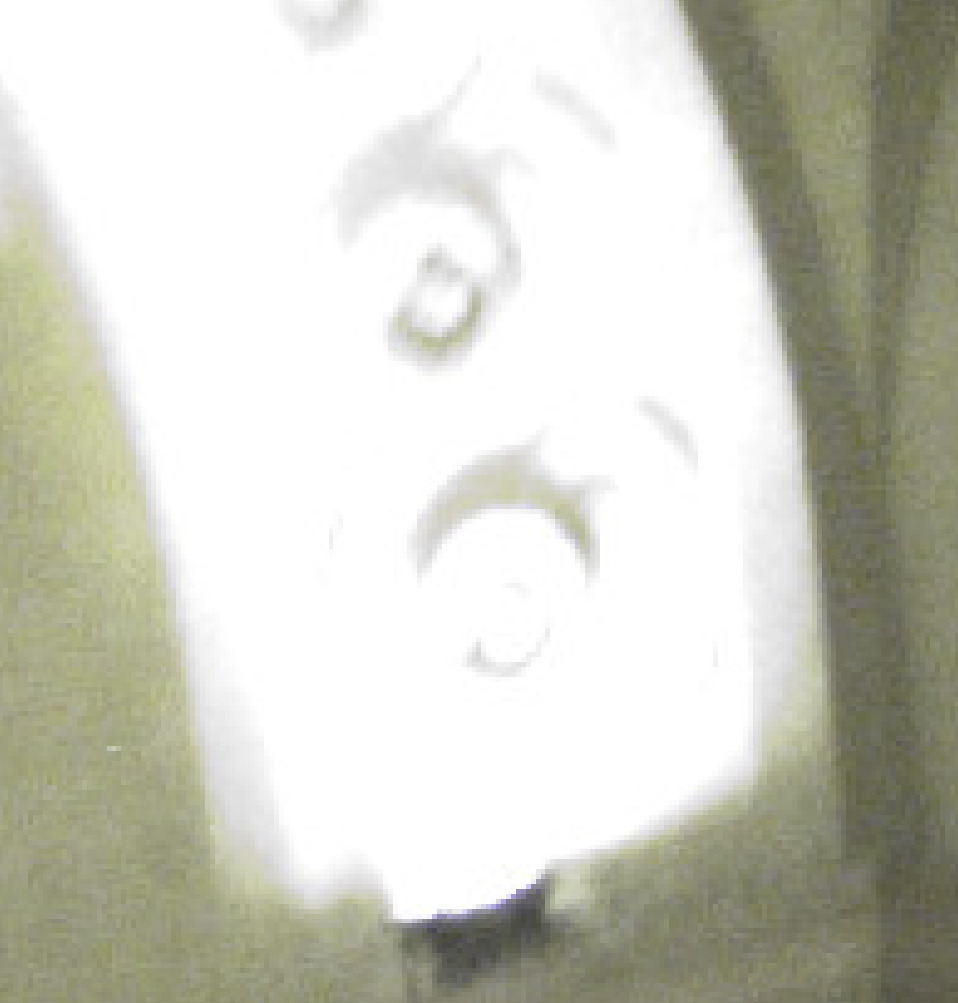}        &  
    \includegraphics[trim=0 0 0 0,clip,height=0.199\textwidth]{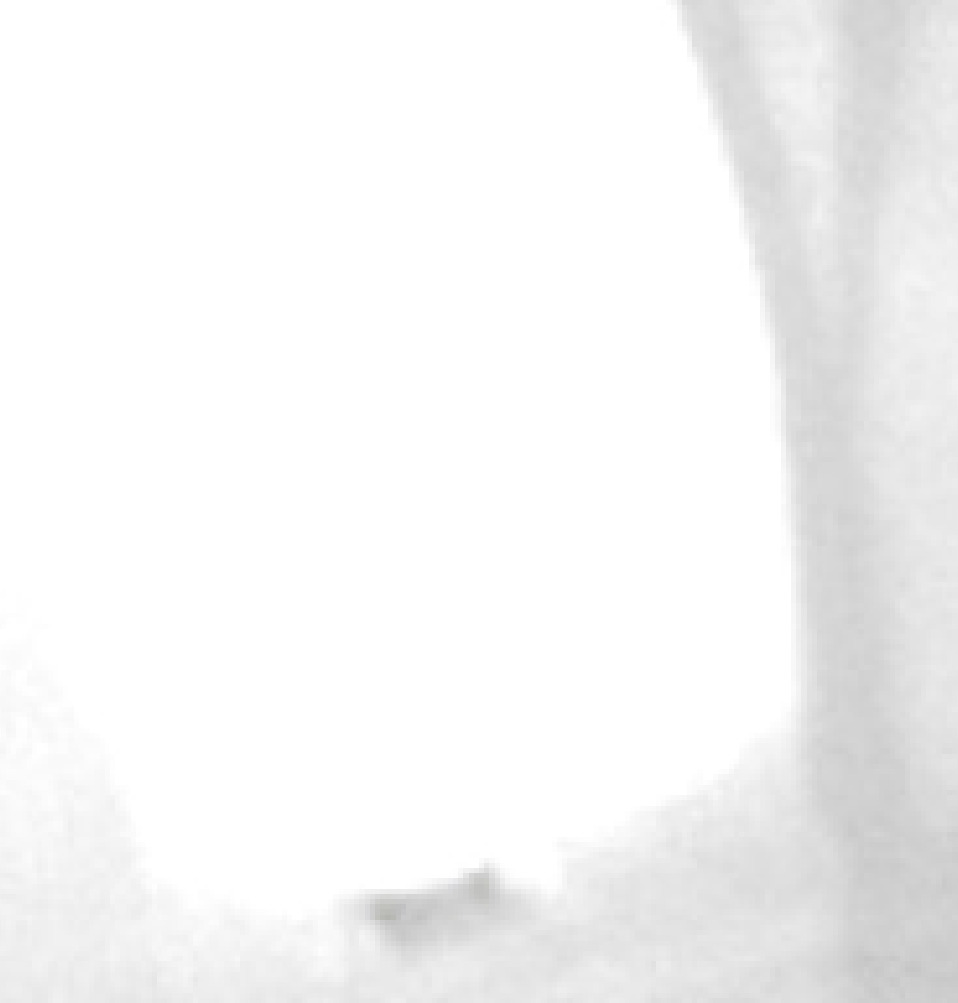}        &  
    \includegraphics[trim=0 0 0 0,clip,height=0.199\textwidth]{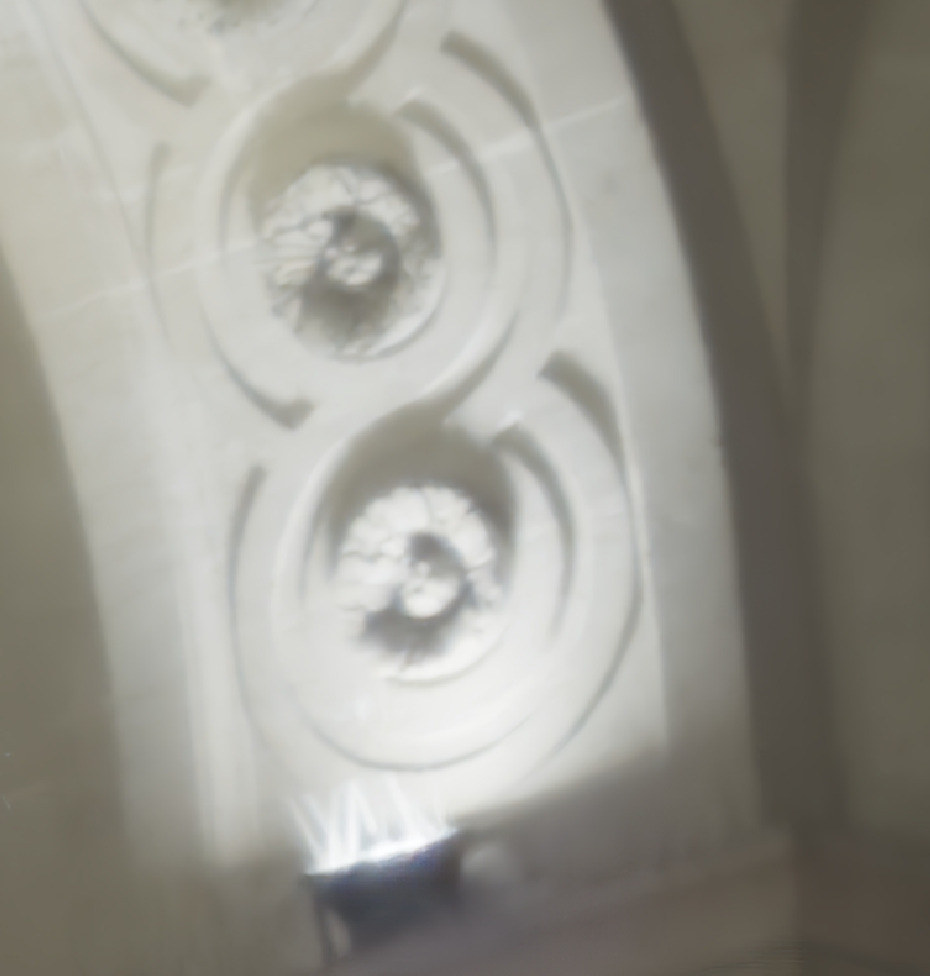}        \\ 
    LR frame 1 &  LR frame 6  & LR frame 12   & LR frame 18     & Ours (HDR+SR $\times 4$)
    \end{tabular}
    \caption{Exposure bracketing: {\bf Left:} Three high dynamic range,
    high-resolution images obtained by our method from 18-image bursts taken
    by a handheld Pixel 4a smartphone with a $\times 4$ super-resolution factor. We
    show post-processed sRGB pictures for the sake of presentation.
    {\bf Right:} Small crops from sample photos in the burst and our
    reconstruction. Note the high level of noise in the short-exposure images, in particular in
    the second row, and the saturated regions in the long-exposure ones. As shown by the
    last column of the figure, our algorithm recovers details in saturated areas and
    remove noise in the darkest regions. The reader is invited to zoom in on a computer screen.}
    \label{fig:samplebracketing}
\end{figure*}

\paragraph{Learning-based methods} for HDR imaging have also been proposed.  \citet{kalantari17deep} introduce a convolutional neural network (CNN) to predict the irradiance from three low-dynamic range (LDR) images, with different exposures, camera poses and possibly moving subjects, pre-aligned with an optical flow algorithm. Most recent CNN-based multi-image methods~\cite{wu18deep, yan19attention, yan20deep, niu21hdrgan, perez21ntire} learn to align and fuse demosaicked images in an end-to-end manner, and they typically
operate on image triplets such as those in the dataset of~\citet{kalantari17deep}.
CNN-based approaches to single-image HDR include \cite{eilertsen17hdr, endo17deep, santos20single, liu20single}. They
rely on machine learning to recover missing details
 in the darkest and saturated areas of tone-mapped images.

\subsection{Super-resolution}
We limit here our discussion to multi-frame super-resolution algorithms. Although single-image learning-based techniques have been used to generate very impressive and highly-detailed images \citep[\it{e.g.}][]{Dahl17,Pulse20}, their objective is not the same as ours: they aim at generating a high-resolution picture {\em compatible} with one input photograph, whereas we want to reconstruct the details that are {\em actually available} in the input burst.

\paragraph{Energy-based methods.}  High-frequency information present in low-resolution (LR) photos with aliasing artefacts is useful for reconstructing a high-resolution (HR) image from multiple LR frames~\cite{farsiu06multi}. Unfortunately, this information is typically lost during the denoising and demosaicking steps performed by the camera ISP pipeline to produce sRGB images.  \citet{farsiu06multi} estimate an HR demosaicked image from a sequence of raw photographs by minimizing a penalized energy---that is, they solve an inverse problem via optimization. \citet{wronski19handheld} adapt the kernel method of \citet{takeda07kernel} and exploit natural hand tremor to jointly demosaick and super-resolve a raw image burst with magnification factors up to $\times 3$ in a fraction of a second on a handheld smartphone. 

\paragraph{Learning-based techniques.}  \citet{bhat21deep} learn a CNN with attention module to align, demosaick and super-resolve a burst of raw images. In a follow-up work,~\citet{bhat21reparameterization} minimize a penalized energy including a data term comparing the sum of parameterized features residuals. 
\citet{lecouat21aliasing}, learn instead a hybrid neural network alternating between aligning the images with the Lucas-Kanade algorithm \cite{lucas81iterative}, predicting an HR image by solving a model-based least-squares problem and evaluating a learned prior function. 
\citet{luo2021ebsr} propose a
neural network architecture that aligns an input burst of images while performing super-resolution with a non-local fusion module.

\subsection{Joint HDR imaging and super-resolution}
The algorithms proposed by~\citet{choi09high, gunturk06high} address joint SR and HDR imaging with an existing SR energy-based solver. To tackle the multi-exposure setting, they introduce weights inspired by bracketing techniques in the least-squares term. 
More generally, this joint image restoration problem has been addressed in a two-stage fashion: (i) image registration with an algorithm robust to varying exposures and (ii) solving a least-squares problem including operators modelling both SR and HDR.  For instance,~\citet{rad07multidimensional} propose an exposure-invariant transform before applying the FFT-based registration technique of \citet{vandewalle06frequency}.  The image is then obtained by solving a penalized least-squares problem.  \citet{zimmer11freehand} use an optical flow approach with normalized gradients for robustness to changes of exposure, and the HR/HDR image is found by solving again a penalized least-squares problem.  \citet{traonmilin14simultaneous} adapt a backprojection algorithm to the multi-exposure setting and simply solve a weighted least-squares problem without prior, with comparable performance but lower computational cost.  \citet{vasu18joint} explore the case where the LDR SR images are also blurred with camera shake or motion blur. 
Similar to the HDR case, CNNs have also been proposed
for single-image joint SR and HDR, \eg\cite{kim19deep}, while \citet{deng21deep} address instead joint SR, HDR and tone mapping
by merging a pair of previously aligned over- and under-exposed images with a two-stream CNN.
In contrast with these techniques, we use trainable image features to adapt the raw image registration module of \citet{lecouat21aliasing} to the varying-exposure setting in a robust manner, and jointly learn these features and a parametric image prior in an end-to-end manner.

Figure~\ref{fig:samplebracketing} shows examples of the input data these methods
use and samples of the the predicted high-resolution HDR images we predict
with the proposed approach.

\section{Image formation model}
We now describe the process generating a burst of low-dynamic low-resolution raw images from a high-resolution HDR image. This process yields a natural inverse problem formulation,
which we will leverage later to build a trainable architecture.

\subsection{Dynamic range}
\label{sec:dynamicalrange}
After analog-to-digital conversion, a camera sensor outputs a
black-and-white mosaicked image whose pixel values are integers
obtained by quantizing the number of photons collected by each photosite on a linear $q$-bit scale~\cite{clark06sensor}, where $q$ is called
the {\em bit depth} of the sensor. We denote by $P_q$ the set of the discrete values a pixel may take, as measured in {\em data numbers}
(or {\em DNs}~\cite{martinec08noise, clark06sensor}), 
from 0 to $2^{q}-1$.

The dynamic range $R(u)$ for a pixel $u$
is defined as the ratio of the largest to the
smallest values this pixel may take:
the larger the bit depth of the sensor, the greater is its maximal
value in $P_q$.
The ratio is usually given in photographic {\em stops}, where each
stop corresponds to a multiple of 2.
In practice, the largest value $u$ can take 
is limited either by the bit depth~$q$ or the 
white level $c$ set by the camera, to prevent color artefacts 
in highlights~\cite{lujik07dcraw}, whereas
the lowest value is actually limited by the noise $\varepsilon(u)$ and by the camera black level~$b$~\cite{foi08practical}. Note that even in the absence of light, $\varepsilon(u)$ is never 0 since  any digital camera suffers from various sources of electronic noise~\cite{hasinoff10noiseoptimal}. This also shows that increasing dynamic range is strongly related to denoising, as discussed later in this section.

\subsection{Exposure}
As mentioned above, raw pixel values depend linearly on the number of
photons captured by each photosite (ignoring quantization effects) and thus on exposure time.
In photography, this effect is quantified by the {\em exposure value} (or {\em EV}):
Increasing it by +1EV (resp. decreasing by -1EV)
corresponds to doubling (resp. halving) the raw pixel values.
The EV depends on the {\em ISO gain}, aperture size and exposure time.
In this work, we will only control the exposure time $\Dt$, keeping
it small enough to (mostly) avoid motion blur, and keep the
other two quantities constant since modifying the ISO gain may change the noise 
distribution~\cite{hasinoff10noiseoptimal} and
adjusting the aperture size changes the blur of out-of-focus regions~\cite{levin07image}.

The raw value $y(u)$  in $P_q$ recorded at some pixel $u$ is thus related to the irradiance $x(u)$  in $\R^+$ at the same location by
\begin{equation}\label{eq:exposure}
    y(u) = S(\Dt x(u)),
\end{equation}
where $S$ is the function mapping pixel values
from $\R^+$ to $P_q$. This equation is only valid when
$S(\Dt x(u))<2^q-1$, with saturation occurring for higher
values. Using short exposure times limits saturation, but,
as shown in the next section, leads to  a poor {\em signal-to-noise
ratio} (or {\em SNR}).

\begin{figure}
    \centering
    \includegraphics[width=0.4\textwidth,trim=15 18 0 10,clip]{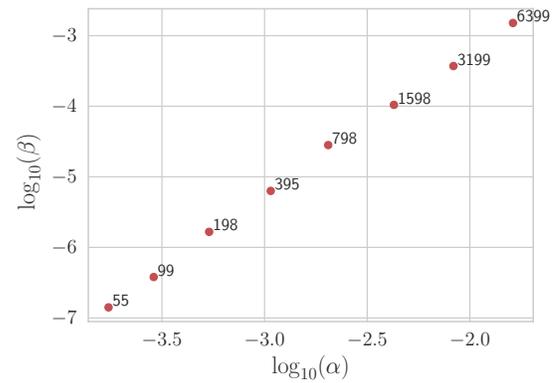}
    \caption{Empirical measurements of the shot and read noise levels $\alpha$ and $\beta$ from 
    the metadata of raw images taken with the Google Pixel3a smartphone.
    The numbers next to the markers are the corresponding 
    empirical ISO levels.
    As observed by \citet{brooks19unprocessing}, there exists a linear 
    relationship between $\log_{10}(\alpha)$ and $\log_{10}(\beta)$ that we
    leverage to train our models. 
    }
    \label{fig:isoplot}
\end{figure}

\subsection{Noise and SNR}
The raw image noise $\varepsilon(u)$ at each pixel
comes from the physics of light and the 
electronics of the camera.
The former is called {\em shot} noise, and it can be
modelled with a Poisson distribution~\cite{foi08practical}.
The latter is often referred to as {\em read} noise and corresponds
to random signal fluctuations caused by the electronics and 
quantization effects. 
It is usually modelled with a zero-mean Gaussian distribution~\cite{foi08practical}.
The combination of shot and read 
noise can be modelled by a single random variable $\varepsilon(u)$ following
a zero-mean Gaussian distribution
with pixel-dependent standard deviation, defined for any pixel value $y(u)$ as~\cite{foi08practical, brooks19unprocessing, plotz17benchmarking}:
\begin{equation}\label{eq:noisevariance}
    s(u) = \sqrt{\alpha y(u) + \beta},
\end{equation}
where $\alpha$ and $\beta$ are respectively the variances of the 
shot and read noise.
Figure~\ref{fig:isoplot} shows the distribution of $\alpha$ (shot noise level) and $\beta$ (read noise level) for the Google
Pixel3a camera. We have obtained theses values from
the EXIF metadata of raw images taken with the smartphone.
Each marker corresponds to a couple 
$(\log_{10}(\alpha), \log_{10}(\beta))$ for an ISO level.
In dark regions, read noise dominates shot noise, and
limits the total dynamic range. 
\comment{**J"ai enleve ca car je pense que c'est mysterieux
et ne sers a rien.On reparler de la rellation entre range et bruit juste en dessous et plus clairement **For instance, the Canon EOS 7D camera has a measured 
read noise with standard deviation of $2.36$DN~\cite{martinec08noise, clark06sensor}
for ISO value of 400. This highlights again the crucial link between high dynamic range and 
low signal-to-noise ratio (SNR).} 

For the Poissonian-Gaussian noise model of Eq.~\eqref{eq:noisevariance}, the SNR is:
\begin{equation}\label{eq:snr}
    \text{SNR}(u) = \frac{m(u)y(u)}{s(u)}
     = \frac{m(u)y(u)}{\sqrt{\alpha y(u) + \beta}},
\end{equation}
where $m$ is a binary mask excluding the saturated pixels.
It is a monotonically increasing function of the pixel value $y(u)$,
essentially linear in dark regions (\eg shadows) 
where read noise dominates shot noise, and essentially proportional to
$\sqrt{y(u)}$ in bright regions (\eg highlights) where the opposite occurs~\cite{granados10optimal}.
As already discussed in the previous section, noise removal
is essential for generating images with high dynamic range, and
\comment{Eq.~\eqref{eq:snr} suggests that the largest SNR gain comes from removing
read noise.}
Equation~\eqref{eq:snr}  shows that high raw pixel values lead to better SNR and thus better dynamic range in both dark and bright image regions.
But high pixels values everywhere in an image can typically only be
achieved at the cost of saturating the brightest areas.
Exposure bracketing 
avoids this problem by using the longest exposures
to eliminate read noise from dark regions  and the shortest ones
to avoid saturation in bright spots.

\subsection{Overall image formation model}
The original analog 
image cannot be recovered on a computer
and we instead focus on estimating a discrete HR/HDR  $sh\times sw \times 3$ photograph $x$ with pixel values in $\R_+$
from a burst of $K$ raw
LR and LDR images $y_k$ ($k=1,\dots,K$) of size $h \times w$ with entries in $P_q$.
The integer~$s$ is the super-resolution factor.
Following~\cite{lecouat21aliasing}, let us introduce the warp operator $W_k$ associated with the $k$th photo in the burst and accounting for camera shake, the blur operator $B$ taking into account the integration of the signal over the pixel area is modeled by a convolution, the decimation operator $D_s$ associated with the super-resolution factor $s$, and the $C$ operator is a binary mask modeling the sensor CFA.  Putting them together and taking into account the exposure time $\Delta t_k$, the analog low-resolution image associated with the irradiance image $x$ is $a_k=CD_sBW_k(\Delta t_k x)$, which can be rewritten as $a_k=A_kx$, where $A_k=\Delta t_k CD_sBW_k$ (the factor $\Delta t_k$ commutes with the operators since it only scales the image values).

Combining this model with Eq.~\eqref{eq:noisevariance}
and~\eqref{eq:exposure} yields, for all $k=1,\dots,K$:
\begin{equation}\label{eq:formatinmodel}
  y_k = S\left( A_k x + \varepsilon_k \right),
\end{equation}
where we abuse the notation so $S$ operates on a whole image instead of a scalar, and $\varepsilon_k$ is a zero-mean Gaussian noise with pixel-dependent variance $\alpha A_kx + \beta$ according to Eq.~\eqref{eq:noisevariance}.  The operator $CD_sB$ impacts the spatial resolution, while $S$ and the noise variance limit the dynamic range of each image $y_k$.

Note that our model assumes that the scene is static during burst acquisition, which may result in ghosting artefacts in the presence of scene motion, when using this model within an inverse problem formulation. We will, however,
introduce in the next section simple weighting strategies to make our approach robust
to moderate scene motion.

\begin{figure*}
    \setlength\tabcolsep{0.5pt}
    \renewcommand{\arraystretch}{0.5}
    \centering
    \scalebox{1.02}{
    \begin{tabular}{cccccccccc}
    \includegraphics[trim=0 0 0 0,clip,width=0.095\textwidth]{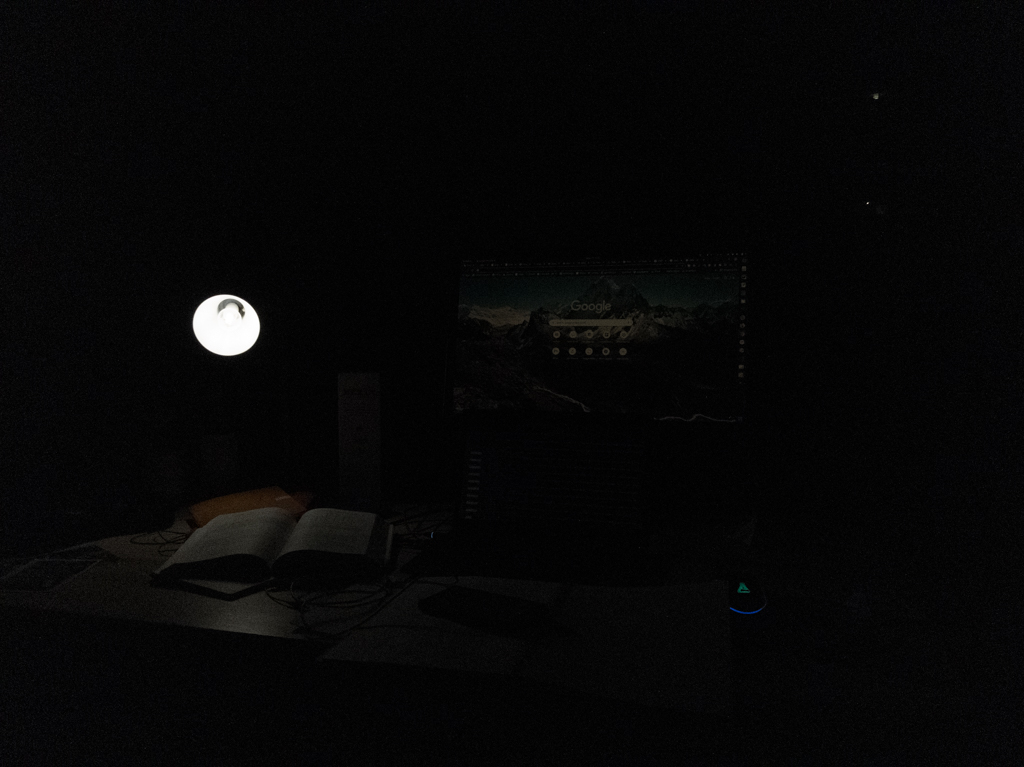}  &
    \includegraphics[trim=0 0 0 0,clip,width=0.095\textwidth]{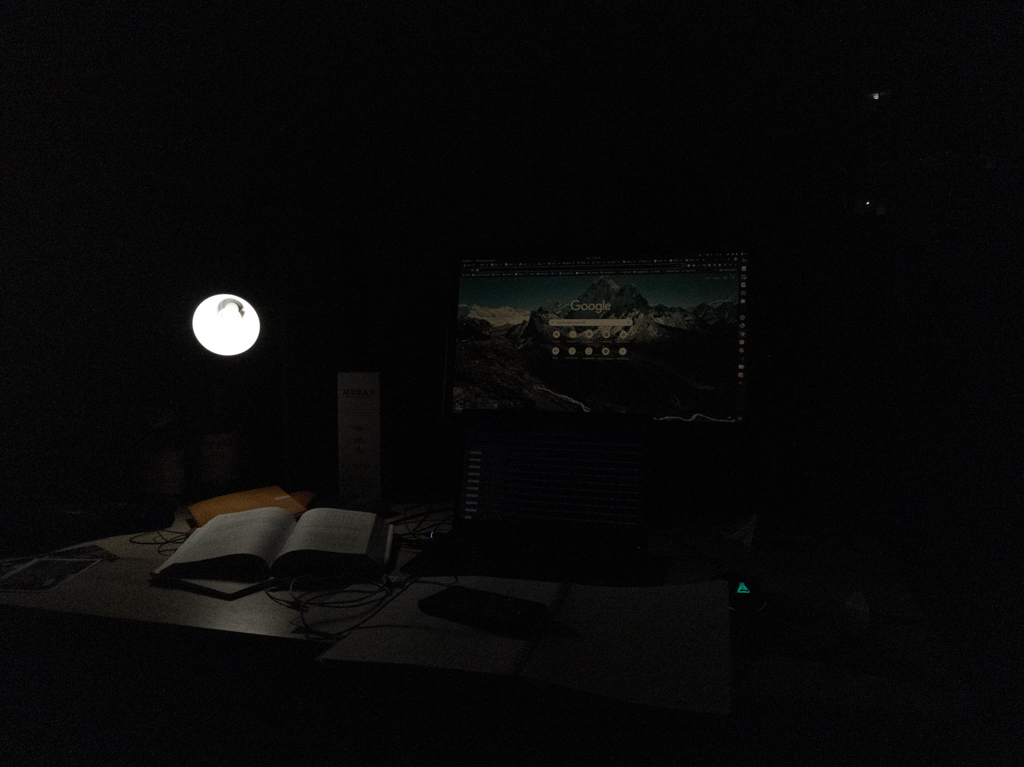}        &  
    \includegraphics[trim=0 0 0 0,clip,width=0.095\textwidth]{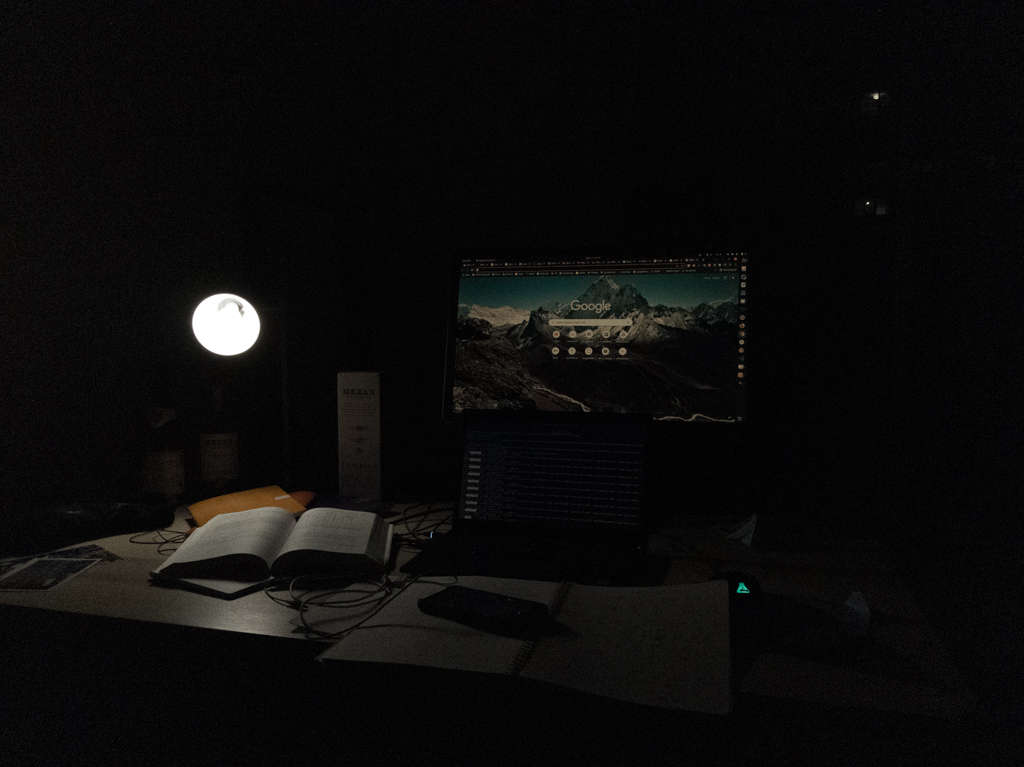}        &  
    \includegraphics[trim=0 0 0 0,clip,width=0.095\textwidth]{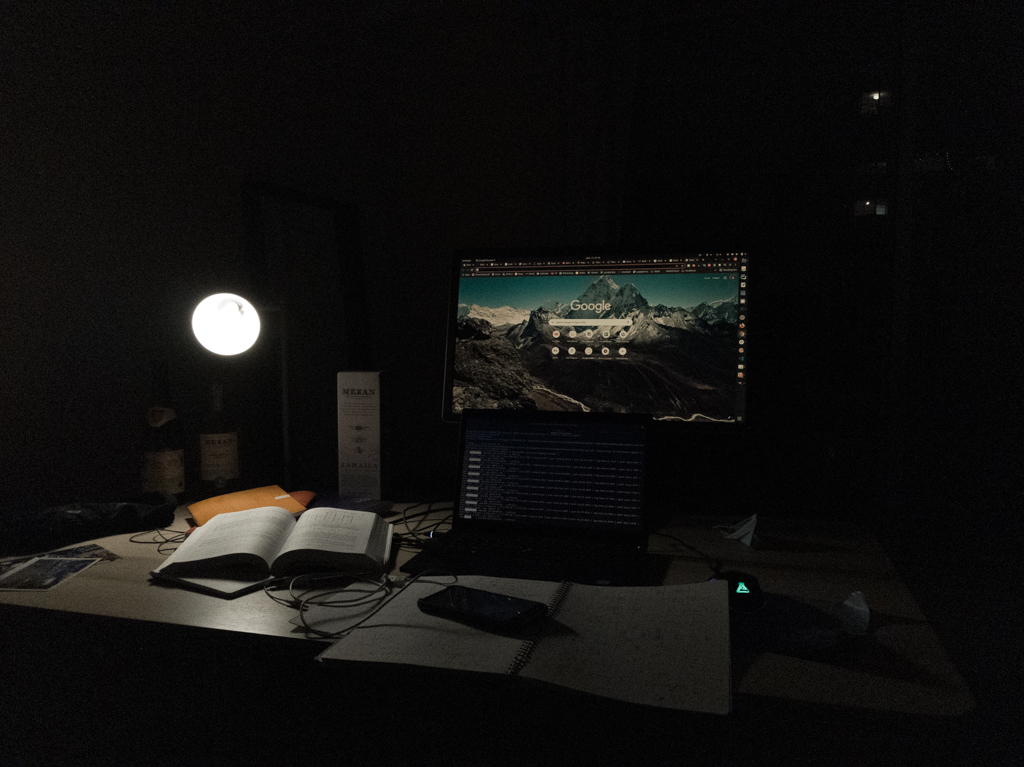}        &  
    \includegraphics[trim=0 0 0 0,clip,width=0.095\textwidth]{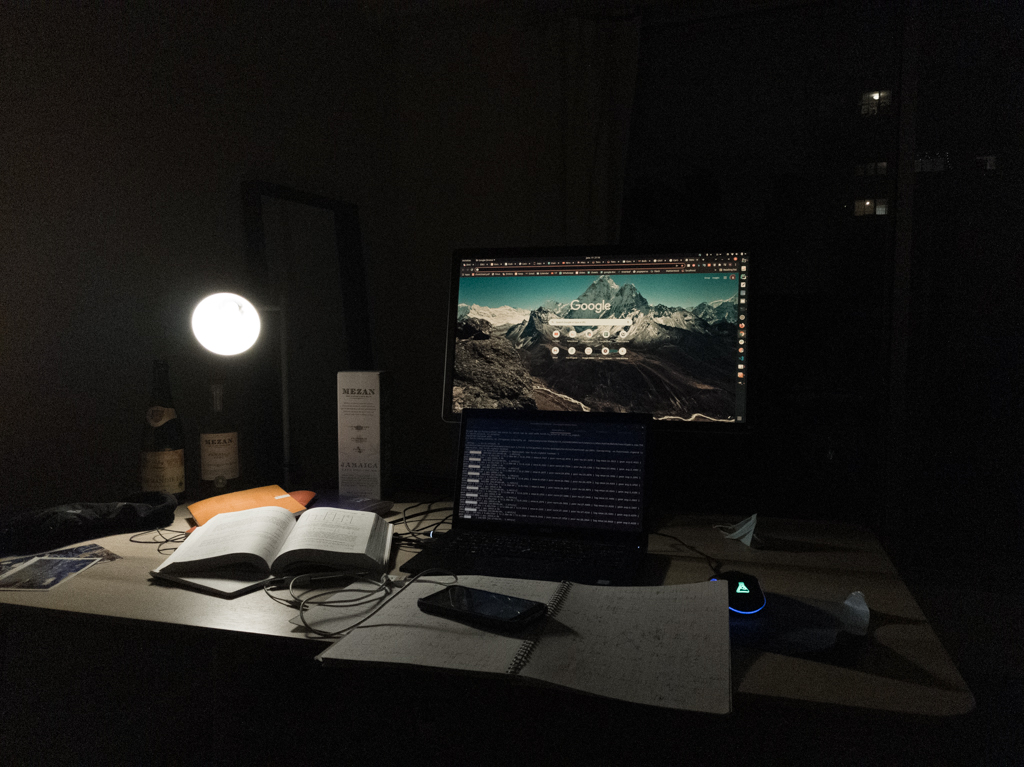}        &  
    \includegraphics[trim=0 0 0 0,clip,width=0.095\textwidth]{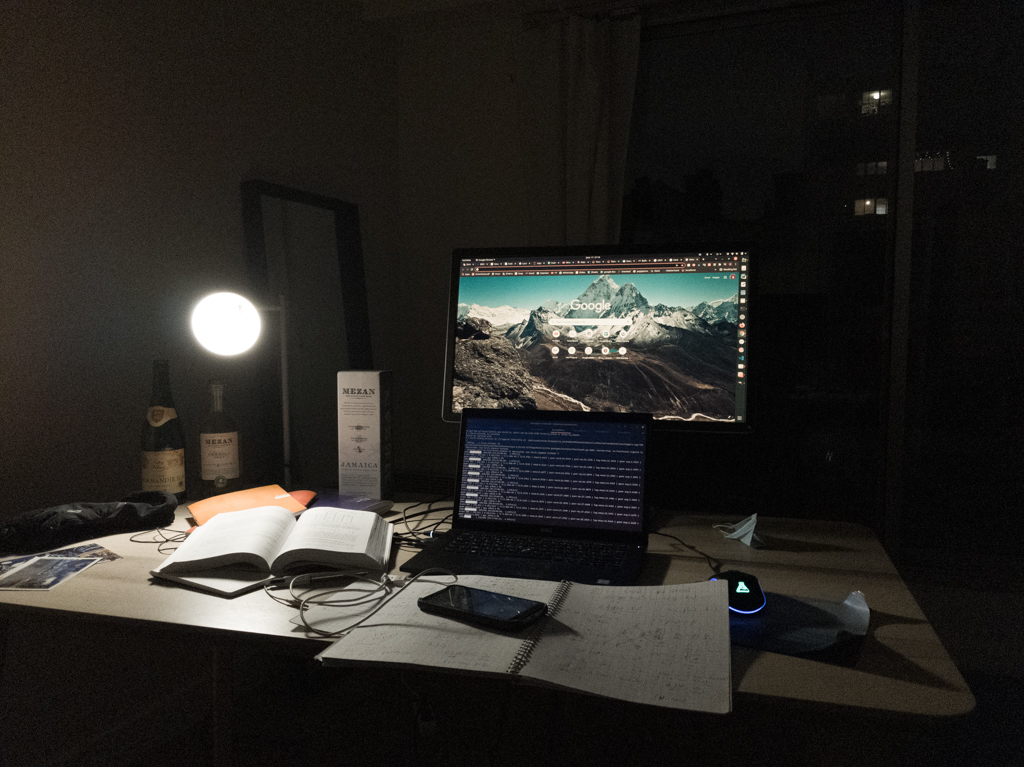}        &  
    \includegraphics[trim=0 0 0 0,clip,width=0.095\textwidth]{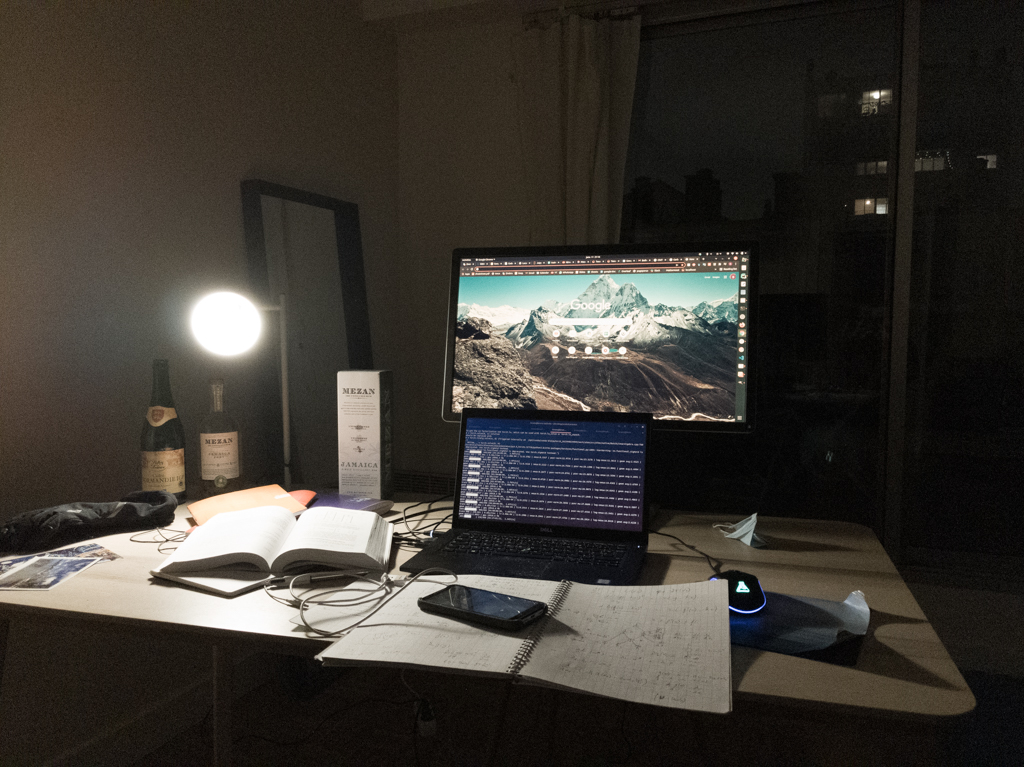}        &  
    \includegraphics[trim=0 0 0 0,clip,width=0.095\textwidth]{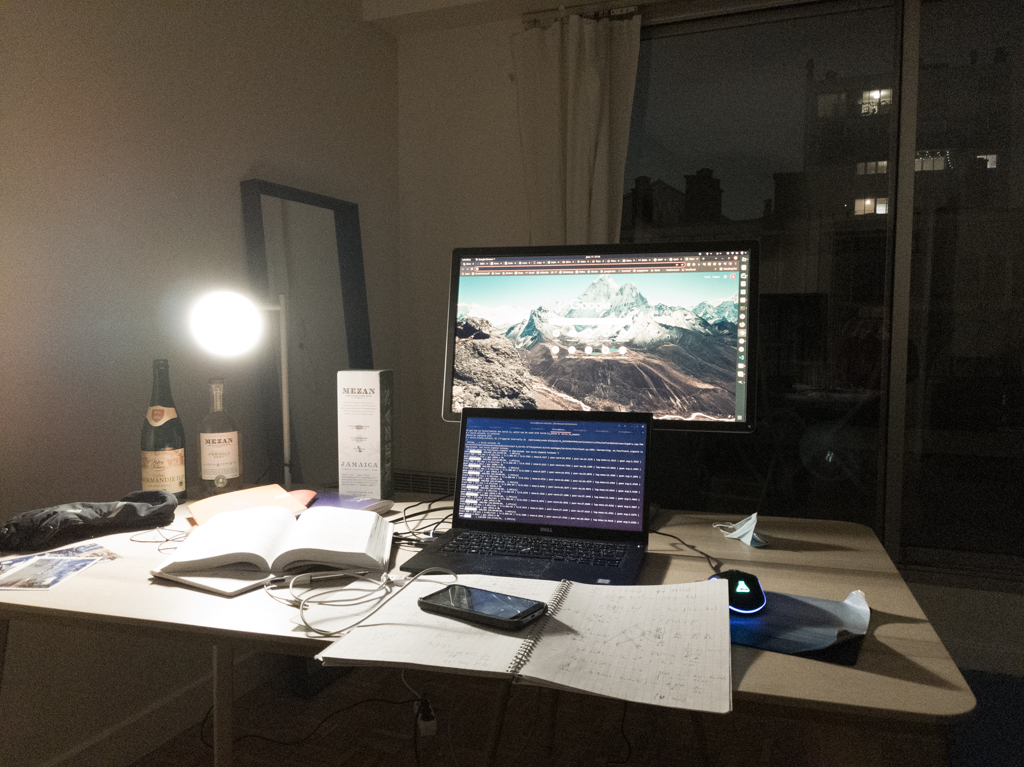}        &  
    \includegraphics[trim=0 0 0 0,clip,width=0.095\textwidth]{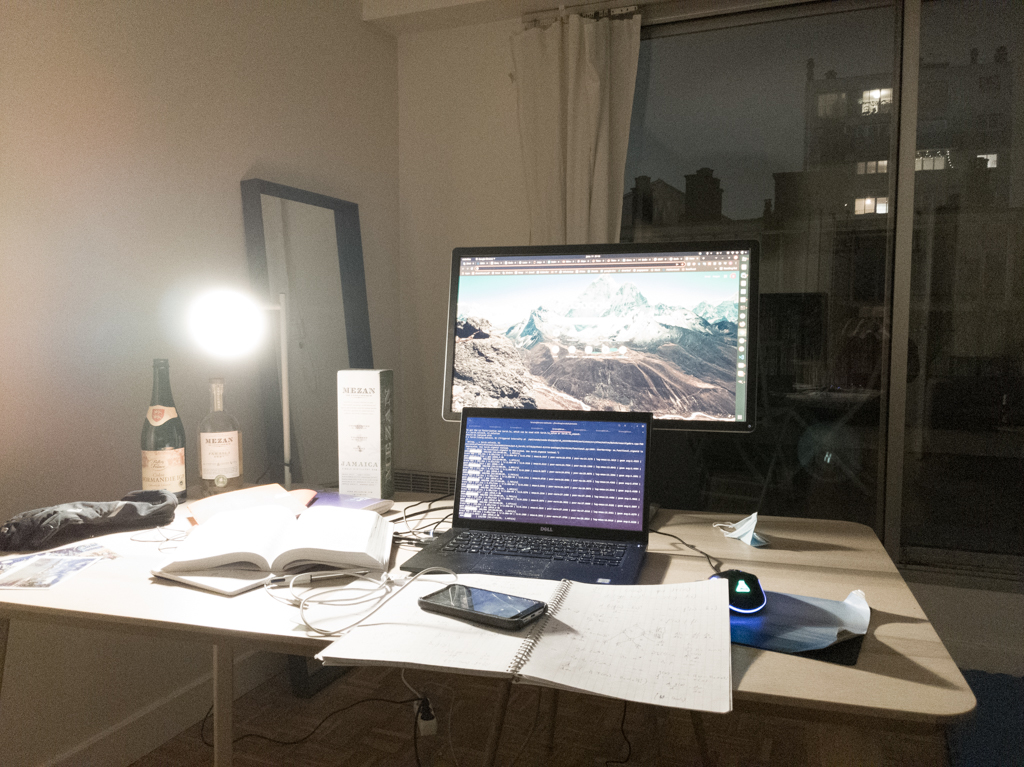}        &  
    \includegraphics[trim=0 0 0 0,clip,width=0.095\textwidth]{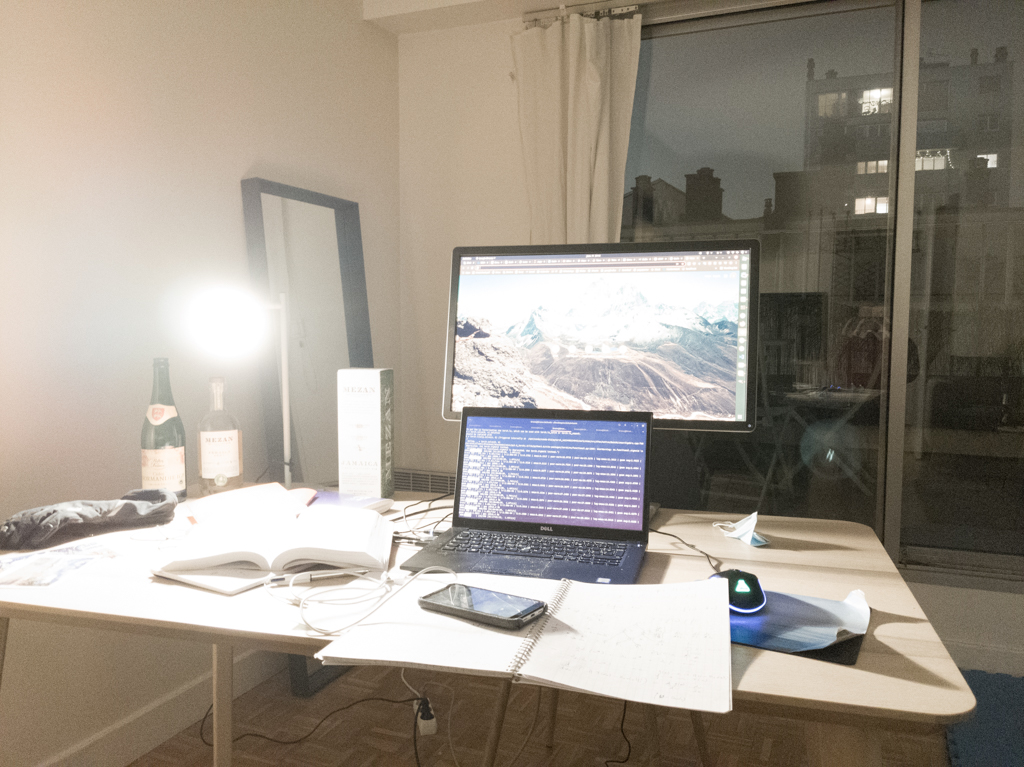}        \\
    \end{tabular}}
    \scalebox{1.02}{
    \begin{tabular}{cccccc}
    \multicolumn{3}{c}{\includegraphics[trim=0 0 0 0,clip,width=0.48\textwidth]{images/dyn/lampe1/lampe6.jpg}} &
    \multicolumn{3}{c}{\includegraphics[trim=0 0 0 0,clip,width=0.48\textwidth]{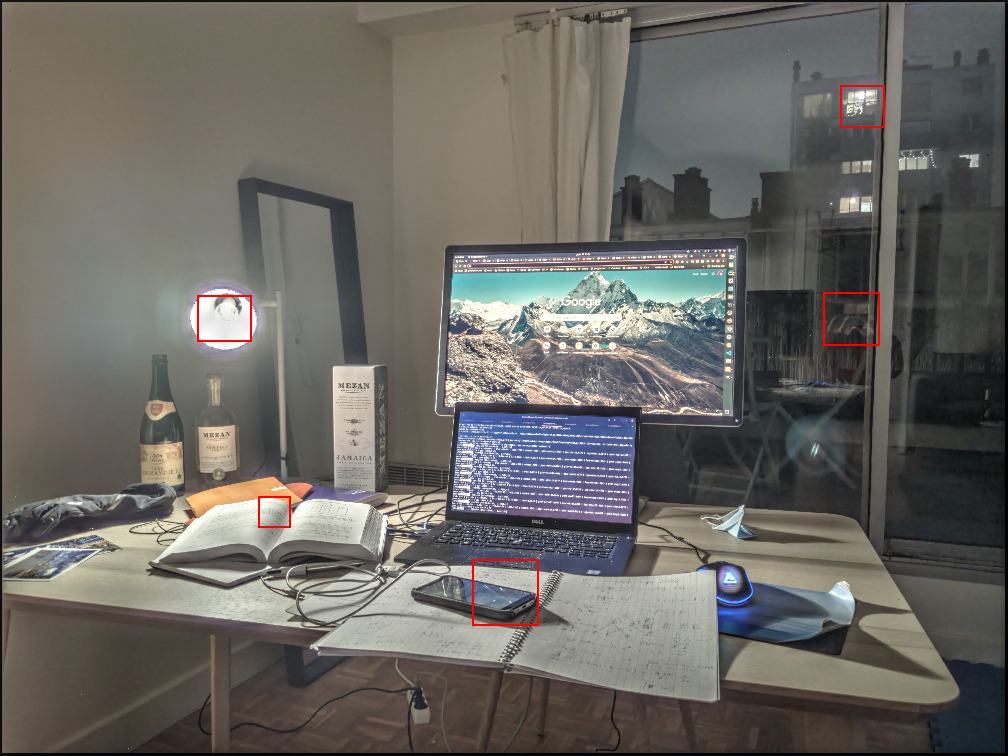}} \\
    \includegraphics[trim=0 80 0 40,clip,cfbox=blue 0pt 0pt,width=0.16\textwidth]{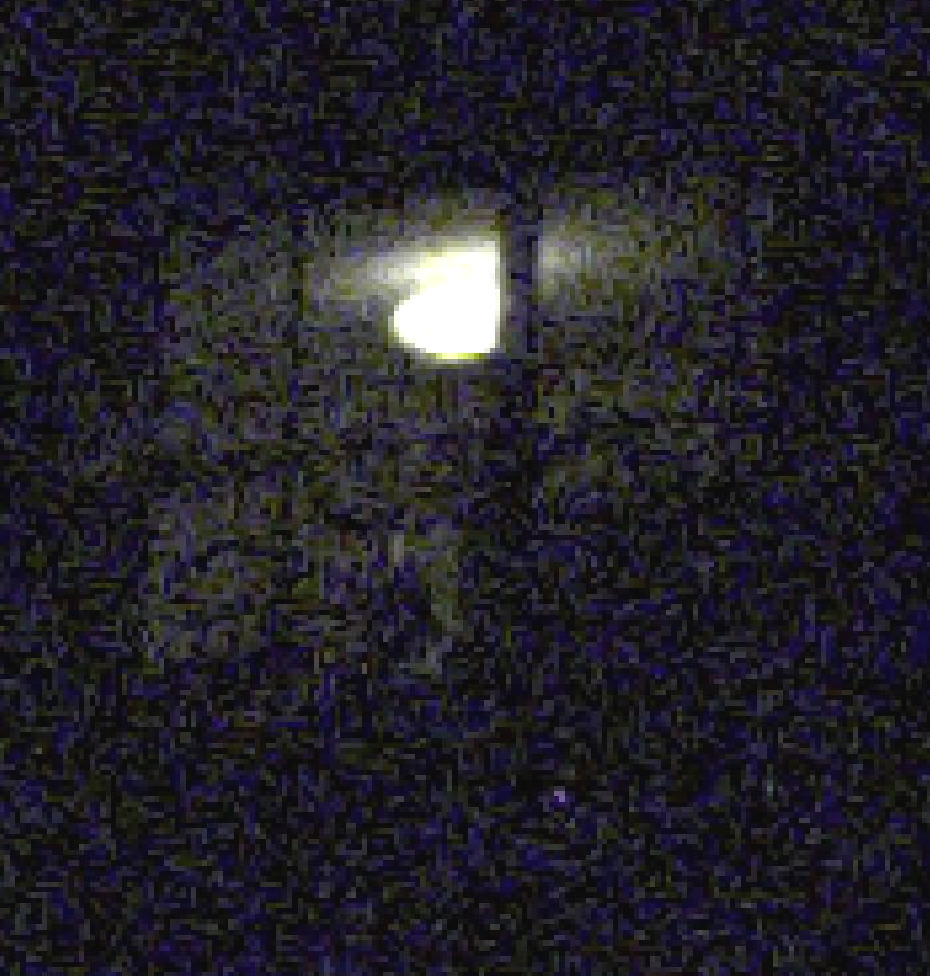}   & 
    \includegraphics[trim=0 80 0 40,clip,cfbox=red 0pt 0pt,width=0.16\textwidth]{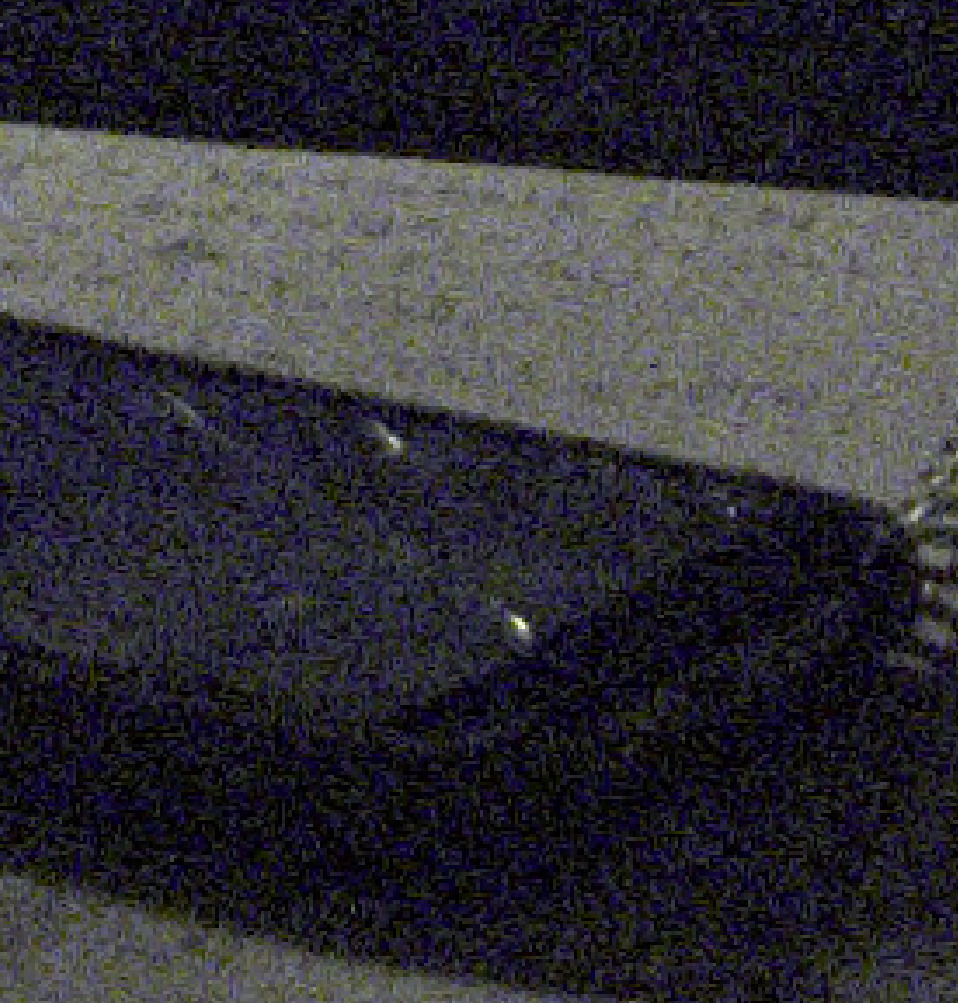}   & 
    \includegraphics[trim=0 80 0 40,clip,cfbox=red 0pt 0pt,width=0.16\textwidth]{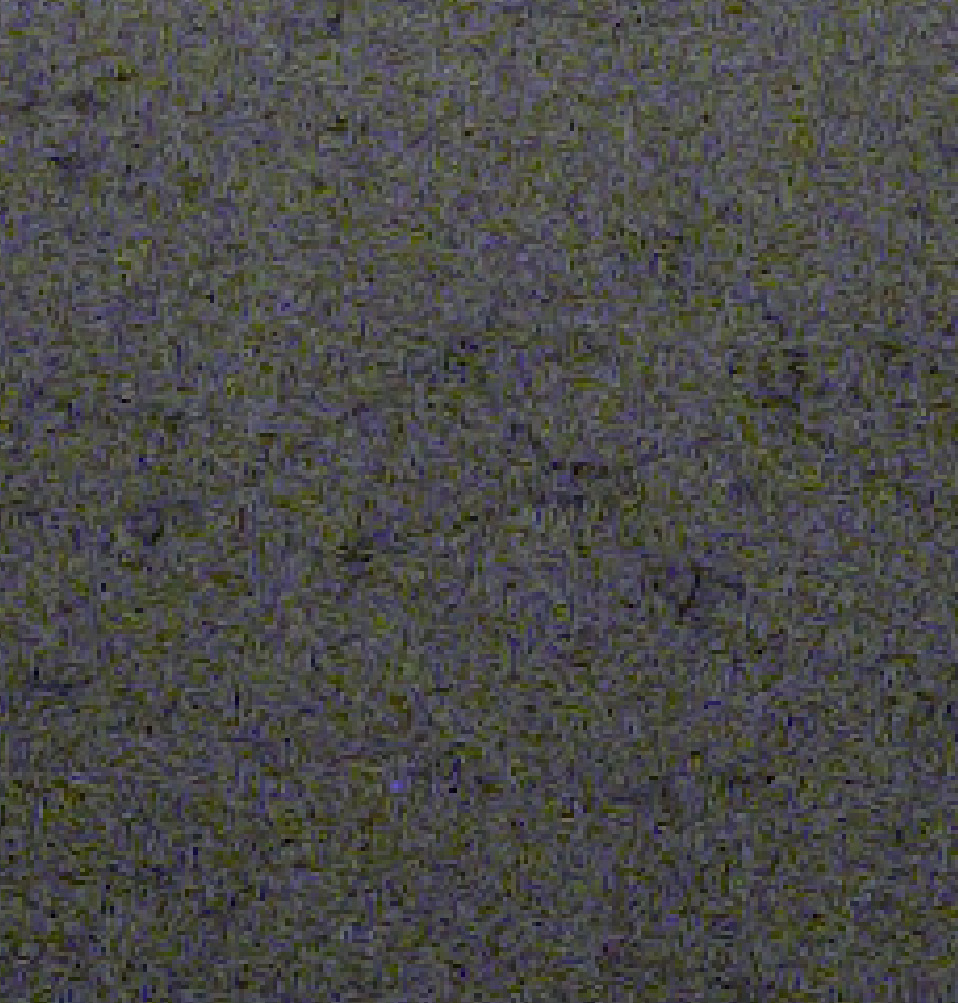}   & 
    \includegraphics[trim=0 80 0 40,clip,cfbox=red 0pt 0pt,width=0.16\textwidth]{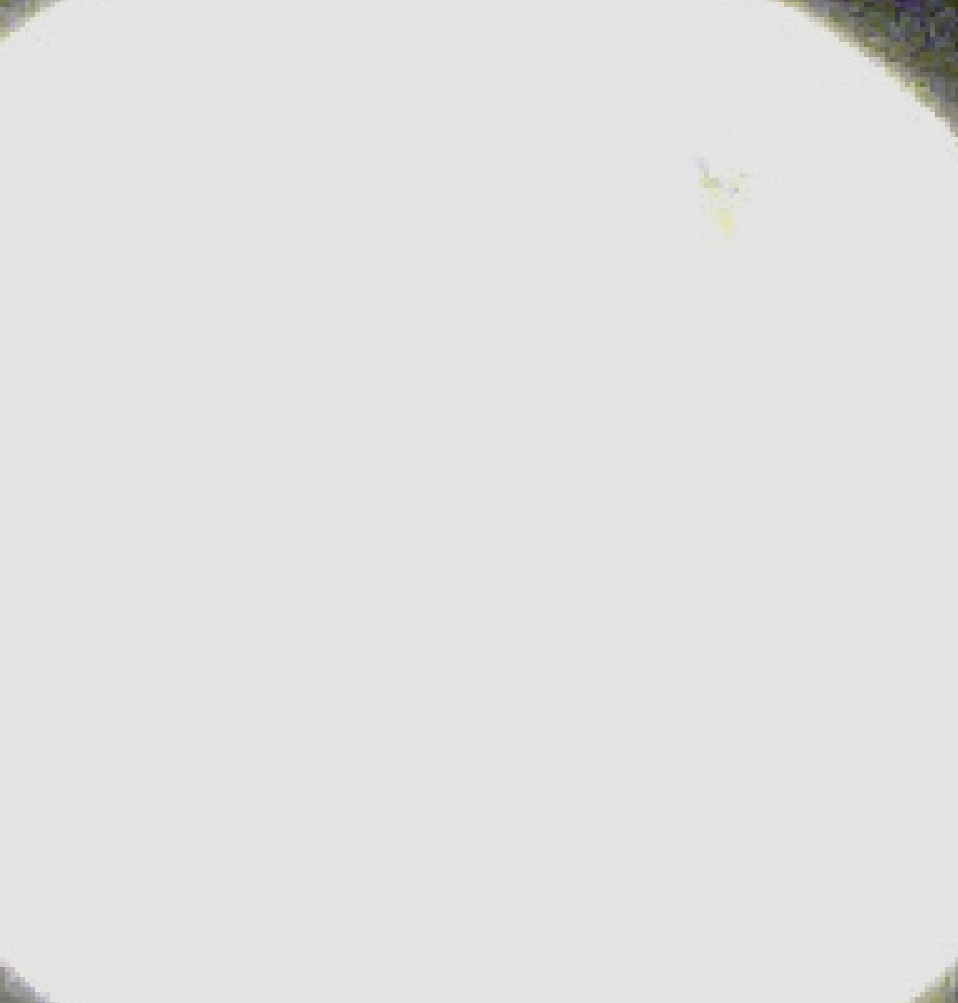}   & 
    \includegraphics[trim=0 80 0 40,clip,cfbox=red 0pt 0pt,width=0.16\textwidth]{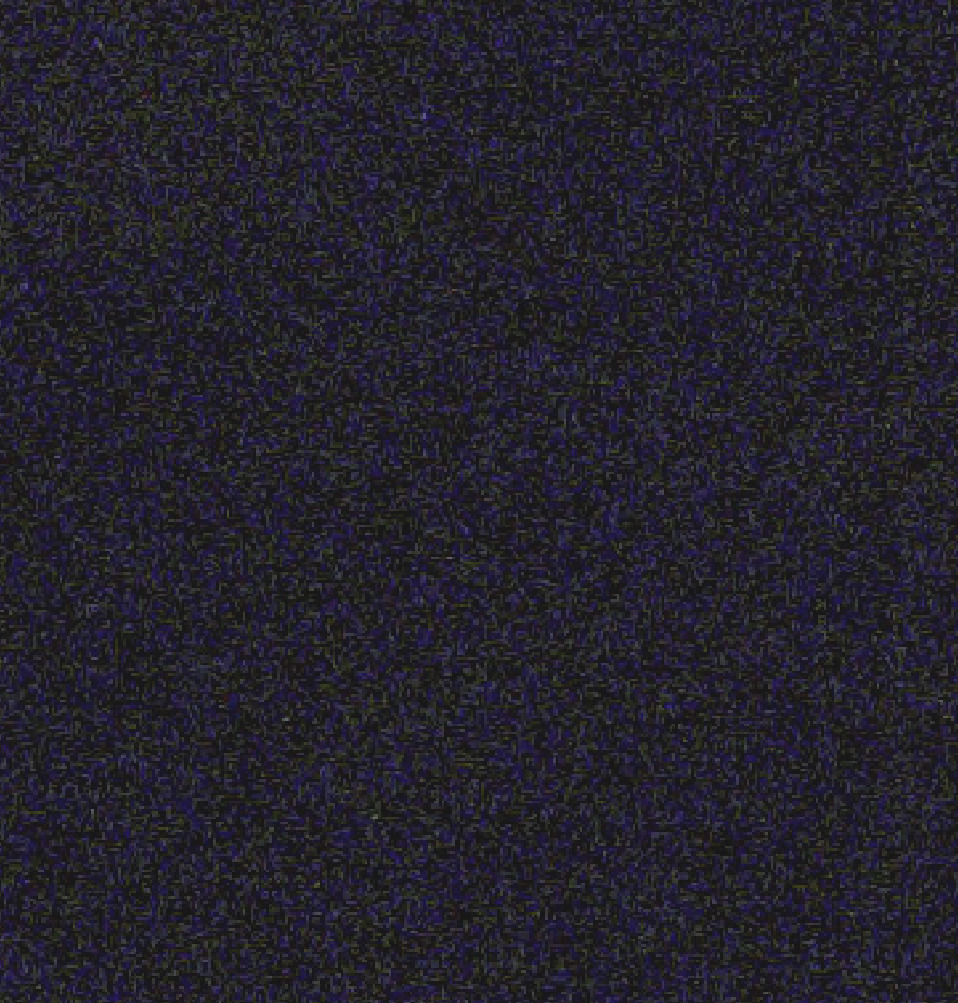}   & 
    \includegraphics[trim=0 80 0 40,clip,cfbox=red 0pt 0pt,width=0.16\textwidth]{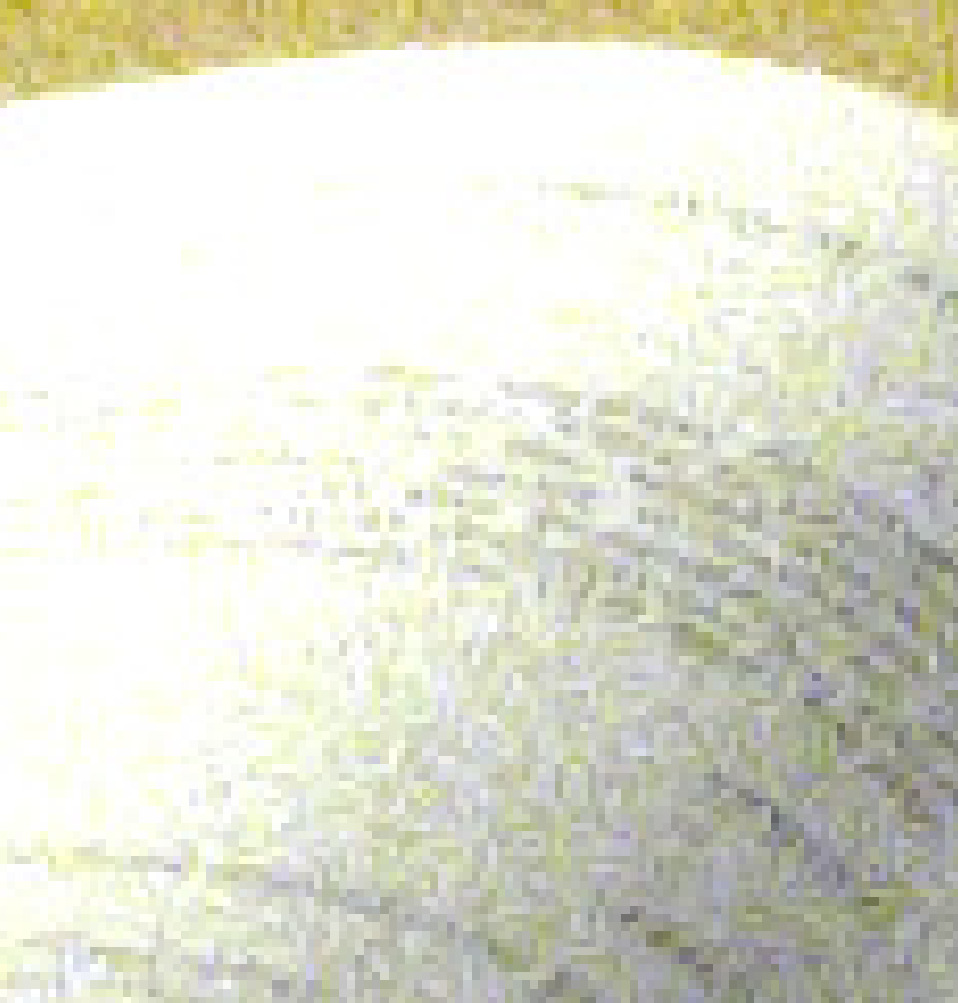}   \\
    \includegraphics[trim=0 80 0 40,clip,cfbox=blue 0pt 0pt,width=0.16\textwidth]{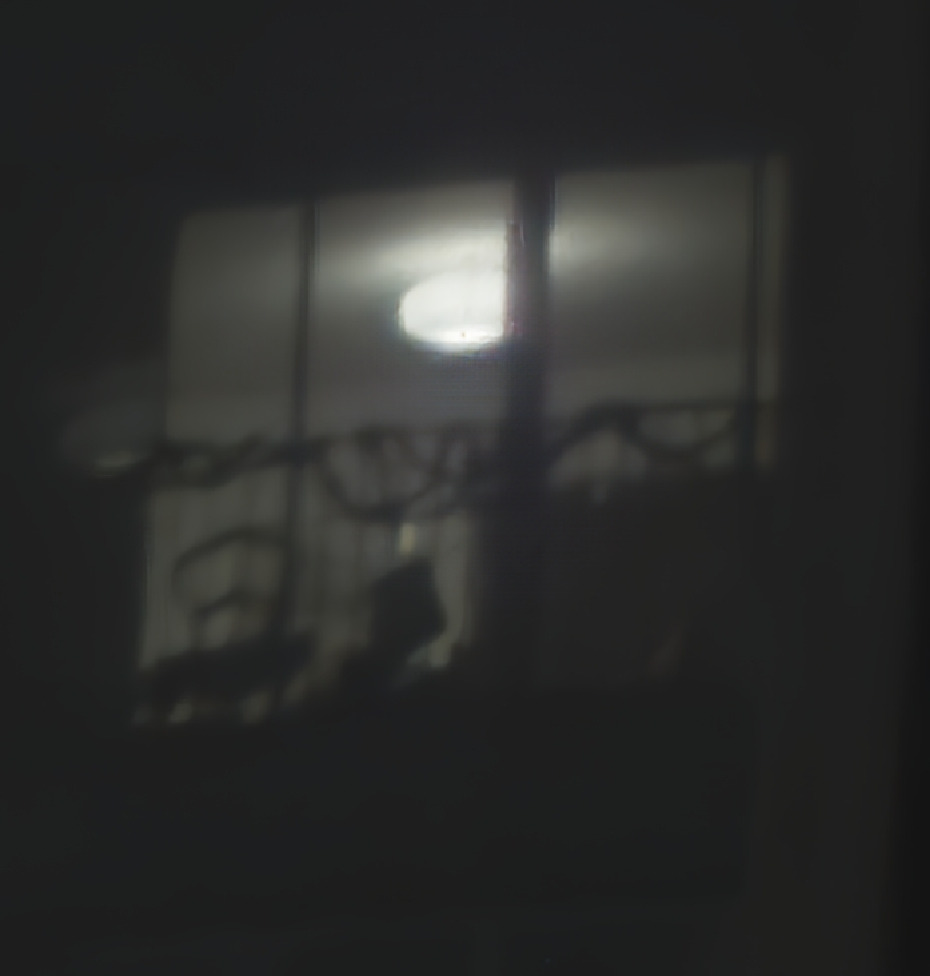}   & 
    \includegraphics[trim=0 80 0 40,clip,cfbox=red 0pt 0pt,width=0.16\textwidth]{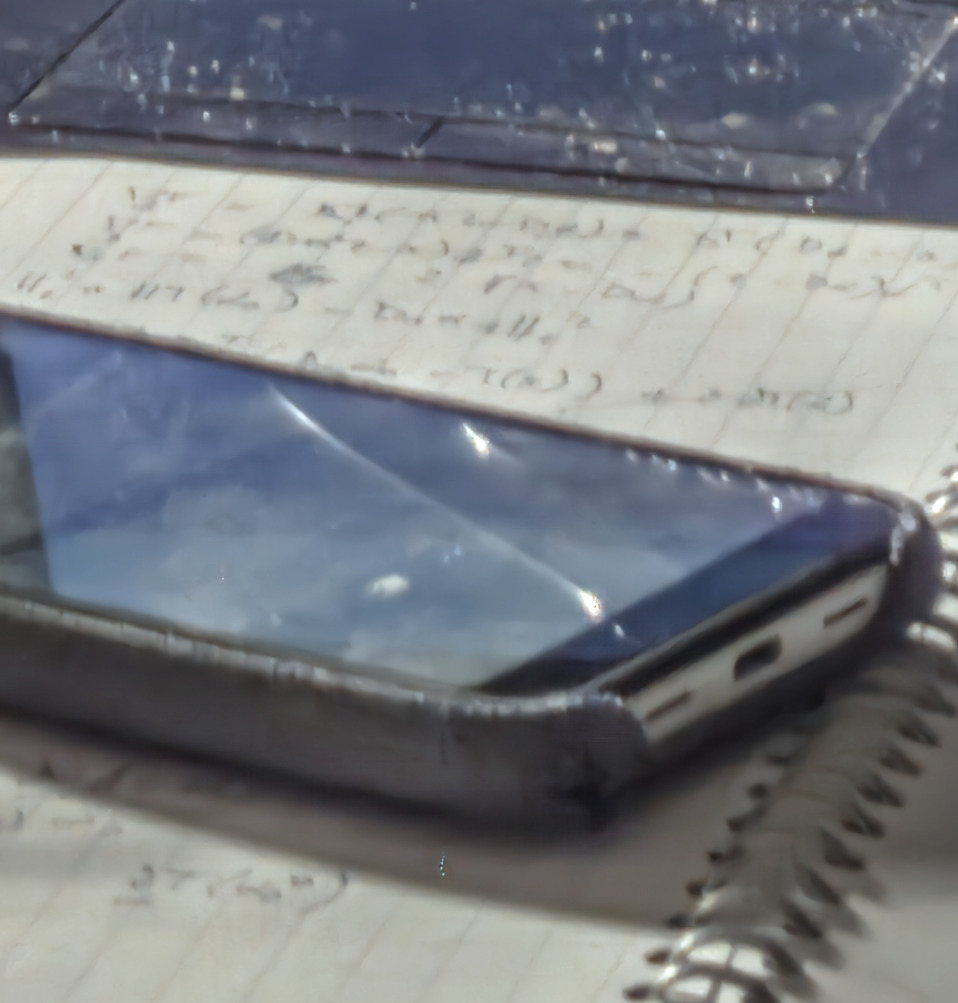}   & 
    \includegraphics[trim=0 80 0 40,clip,cfbox=red 0pt 0pt,width=0.16\textwidth]{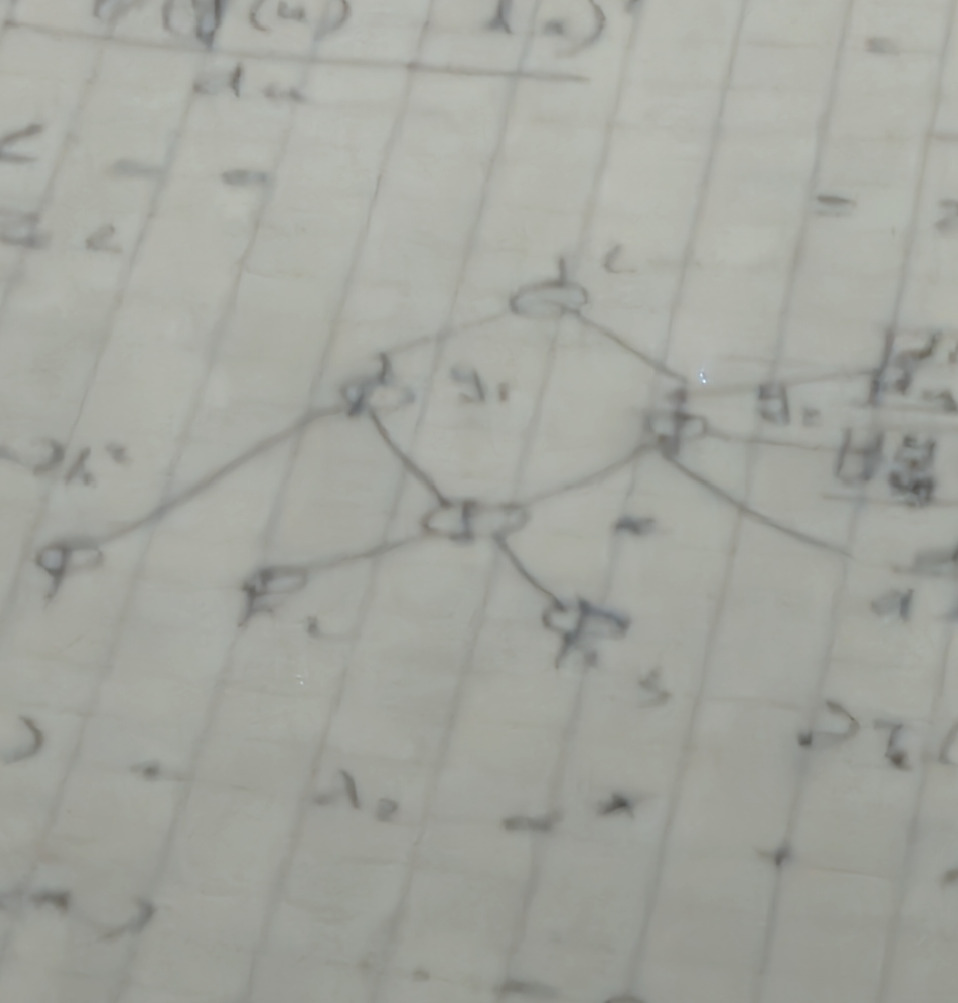}   & 
    \includegraphics[trim=0 80 0 40,clip,cfbox=red 0pt 0pt,width=0.16\textwidth]{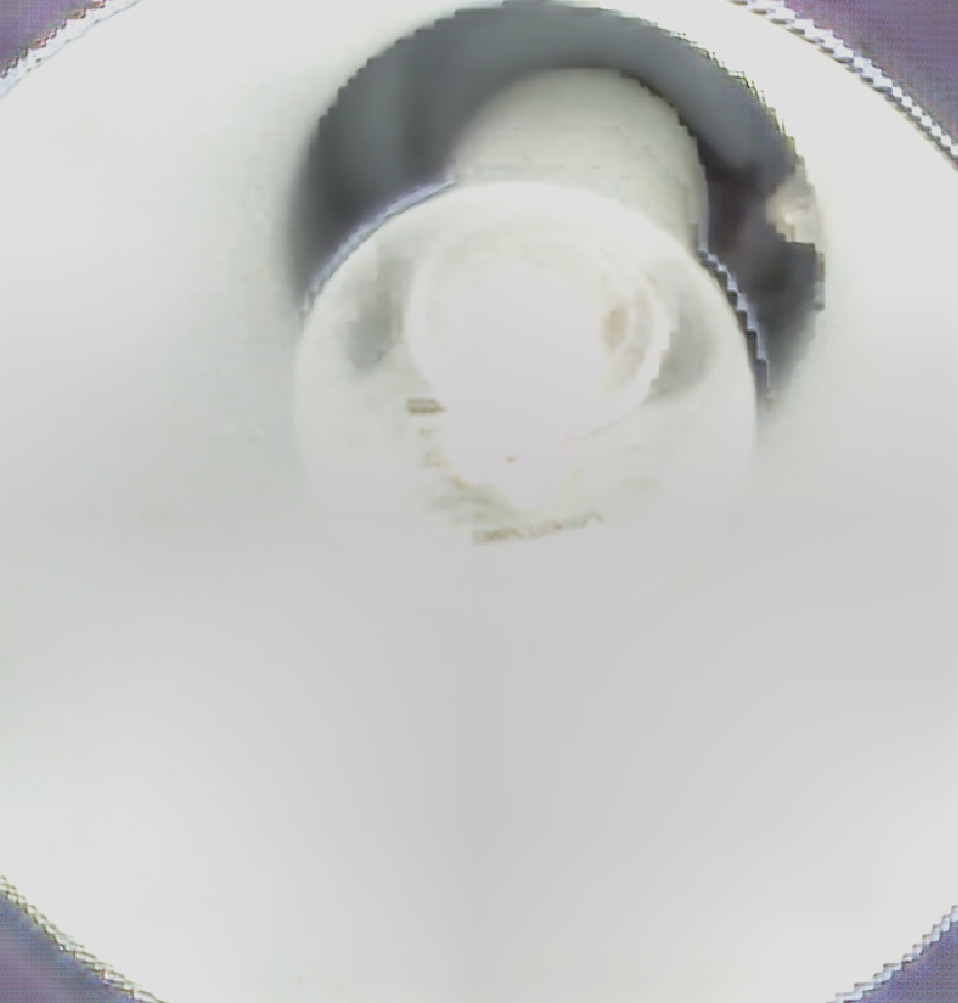}   & 
    \includegraphics[trim=0 80 0 40,clip,cfbox=red 0pt 0pt,width=0.16\textwidth]{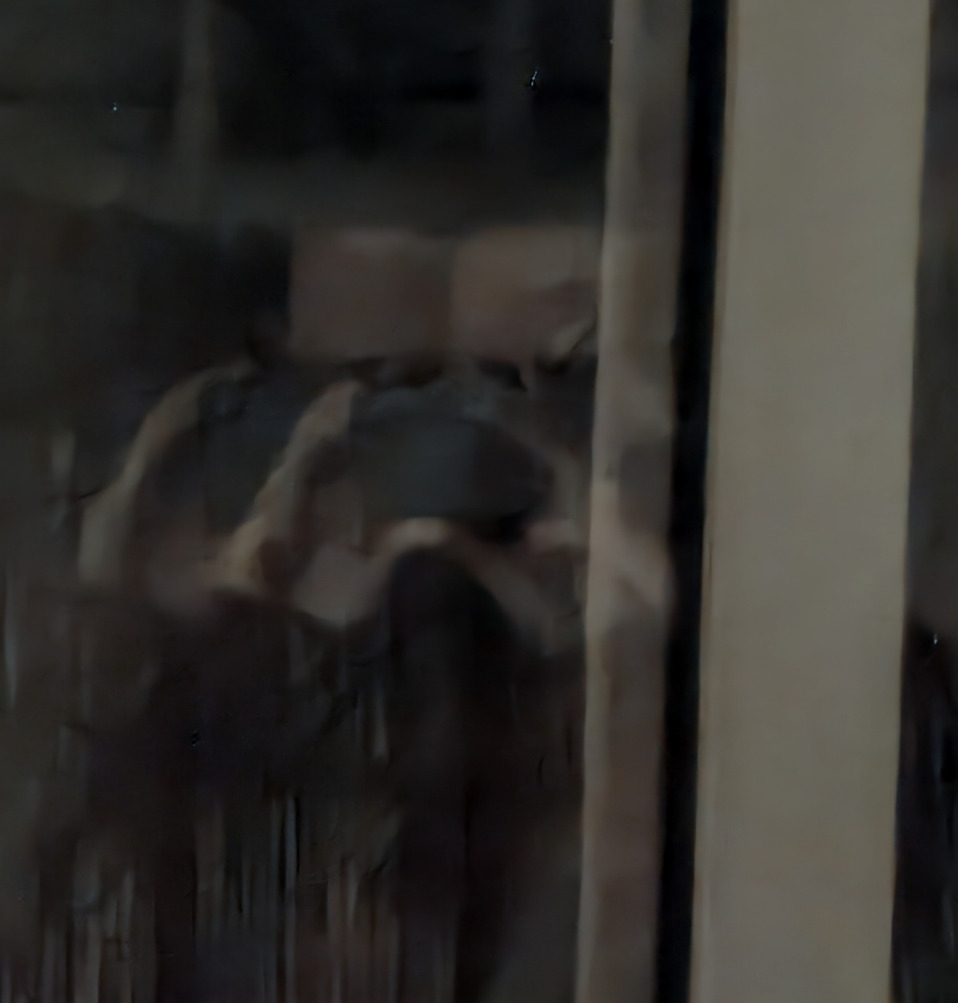}   & 
    \includegraphics[trim=0 80 0 40,clip,cfbox=red 0pt 0pt,width=0.16\textwidth]{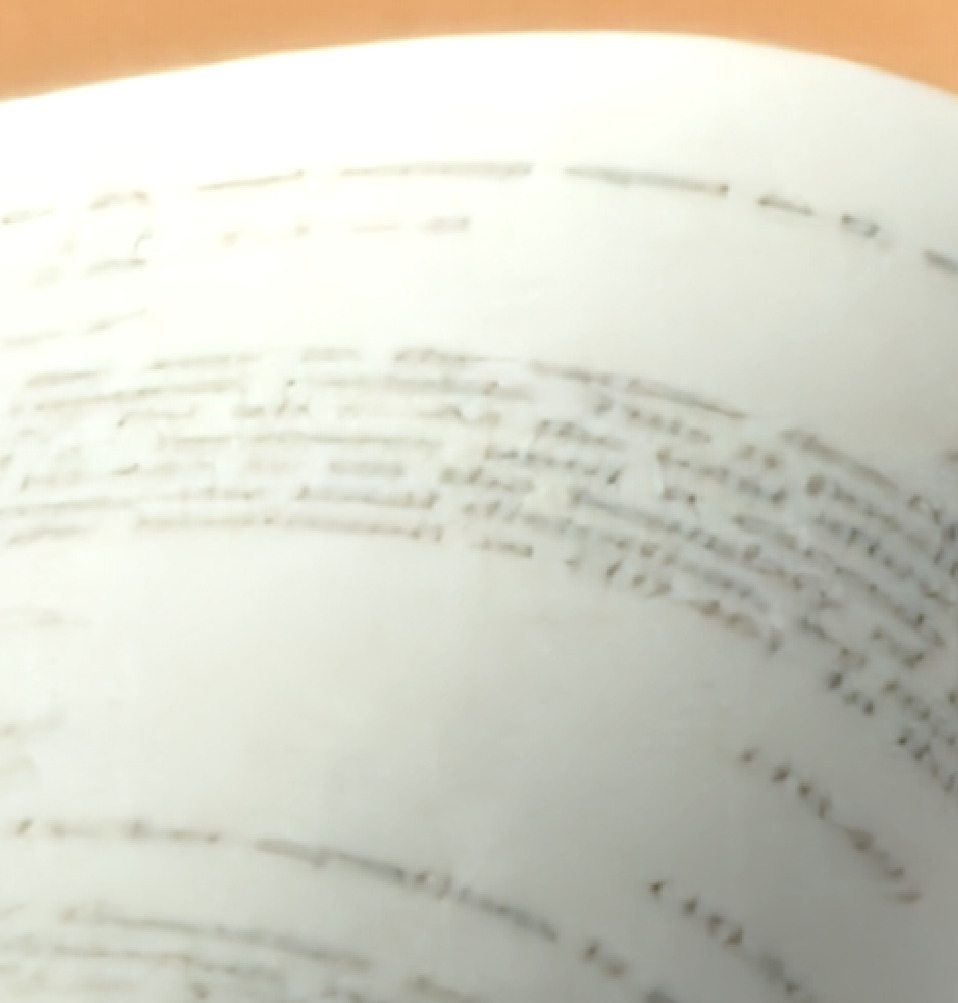}   \\
    \end{tabular}}
    \caption{
    Joint HDR imaging and super-resolution $\times 4$ with a burst taken with a hand-held Pixel4a at night, facing a spotlight. {\bf Top:} The original burst. {\bf Middle:} The central image in the burst (left) and the reconstructed HDR/SR image after tone mapping (right). {\bf Bottom:} Six crops showing details of the original and HDR/SR images, presented respectively in the first and second rows.
    }
    \label{fig:lowlighqualitative1}
    \end{figure*}

\section{Proposed approach}
The goal of this work is to design a  function
$F_\theta$ with learnable parameter $\theta$
which, given $K$ raw images 
$Y = \left\{y_1,\dots,y_K\right\}$ and corresponding exposure times
$\Delta = \{\Dt_1,\dots,\Dt_K\}$,
predicts a single-precision floating-point estimate $\hat{x}$ of the 
the HR $sh \times sw\times3$ irradiance map:
\begin{equation}\label{eq:learntfunction}
    \hat{x} = F_\theta(Y,\Delta).
\end{equation}
As explained later in this section, all images of the burst are automatically aligned on a reference frame $y_{k_0}$ (typically the central one that has in general a reasonable exposure).

\comment{Since the HDR image $\wx$ has finer precision than the images of the burst
encoded on $M=12$ or 14 bits and should have little read noise in the shadows,
it}
\subsection{Formulation of the problem}
\paragraph{Inverse problem.} Our image formation model~(\ref{eq:formatinmodel}) suggests using an inverse problem formulation to the design of $F_\theta$ and the recovery of $\wx$. We first convert the discrete raw pixel values from $y_k$ in $P_q$ into $32$-bits real values in $[0,1]$, and construct 
the binary mask $m(y_k)$ representing saturated pixels containing non-informative values. With an abuse of notation, we keep the notation $y_k$ for the floating-point burst images in the rest of this presentation, and formulate the solution of
our inverse problem as the joint recovery of the warp operators $W_1,\ldots,W_K$ (parameterized with a piecewise-affine model, as detailed later), and the irradiance image~$x$:
\begin{equation}\label{eq:pboriginal}
    \min_{x,W_1,\dots,W_K} \frac{1}{2} \sum_{k=1}^K \| {w_k} \odot( y_k - A_k x ) \|_F^2 + \lambda \Omega(x),  
\end{equation}
where $A_k$ is the image formation operator
defined in the previous section, $\odot$ denotes pointwise multiplication, and the function~$\Omega$ is a regularizer, and it
will be discussed in details later. The $h \times w$ maps~$w_k$
store pixel-wise weights used to control the relative contribution of each frame to the reconstruction of each pixel, a key factor for robustness in bracketing methods~\cite{granados10optimal, aguerrebere14best}.

\paragraph{A robust weighting strategy.}
We write
 \begin{equation}\label{eq:LSweights}
     w_k = \frac{\Delta t_k m(y_k)}{\sum_{j=1}^{K}  \Delta t_j m(y_j)} \odot g\left (y_k, W_k y_{1} \right ),
 \end{equation}
where $m(y_k)$ is the binary with zero values at saturated
 pixels (this formulation assumes the existence of non-saturated
 pixels at corresponding locations in the burst; when all pixels are saturated, we use uniform weights instead). Here, 
 the function $g$ is a confidence factor, often used in HDR imaging to weight down images incorrectly aligned~\cite{tursun16deghosting} and avoid ghosting effects. It can be handcrafted from classical image features and/or
priors, but we will instead follow a plug-and-play strategy (detailed in the next section) to directly learn a parametric function $g$ from supervisory data.
Our overall weighting strategy is useful for HDR since it provides larger weights to frames obtained with longer exposure time that are less noisy, but it also accounts for registration errors through the learned function~$g$, which turns out to be critical for robustness to moderate scene motion.

\paragraph{Warp parameterization.}
We align images with piecewise-affine warps $W_k=W_{p_k}$,
where $W_{k_0}$ is the identity and  $p = \{p_1,\dots,p_K\}$
is the set of warp parameters. This is implemented by
tiling the images into small (\eg $200 \times 200$) crops,
that are aligned independently with affine transformations with
6 parameters. 

\paragraph{Regularizer.}
Many classical regularizers can be used in the formulation of inverse problems in image processing applications, for example
sparse total-variation priors~\cite{choi09high} or
combinations of penalty functions computed from pixel or histogram values~\cite{pulli14flexisp,rad07multidimensional,debevec97recovering}.
We instead follow the same plug-and-play strategy as for the confidence function $g$, and learn a CNN in
place of the proximal operator~\cite{parikh14proximal} 
of the penalty function $\Omega$. We detail its implementation in Sec.~\ref{sec:learn}.

\subsection{Optimization strategy}
We solve our optimization problem with half-quadratic splitting (or HQS)~\cite{geman95hqs} 
by introducing an auxiliary 
variable $z$ and minimize
\begin{equation}
    \min_{x,z,p} \frac{1}{2} \sum_{k=1}^K \| w_k \odot\left( y_k - A_k  z \right) \|_F^2 + \frac{\eta}{2} \|x - z \|_F^2 +\lambda \Omega(x). \label{eq:hqs} 
\end{equation}
The parameter $\eta$ is usually increased at each iteration according to some preset schedule, which
guarantees that,  as $\eta$ grows,
the solution of this relaxed problem converges to that of the original one~\eqref{eq:pboriginal}~\cite{geman95hqs}. As detailed in Sec.~\ref{sec:learn}, we choose
instead to learn this parameter from training data,
which improves performance in practice.
Note that we now find the warp
operators by minimizing the energy with respect to the warp parameters $p$, and that all operators involved are implemented efficiently
by exploiting the image structure (\eg convolutions instead of
large sparse operators, etc.).
The optimization is carried out by first initializing 
$z$ and $p$, then, in an alternate fashion, repeating~$T$ times ($T=3$ in our implementation) an HQS stage consisting of the three steps detailed below. The motivation for this strategy is that it allows us to gracefully convert our optimization method into a trainable architecture, as discussed in Sec.~\ref{sec:learn}, thanks to automatic differentiation tools~\citep{baydin2018automatic} implemented in modern deep learning frameworks.

\paragraph{Updating $z$.}
The auxiliary image $z$ is updated by a few steps
of a simple gradient descent (GD) algorithm:
\begin{equation}\label{eq:weightedLS}
    z \leftarrow z - \delta \left(\eta(z - x) + \sum_{k=1}^K A_k^\top \left(w_k^2 \odot (A_k z - y_k)\right)    \right),
\end{equation}
where $\delta$ is a step size (which will be learned automatically by the procedure presented in the next section), and of course $A_k$ depends on the current warping parameters $p$.

\paragraph{Updating $x$.}
Minimizing~(\ref{eq:hqs}) with respect to the image $x$ while keeping the other variables fixed amounts to compute the so-called proximal
operator $G$ of $\Omega$~\cite{parikh14proximal}:
\begin{equation}\label{eq:proximal}
    x = G(z,\lambda/\eta) = \argmin_{x} \frac{1}{2}\|x-z\|_{\text{F}}^2 + \frac{\lambda}{\eta} \Omega(x).
\end{equation}
We will detail in the next section how we implement $G$.

\paragraph{Updating $p$.}
\citet{lecouat21aliasing} estimate the warp 
parameters $p_k$ ($k\neq k_0$) on 200$\times$200 
tiles in a 4-scale Gaussian image pyramid, running
three stages of the Lucas-Kanade algorithm~\citep{lucas81iterative} at each stage. We will show in Sec.~\ref{sec:learn} how to
do significantly better, both quantitatively and qualitatively, by using a similar approach to align {\em learned} features
instead. 

\paragraph{Initialization of $p$ and $z$.}
A fast and coarse initialization of the warp parameters $p$ is 
obtained using a sub-pixel variant of the FFT-based
algorithm of~\cite{anuta70phase} with the features of~\cite{ward03fast}.
After having estimated $p$ for the first time with the Lucas-Kanade algorithm and
before the first $z$-update stage, we initialize $z$ as follows: 
we demosaick each frame $y_k$ with bilinear 
interpolation, align them with the warping operators $W_k$, average them with
the normalized weights $\Delta_k/\sum_{j=1}^K\Delta_j$, and finally upscale the resulting image
by a factor $s$ with bilinear interpolation.
This procedure yields a fast and coarse estimate of the HR and HDR image
to start the GD algorithm in Eq.~\eqref{eq:weightedLS}.

\subsection{Learnable architecture\label{sec:learn}}
The optimization procedure described in the previous section is
implemented as a function $F_\theta$ that produces an estimate $\hat{x}$ from a burst $Y$ and exposure times $\Delta$, according to Eq.~(\ref{eq:learntfunction}). By writing this function as a finite sequence of
operations that are differentiable with respect to the model parameters~$\theta$, it is then possible to leverage training data---that is, pairs of HR/HDR images~$x$ associated to LR/LDR bursts---to \emph{learn} these parameters for the reconstruction task. This of course raises questions about data collection and generation, which are discussed later, but it also opens up many possibilities for further improvements. In particular, as described in the rest of
this section, this allows us to learn implicitly the regularization function~$\Omega$ by taking advantage of deep learning principles, as well
as learning appropriate weighting strategies, and robust features to improve image alignment.

\paragraph{Learnable proximal operator~$G$.}
Following the {\em plug-and-play} strategy~\cite{plugandplay} which has proven powerful in the signal processing literature, we
replace the proximal operator~$G$ above by a function $G_\omega$ represented by a CNN and parameterized by~$\omega$, such that the update~(\ref{eq:proximal}) becomes
\begin{equation}\label{eq:learntproximal}
    x = G_\omega(z, \gamma),
\end{equation}
where $\gamma$ is also a trainable parameter.
The CNN has a residual U-net architecture, which is a smaller variant of the network of~\citet{zhang2020deep} for single-image super resolution. 
This network has four scales with respectively 32,64,128,128 channels per scale.  We also run experiments with an even smaller version of the network with 32 features per channel (dubbed {\em small}) and 16 features per channel (dubbed {\em tiny}). 
Note that for our problem, the first layer has 4 input channels: three for the predicted RGB auxiliary variable $z$ and 
one for the scalar~$\gamma$.

\paragraph{Learnable confidence function~$g$.}
Similarly, since designing the function $g$ by hand is difficult, we choose to learn instead a CNN $g_\rho$, and the fusion weights $w_k$ become for all $k\neq k_0$:
\begin{equation}\label{eq:learntLSweights}
    w_k = \frac{\Delta t_k m(y_k, c)}
    {\sum_{j=1}^K \Delta t_j m(y_j, c)} \odot g_\rho(y_k, W_k y_{k_0}), 
\end{equation}
The function $g_\rho$ is implemented with the tiny variant of the U-Net architecture used above. The network takes as input the concatenation along the channel dimension of RGB versions of the images $y_k$ and $W_k y_{k_0}$ obtained by bilinear interpolation.

\paragraph{Learnable features for alignment.}
A classical approach to the registration of frame captured with different exposure times is to use MTB features~\cite{ward03fast}. Here, we construct instead a single-channel feature map for each raw image using again the tiny CNN with U-net architecture, then perform the multi-scale Lucas Kanade algorithm for a fixed number of iterations (3 iteration per scale of the pyramid) {\em directly on the feature map}. Our implementation of the forward additive version of the Lucas Kanade algorithm is fully differentiable. Therefore we can learn the parameters of the feature map jointly with all the trainable parameters of our model, following a strategy similar to~\cite{chang2017clkn}. As shown
in the experimental section this significantly improves registration performance.

\subsection{Learning the model parameters $\theta$}
We denote here by $\theta$ all the learnable parameters of our methods, including those of the CNNs and the scalar parameters involved in the HQS optimization procedure introduced above (\eg, $\delta$, $\eta$, \ldots). 
We use triplets of the form $(x^{(i)}, Y^{(i)}, \Delta^{(i)})$ ($i=1,\ldots,n$) of training data to supervise the learning
procedure. In our setting where ground-truth HDR/HR images are
normally not available for real image bursts, the training data
is necessarily semi-synthetic, that is, obtained by applying various
transformations to real images. Obtaining robust inference with real raw bursts is thus challenging. The hybrid nature of our algorithm, which exploits both a learning-free inverse problem formulation and data-driven priors, appears to be a key to achieving good generalization on real raw data acquired in various conditions that do not necessarily occur in the training dataset.

\begin{figure*}
    \setlength\tabcolsep{0.5pt}
    \renewcommand{\arraystretch}{0.5}
    \centering
    \scalebox{1.02}{
    \begin{tabular}{cc}
        \begin{tabular}{c}
            \includegraphics[width=0.305\textwidth,trim=0 0 0 100]{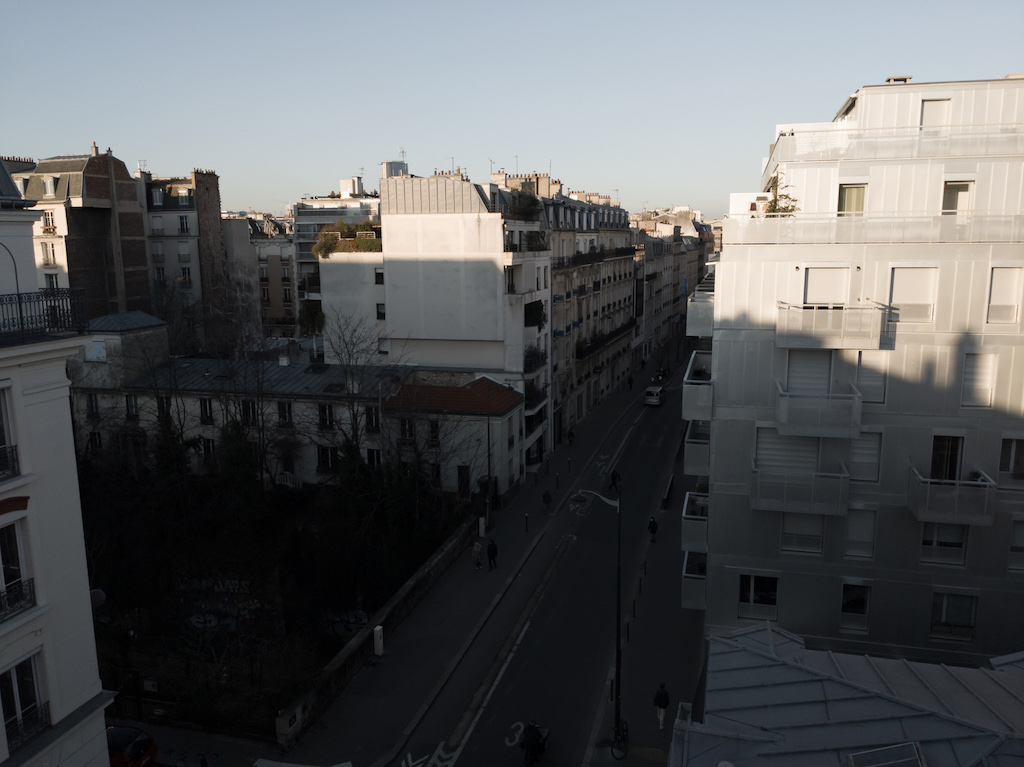}  \\ 
            \includegraphics[width=0.305\textwidth,trim=5 5 5 5, clip]{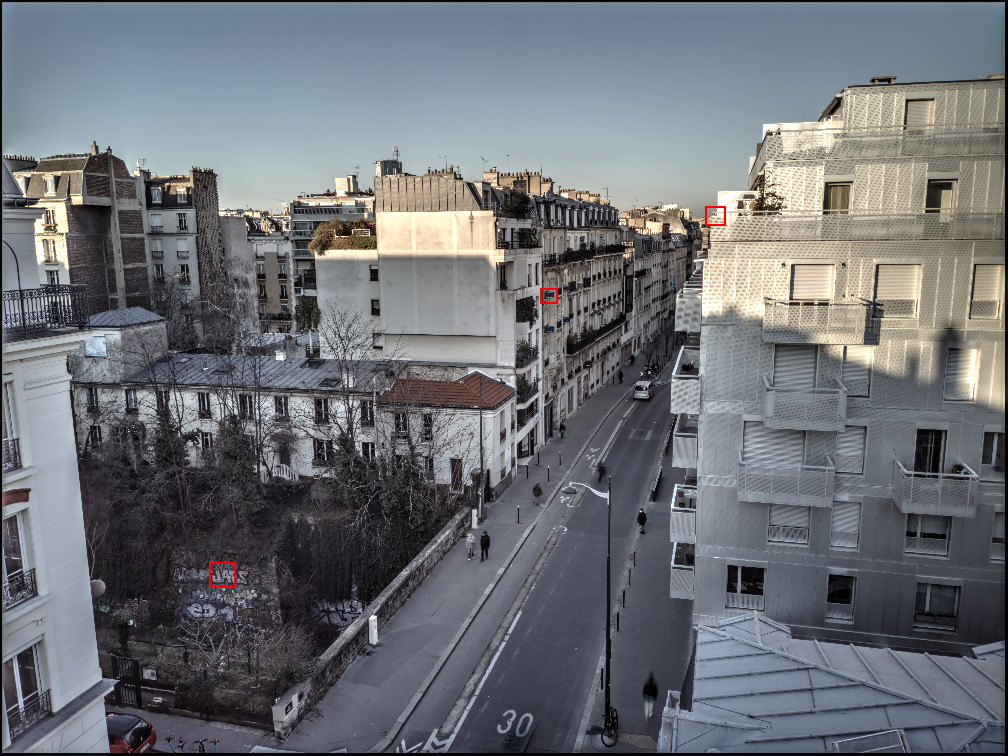}        \\
        
        \end{tabular}
        &
        \begin{tabular}{cccc}
            \includegraphics[trim=0 0 0 100,clip,cfbox=green 0pt 0pt,width=0.16\textwidth]{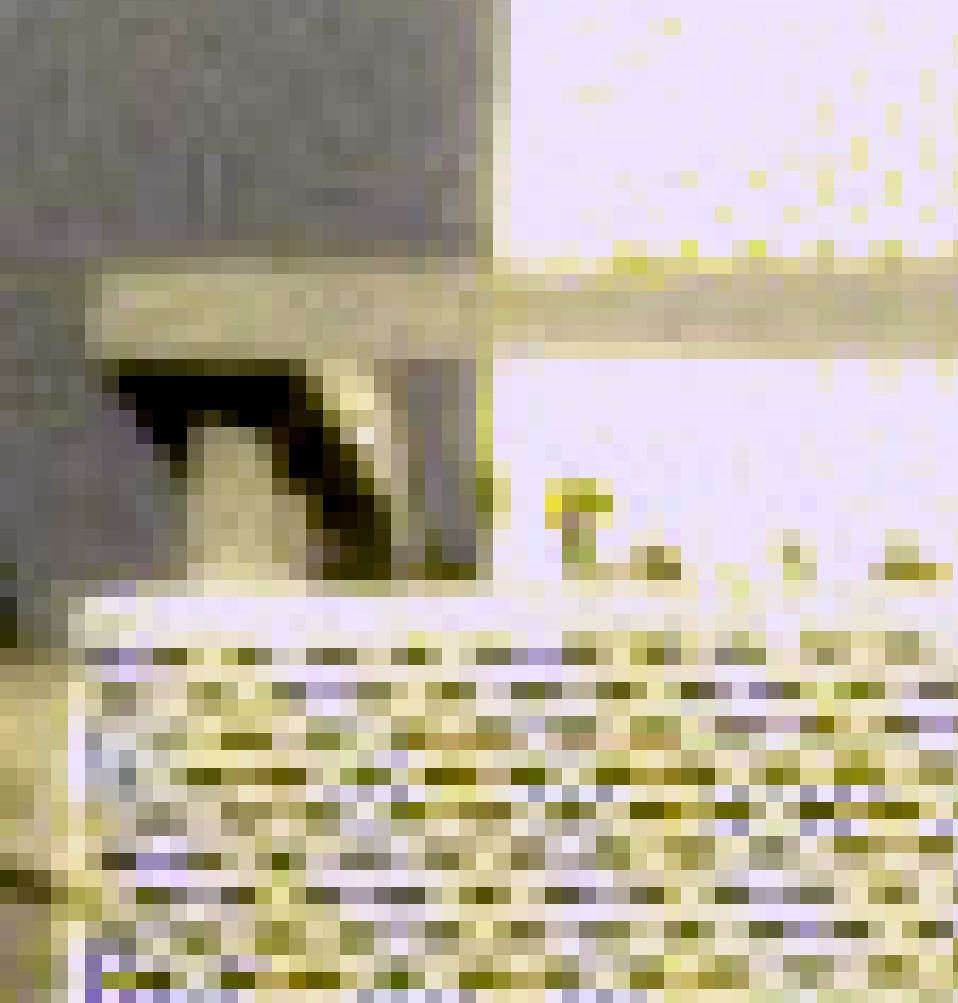}      & 
            \includegraphics[trim=0 0 0 100,clip,cfbox=green 0pt 0pt,width=0.16\textwidth]{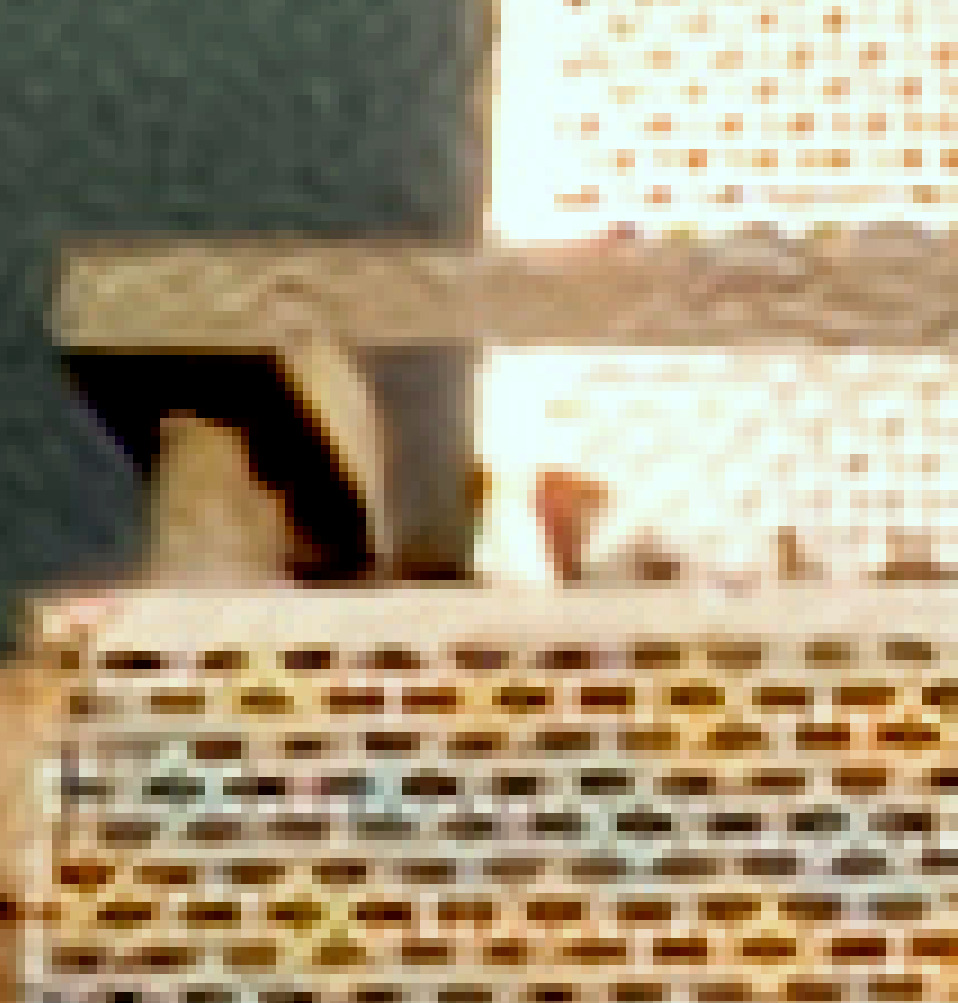}      & 
            \includegraphics[trim=0 0 0 100,clip,cfbox=green 0pt 0pt,width=0.16\textwidth]{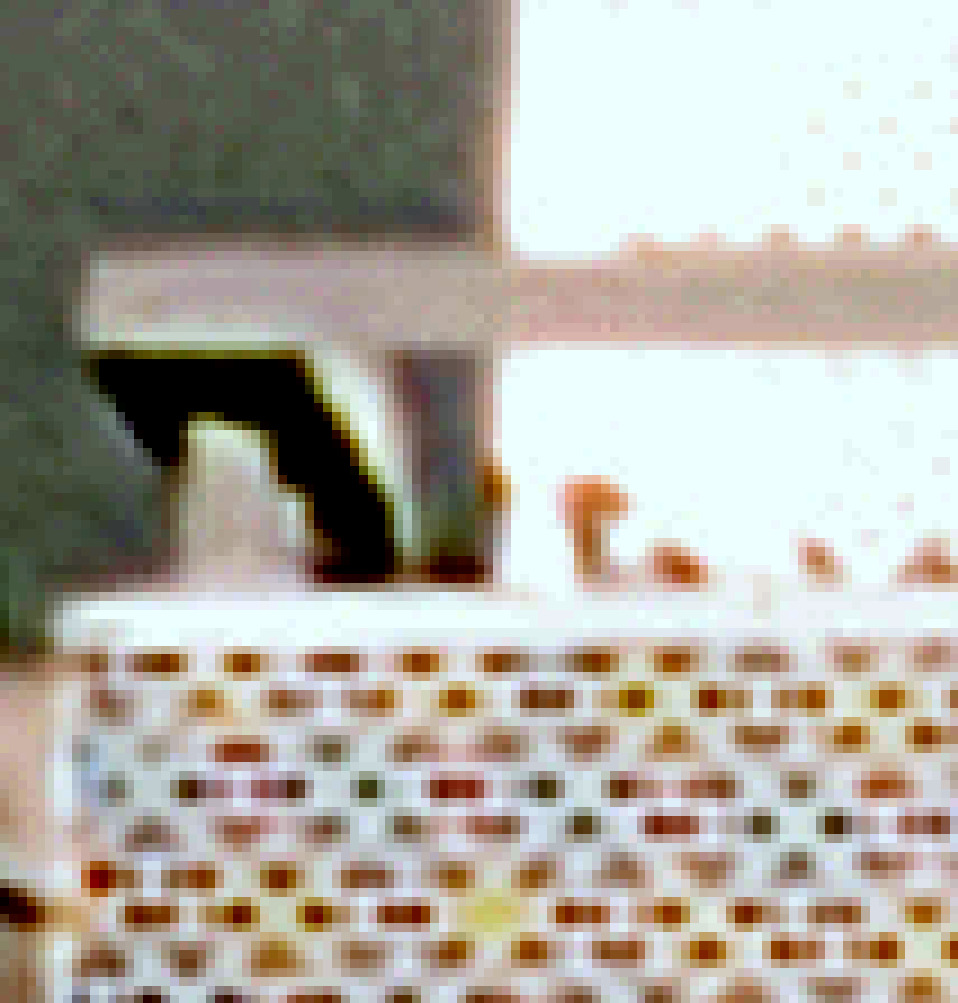}      & 
            \includegraphics[trim=0 0 0 100,clip,cfbox=green 0pt 0pt,width=0.16\textwidth]{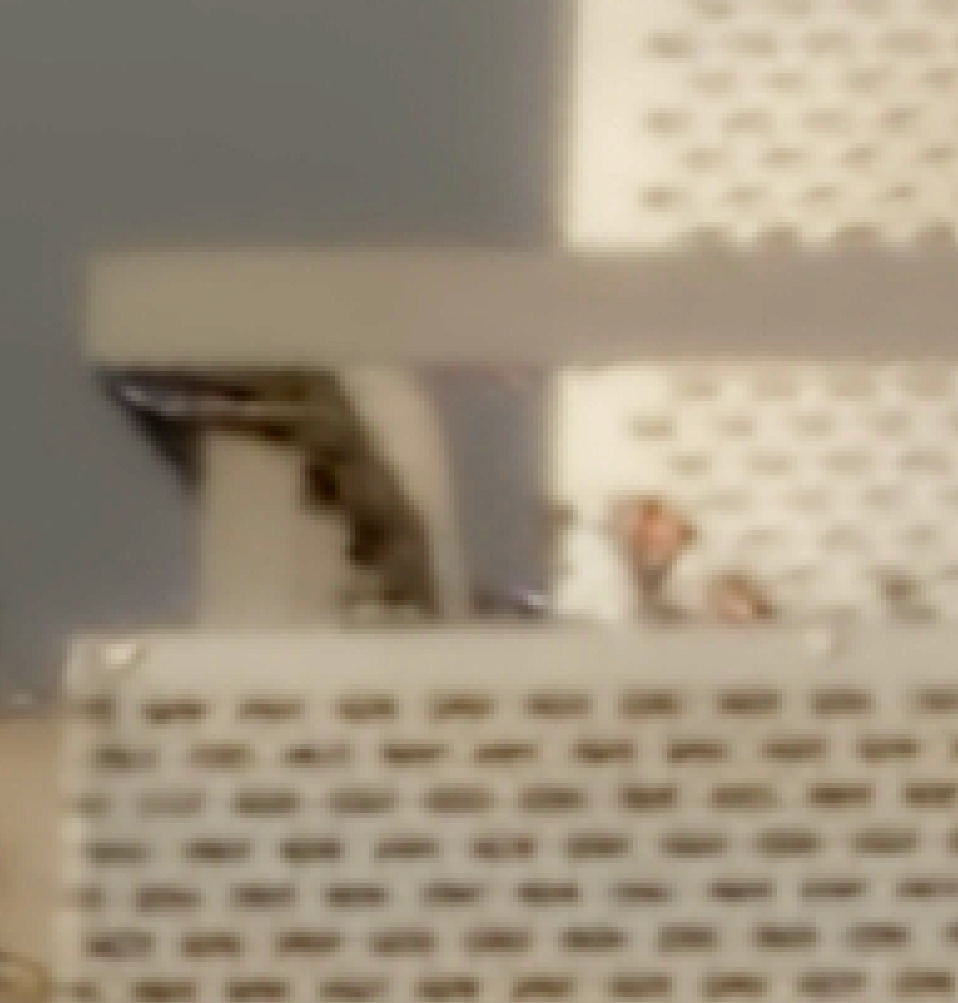}      \\ 
            \includegraphics[trim=0 0 0 100,clip,cfbox=red 0pt 0pt,width=0.16\textwidth]{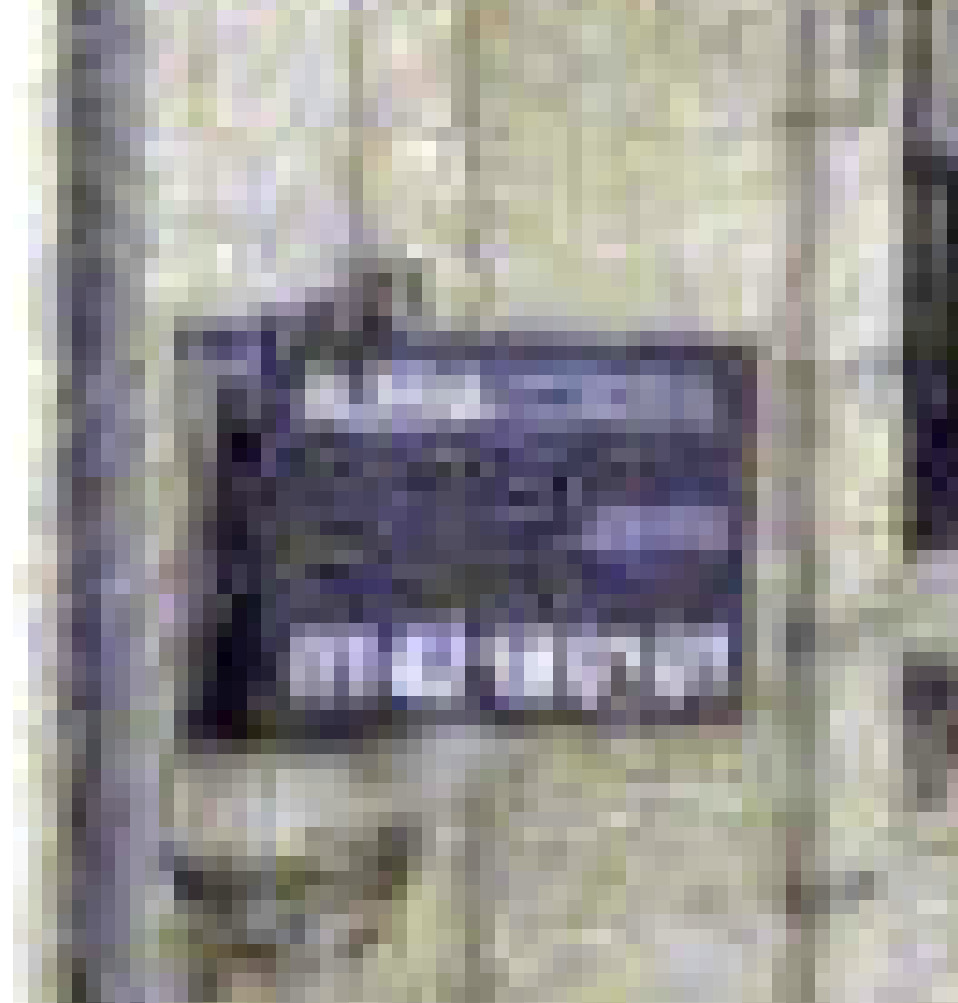}      & 
            \includegraphics[trim=28 28 0 100,clip,cfbox=red 0pt 0pt,width=0.16\textwidth]{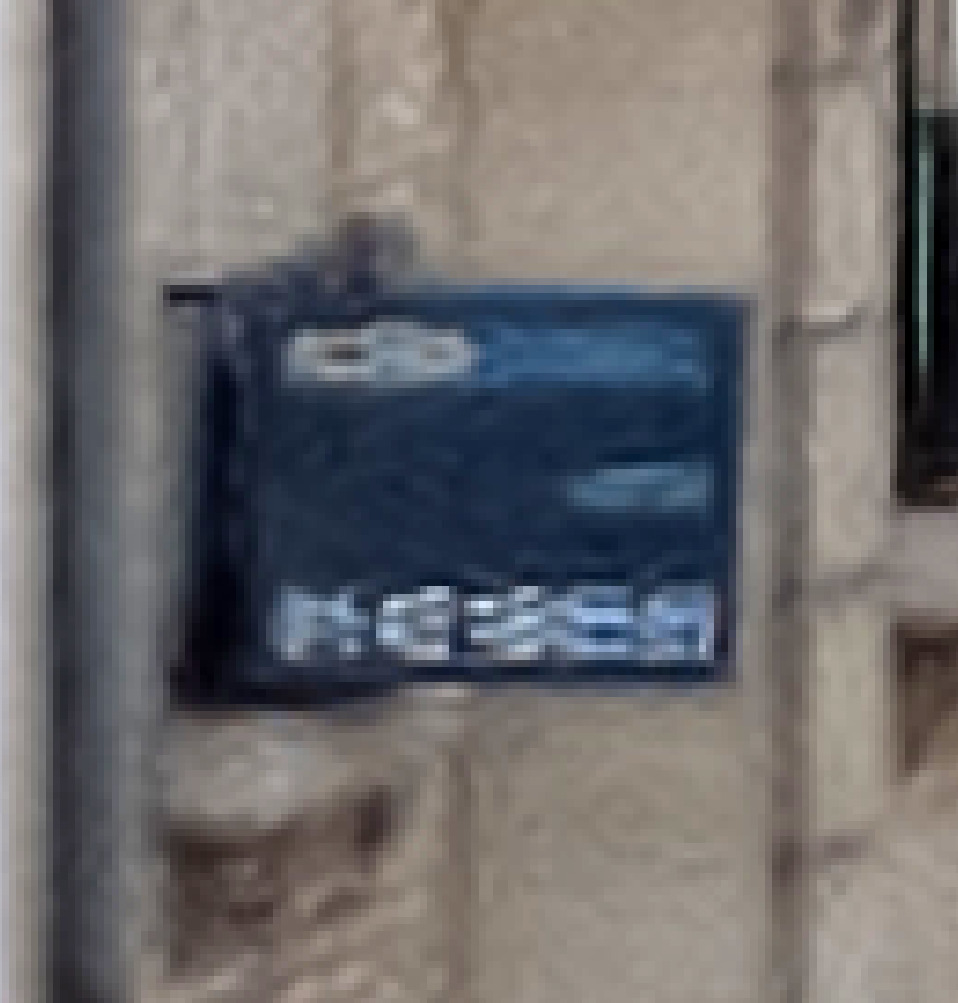}      & 
            \includegraphics[trim=0 0 0 100,clip,cfbox=red 0pt 0pt,width=0.16\textwidth]{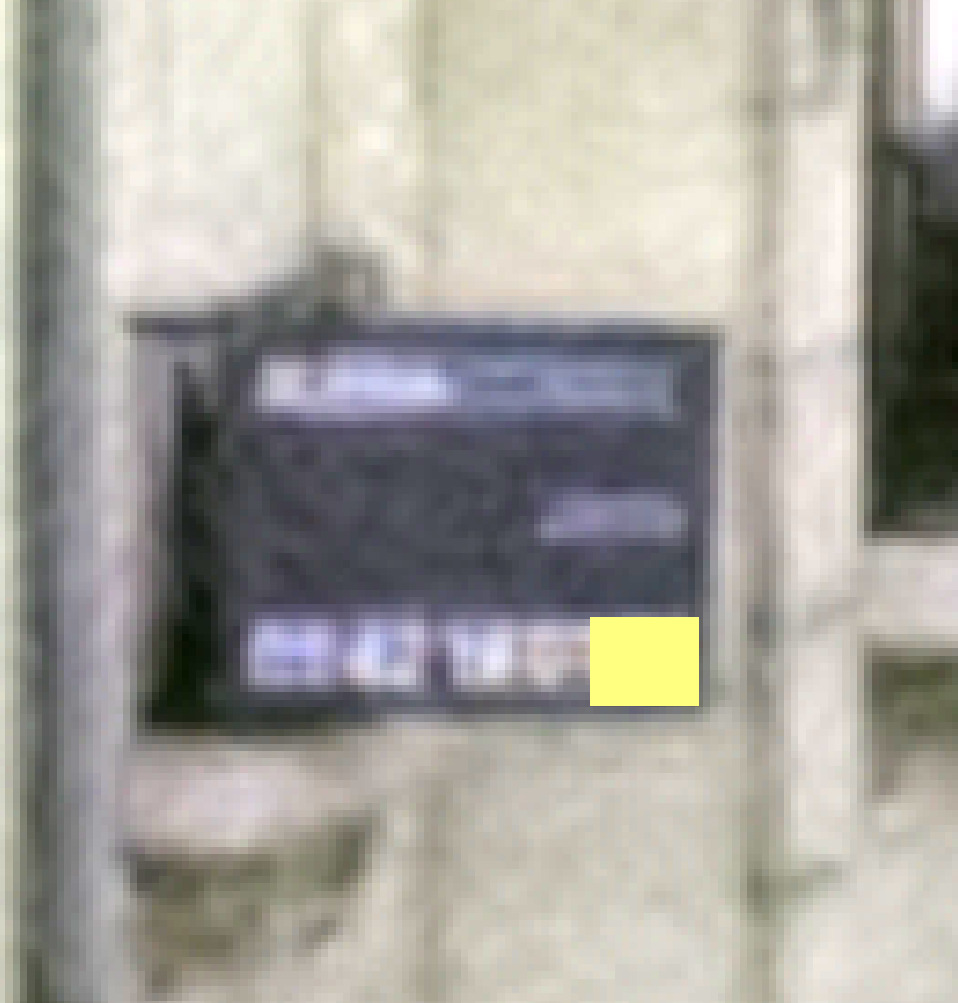}      & 
            \includegraphics[trim=0 0 0 100,clip,cfbox=red 0pt 0pt,width=0.16\textwidth]{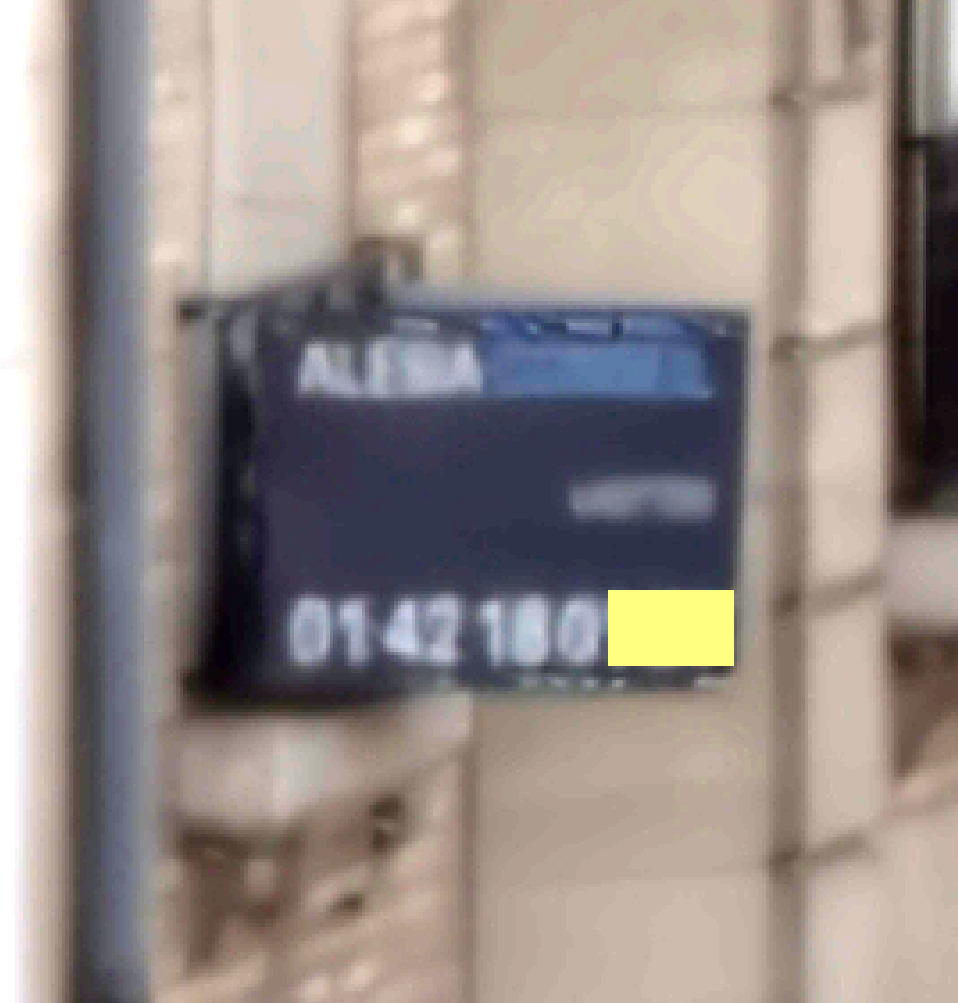}      \\ 
            \includegraphics[trim=0 0 0 100,clip,cfbox=blue 0pt 0pt,width=0.16\textwidth]{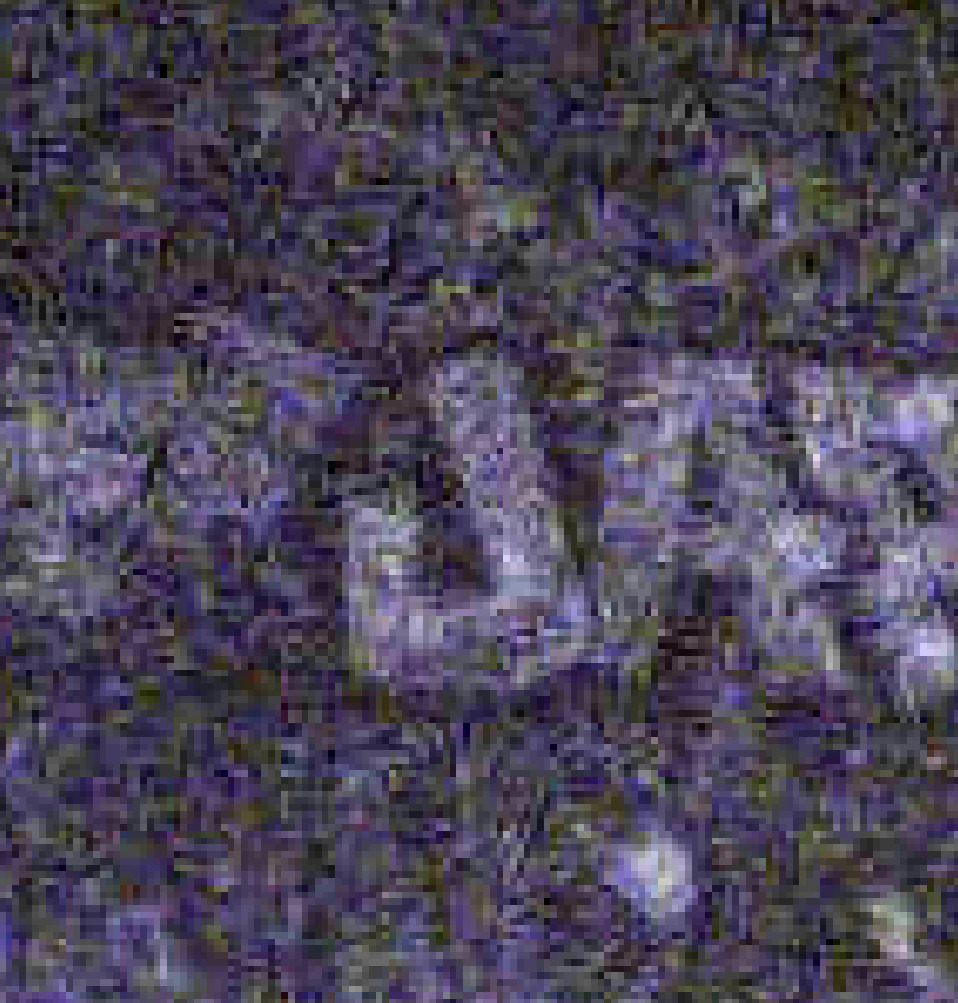}      & 
            \includegraphics[trim=0 0 0 100,clip,cfbox=blue 0pt 0pt,width=0.16\textwidth]{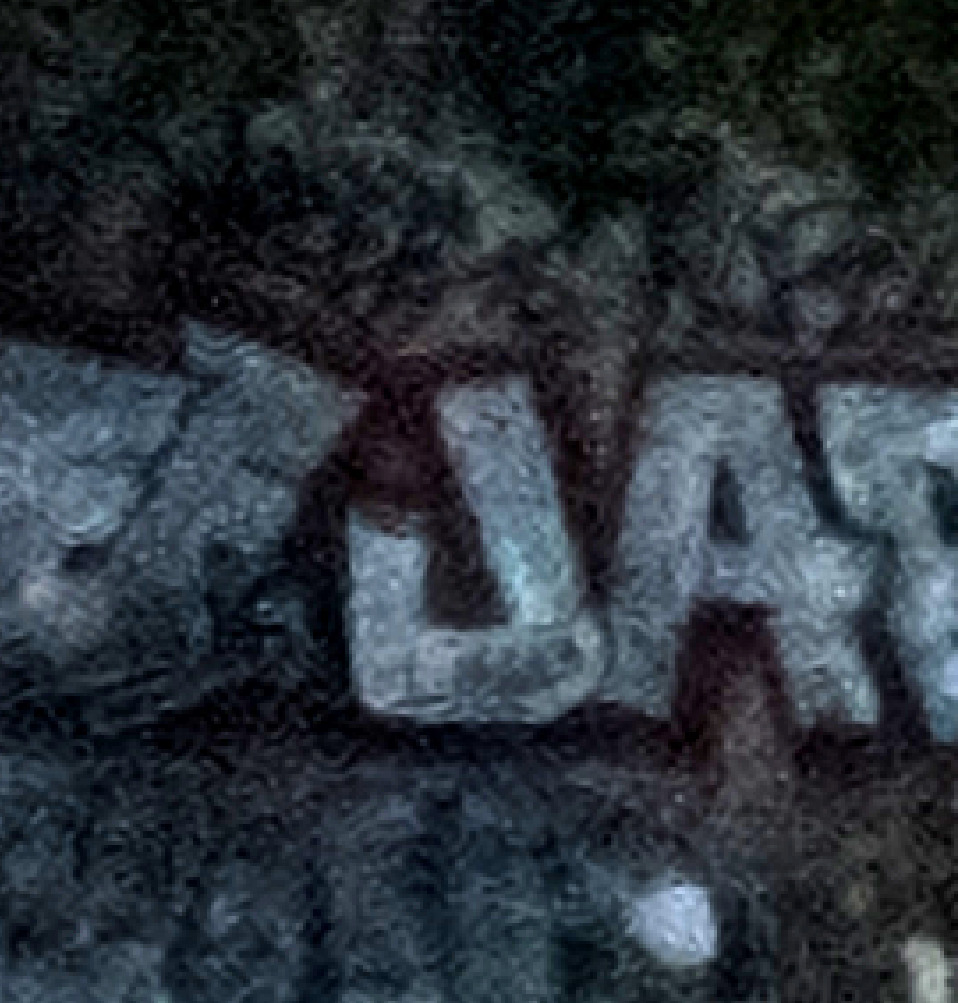}      & 
            \includegraphics[trim=0 0 0 100,clip,cfbox=blue 0pt 0pt,width=0.16\textwidth]{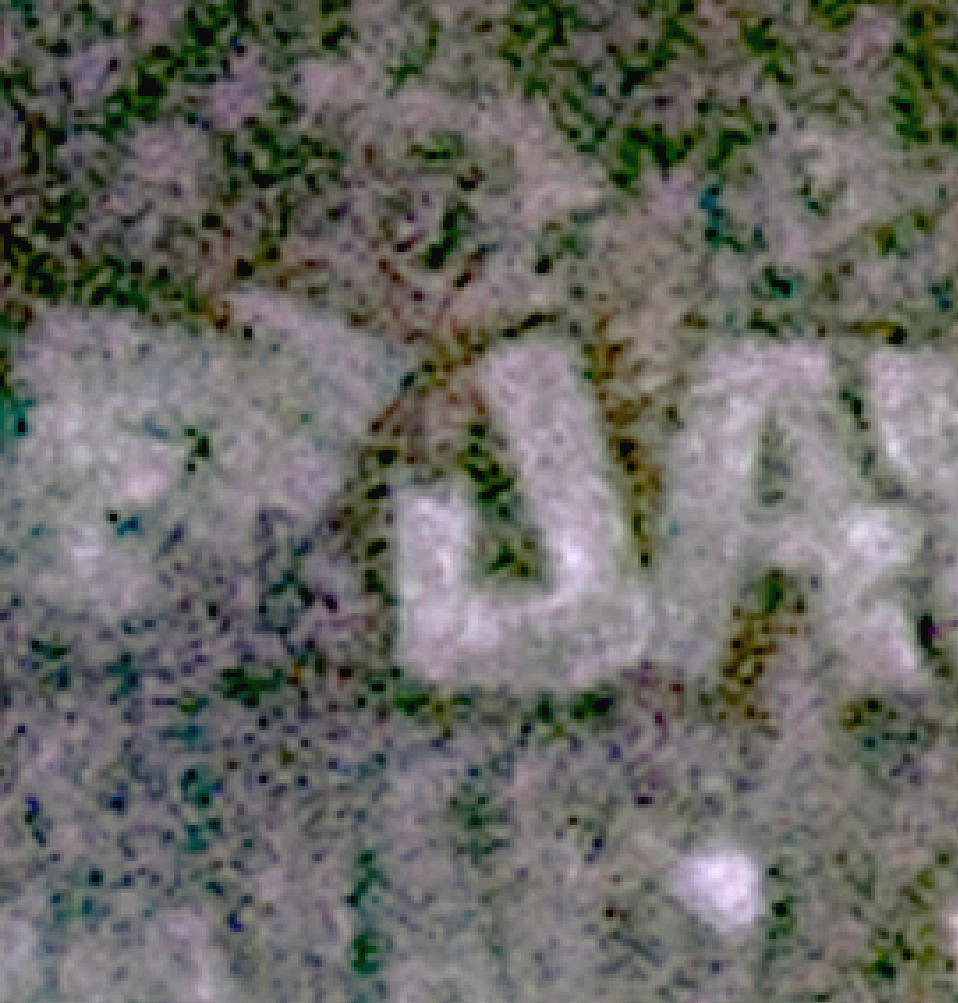}      & 
            \includegraphics[trim=0 0 0 100,clip,cfbox=blue 0pt 0pt,width=0.16\textwidth]{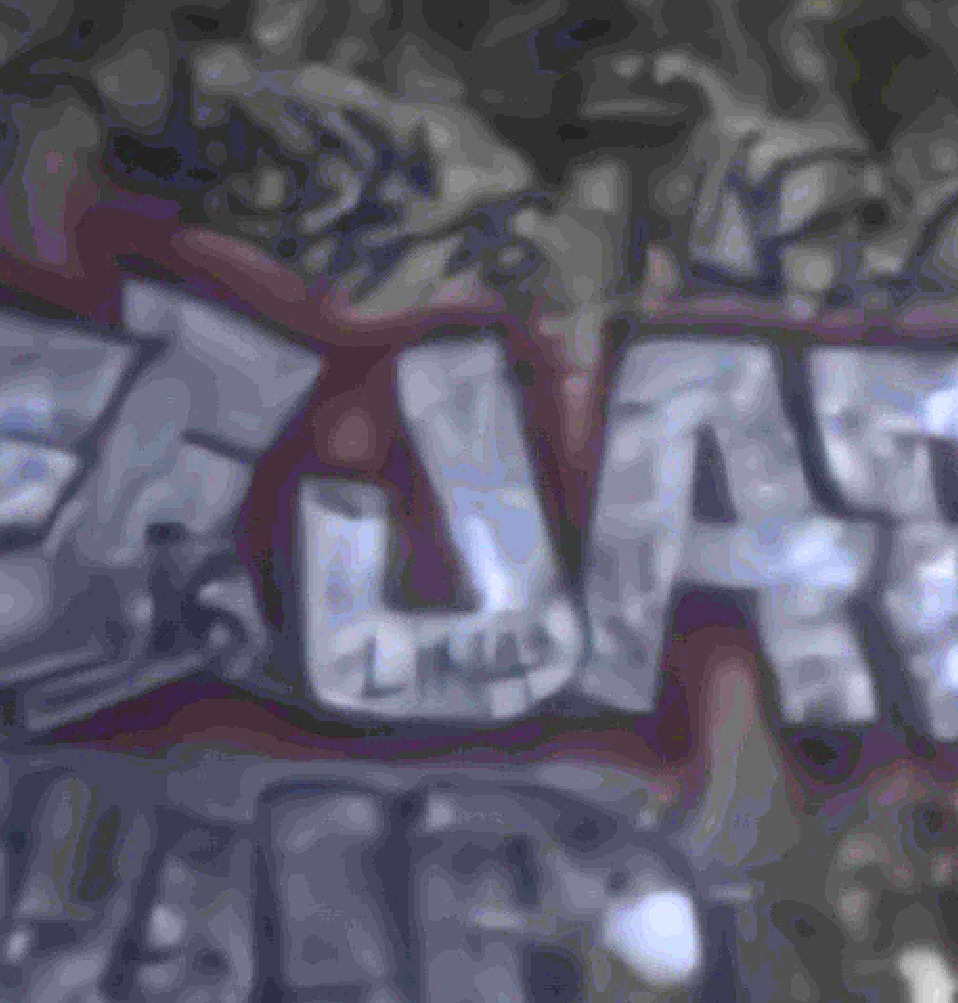}      \\
            \begin{tabular}{c} Low resolution \\ (dcraw) \end{tabular} &
            \begin{tabular}{c} ACR (x2) \\ (1 frame) \end{tabular} &
            \begin{tabular}{c} Kim et al (x2) \\ (1 frame)  \end{tabular} &
            \begin{tabular}{c} Ours (x4) \\ (18 frames) \end{tabular}
        \end{tabular}
   \end{tabular}
   }
    \caption{
    Day-time comparisons of joint HDR imaging and super-resolution algorithms with bursts acquired by a Pixel4a. {\bf Left:} The central image in the burst (top) and our reconstruction (bottom). {\bf Right:} Comparison of close-ups of the reconstructions obtained by the CNN-based Adobe Camera Raw single-image algorithm for $\times 2$ super-resolution and demosaicking, the CNN-based $\times 2$ super-resolution method of \cite{kim19deep}, and our method. (Note: part of the phone number legible in our case is masked for privacy reasons.)
    }
    \label{fig:dayqualitative2}  
\end{figure*}
 
\begin{figure*}
    \setlength\tabcolsep{0.5pt}
    \renewcommand{\arraystretch}{0.5}
    \centering
    \scalebox{1.05}{
    \begin{tabular}{cc}
        \begin{tabular}{c}
            \includegraphics[width=0.28\textwidth,trim=0 0 0 105]{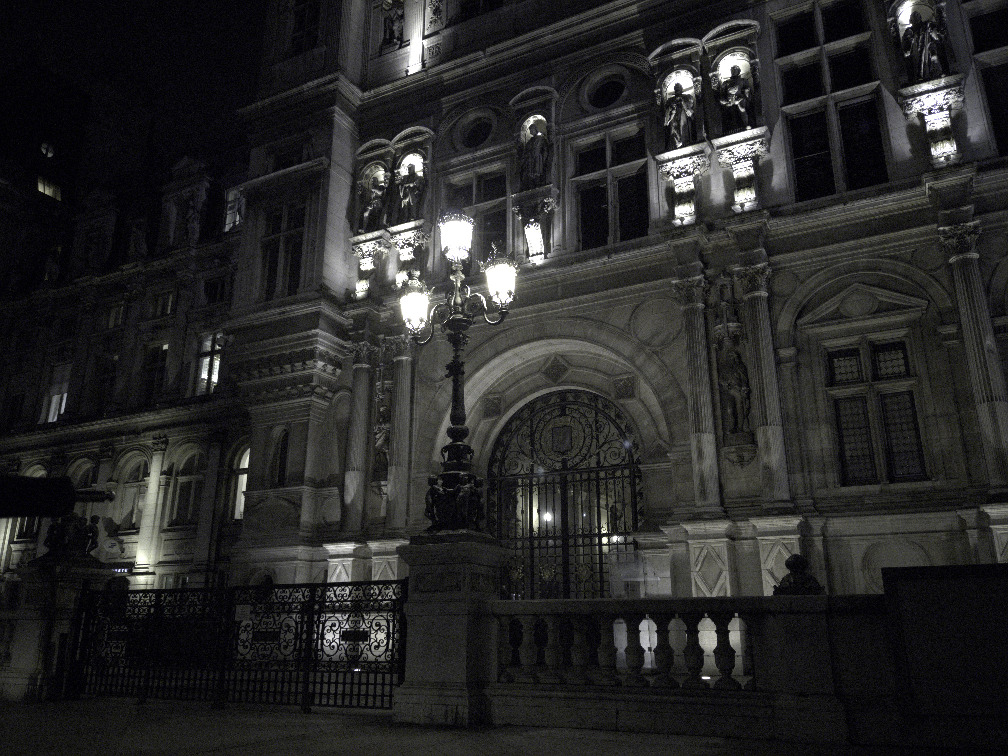}  \\ 
            \includegraphics[width=0.28\textwidth]{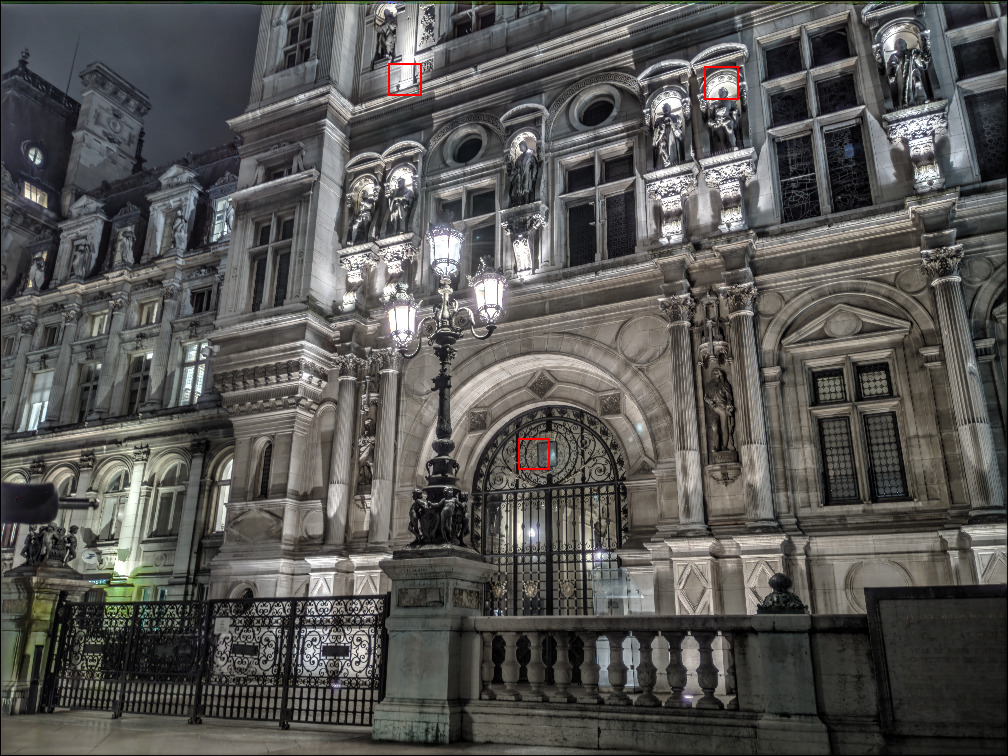}      \\
        \end{tabular}
        &
        \begin{tabular}{cccc}
            \includegraphics[trim=0 0 0 100,clip,cfbox=green 0pt 0pt,width=0.16\textwidth]{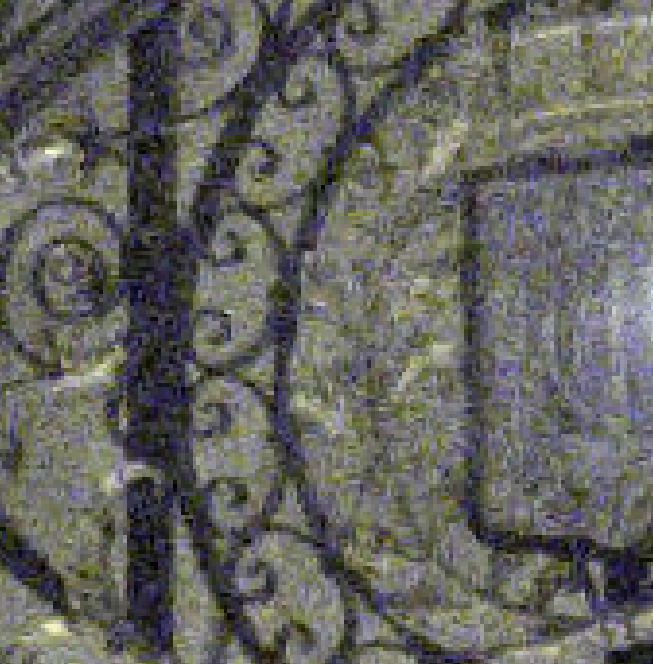}      & 
            \includegraphics[trim=0 0 0 100,clip,cfbox=green 0pt 0pt,width=0.16\textwidth]{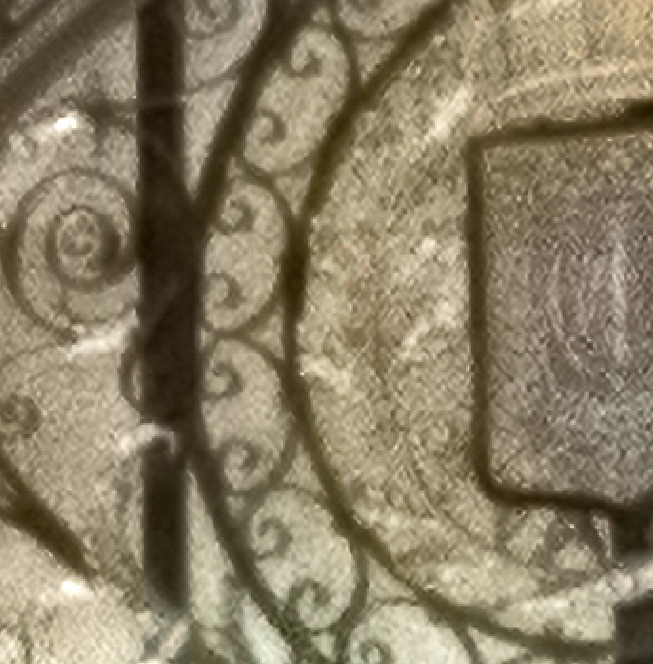}      & 
            \includegraphics[trim=0 0 0 100,clip,cfbox=green 0pt 0pt,width=0.16\textwidth]{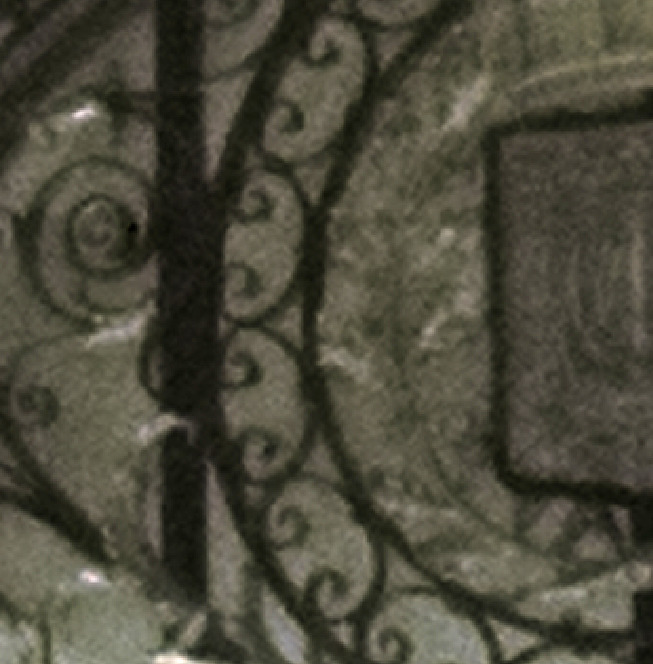}      & 
            \includegraphics[trim=0 0 0 100,clip,cfbox=green 0pt 0pt,width=0.16\textwidth]{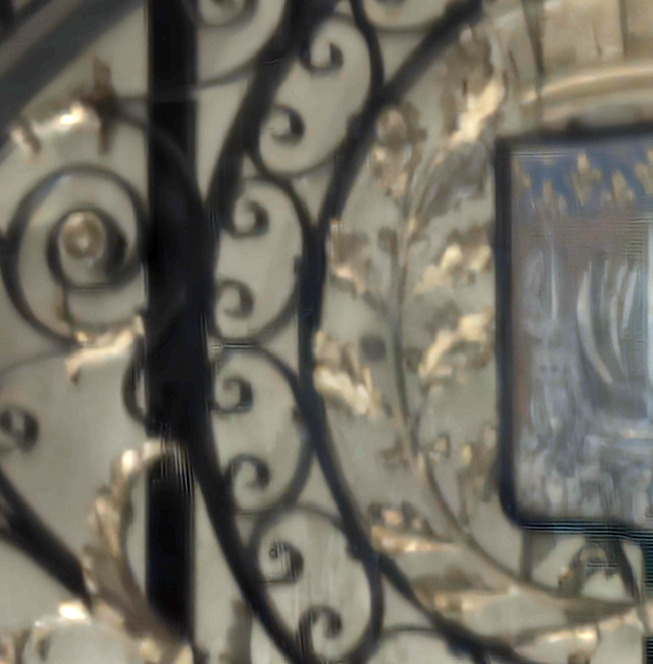}    \\ 
            
            \includegraphics[trim=0 0 0 100,clip,cfbox=red 0pt 0pt,width=0.16\textwidth]{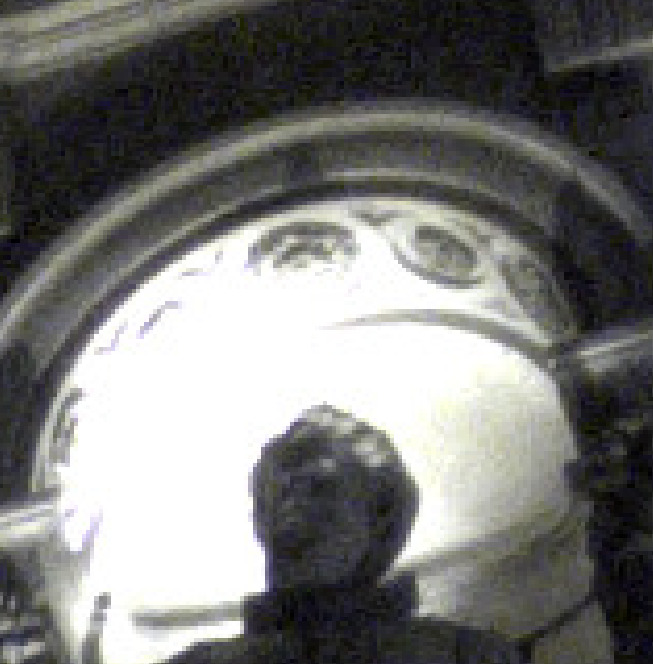}      & 
            \includegraphics[trim=0 0 0 100,clip,cfbox=red 0pt 0pt,width=0.16\textwidth]{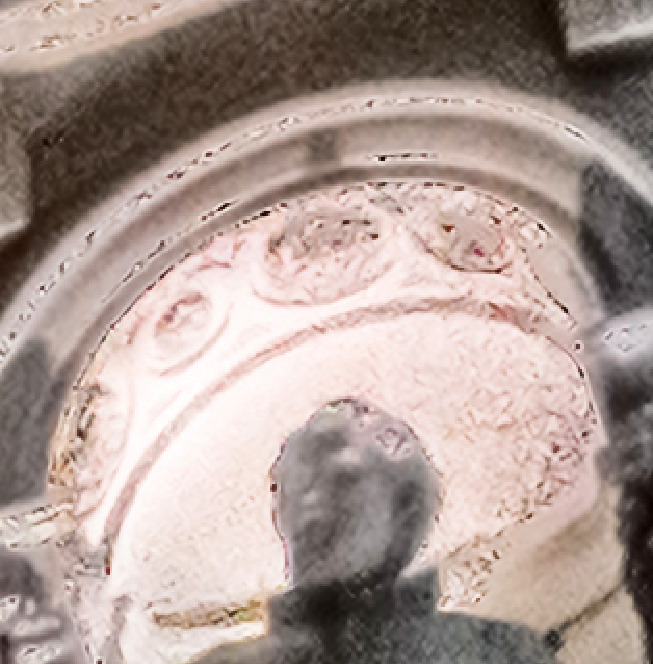}      & 
            \includegraphics[trim=0 0 0 100,clip,cfbox=red 0pt 0pt,width=0.16\textwidth]{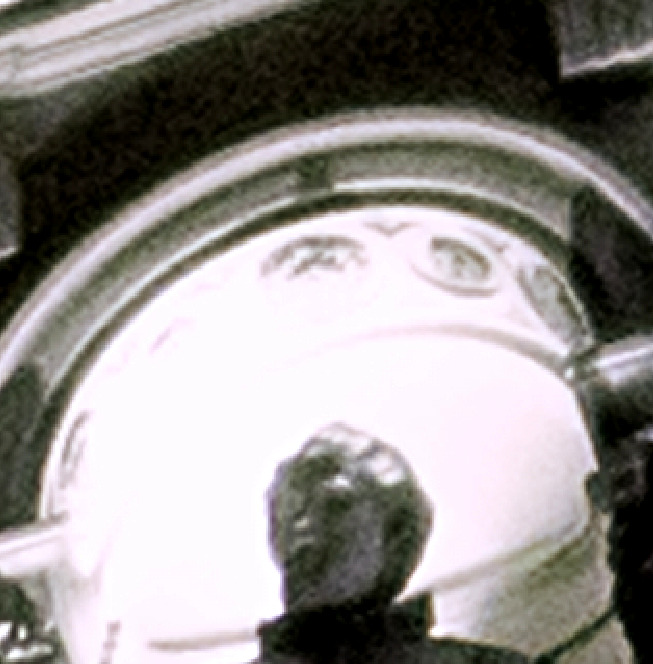}      & 
            \includegraphics[trim=0 0 0 100,clip,cfbox=red 0pt 0pt,width=0.16\textwidth]{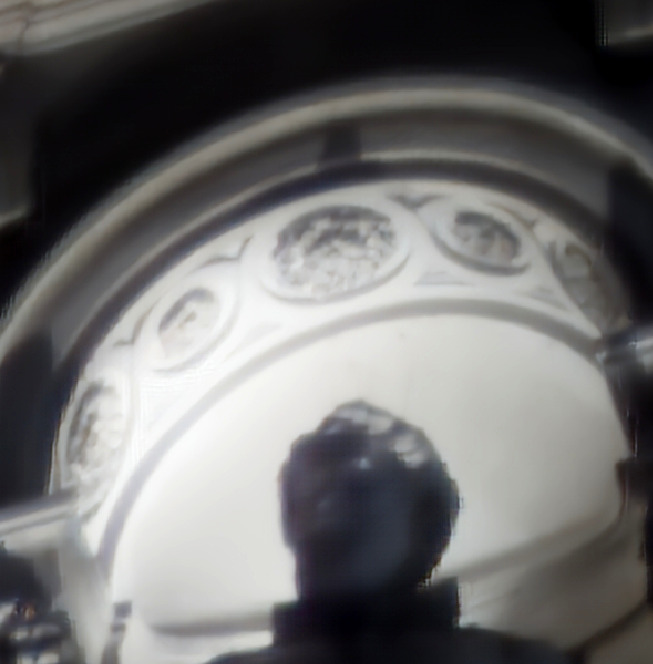}    \\ 

            \includegraphics[trim=0 0 0 100,clip,cfbox=blue 0pt 0pt,width=0.16\textwidth]{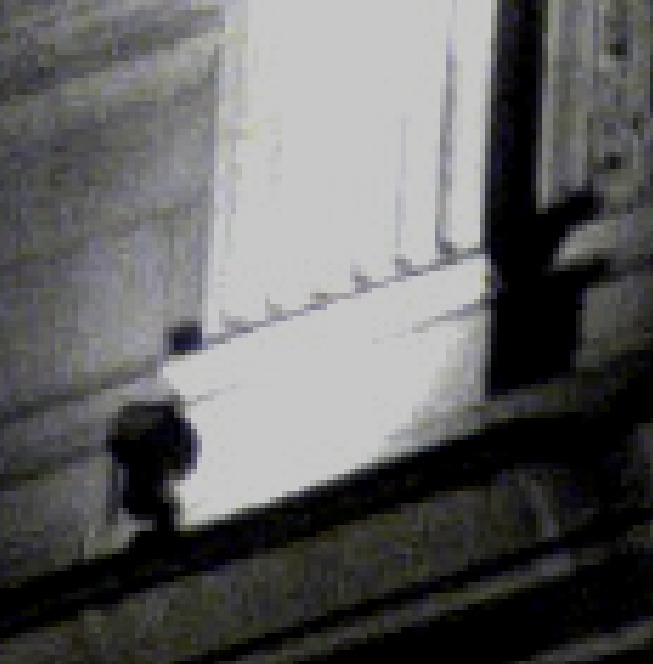}      & 
            \includegraphics[trim=0 0 0 100,clip,cfbox=blue 0pt 0pt,width=0.16\textwidth]{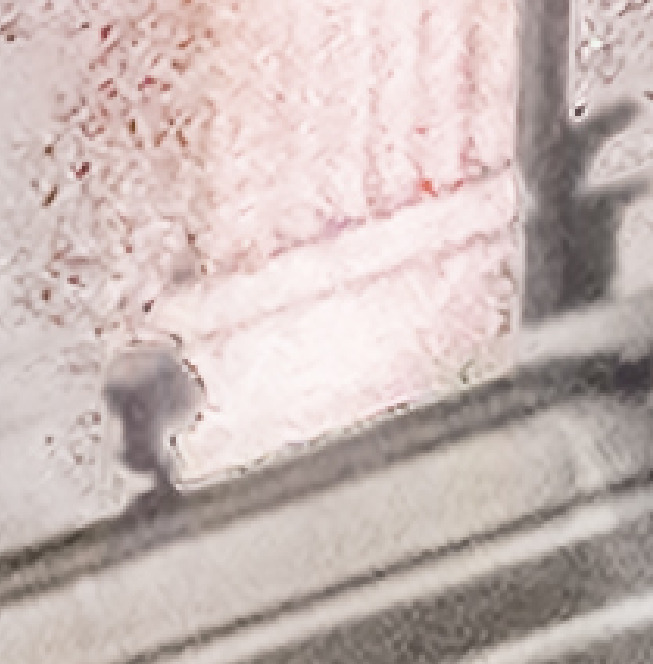}      & 
            \includegraphics[trim=0 0 0 100,clip,cfbox=blue 0pt 0pt,width=0.16\textwidth]{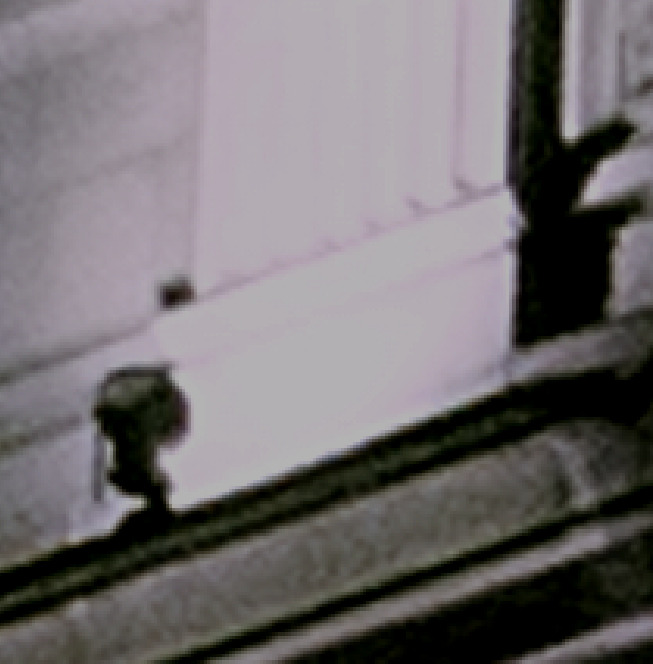}      & 
            \includegraphics[trim=0 0 0 100,clip,cfbox=blue 0pt 0pt,width=0.16\textwidth]{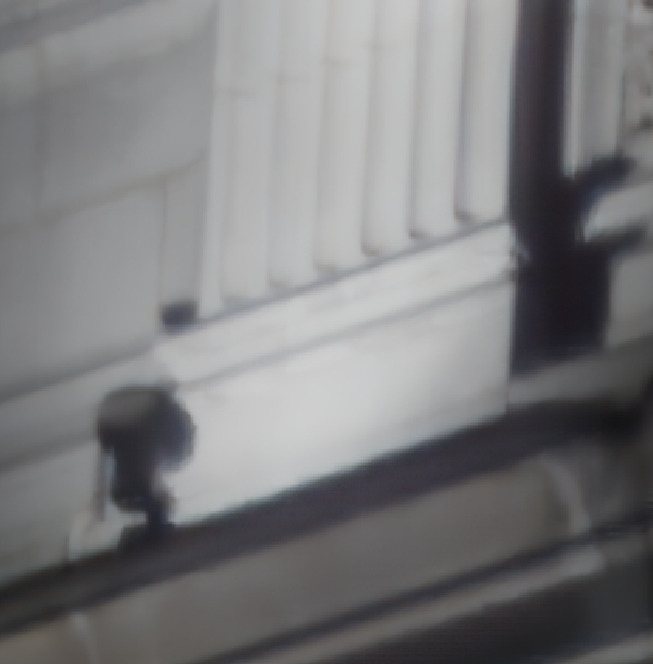}    \\ 

            \begin{tabular}{c} Low resolution \\ (dcraw) \end{tabular} &
            \begin{tabular}{c} ACR (x2) \\ (1 frame) \end{tabular} &
            \begin{tabular}{c} Kim et al (x2) \\ (1 frame)  \end{tabular} &
            \begin{tabular}{c} Ours (x4) \\ (18 frames) \end{tabular} 
        \end{tabular}
    \end{tabular}
    }
        \caption{ 
    \red{
Night-time comparisons of joint HDR imaging and super-resolution algorithms with bursts acquired by a Pixel4a. {\bf Left:} The central image in the burst (top) and our reconstruction (bottom). {\bf Right:} Comparison of close-ups of the reconstructions obtained by the CNN-based Adobe Camera Raw single-image algorithm for $\times 2$ super-resolution and demosaicking, the CNN-based $\times 2$ super-resolution method of \cite{kim19deep}, and our method.
} 
}

    \label{fig:lowlighqualitative2}
\end{figure*}

\paragraph{Dataset generation.}
Given a collection of $sRGB$ images, we construct bursts of LDR/LR raw images and HDR/HR RGB targets using the ISP inversion method of~\cite{brooks19unprocessing} and
our image formation pipeline, adjusting the gain
to simulate different exposure times.
The noise levels are sampled following the empirical model of Figure~\ref{fig:isoplot}.

\comment{We generate $n$ bursts $Y^{(i)}$ of synthetic raw SR images from both \texttt{.jpg} and 
\texttt{.hdr} images with various simulated exposures $\Delta^{(i)}$. The latter images are important to make our network
robust to artefacts occurring near saturated areas. We use the \texttt{.hdr} images from \cite{liu20single}.}

\paragraph{Training loss.}
With this training data in hand, we supervise our model using the $\ell_1$  distance between 
the target irradiance images $x^{(i)}$ and the predicted ones 
$F_\theta (Y^{(i)}, \Delta^{(i)})$, and minimize the cost function:
\begin{equation}\label{eq:trainingloss}
    \min_\theta \sum_{i=1}^n \left\|x^{(i)} - F_\theta (Y^{(i)},\Delta^{(i)}) \right\|_1.
\end{equation}
\comment{By using a normalized scheme, we avoid the sigmoid
activation at the top layer of recent CNNs for HDR imaging~\cite{kalantari17deep, wu18deep, yan19attention} forcing the output to be between 0 and 1.}
We have also tried to use the so-called $\mu$-law~\cite{kalantari17deep}
to include some kind of tone mapping in the supervision but it only marginally
improved the visual quality of the images predicted by our model.

\paragraph{Optimizer.} We minimize Eq.~\eqref{eq:trainingloss} using
Adam optimizer with learning rate set to $10^{-4}$
for 400k iterations. We decrease the learning
rate by 0.5 every 100k iterations. The weights of the CNNs are randomly initialized with the default setting of the PyTorch library.

\section{Results}
We first show in Section~\ref{subsec:srhdr} several qualitative results illustrating
the performance of our method for joint HDR imaging, super-resolution,
demosaicking and denoising from real raw image bursts. Qualitative and
quantitative comparisons with existing methods for super-resolution, HDR
imaging, and registrations are presented in Section~\ref{subsec:sr}, Section~\ref{subsec:hdr}  and Section~\ref{subsec:registr} 
respectively. The effect of the choice of prior and the robustness of
our method for real images are discussed in Section~\ref{subsec:discussions}. Additional
results, ablations studies, and discussions of its limitations can be
found in the appendix.

Note that all the HDR images are rendered using  {\em Photomatix}\footnote{\texttt{https://www.hdrsoft.com/}} for tone mapping, which is itself a challenging task~\cite{reinhard02photographic} beyond the scope of this paper.
For baselines operating on RGB images instead of raw photographs, we first process raw files with Adobe Camera Raw to generate RGB images with the highest quality possible.

\subsection{Joint SR and HDR on raw image bursts}\label{subsec:srhdr}

To the best of our knowledge, we are the first to address jointly HDR, super resolution, demosaicking, and denoising on bursts of raw images. Therefore, we will mostly
present here  qualitative results, and will defer quantitative comparisons to the following sections that evaluate the  performance of our algorithm on separate HDR or SR tasks.

We consider bursts acquired in different settings by a Pixel 3a or 4a camera, by using an Android application to shoot bursts of 11 to 18 raw images. We choose an EV step of 1/3 to 2/3 between each shot. This is particularly important for night scenes to  avoid motion blur in the longest-exposure frames.
Our method successfully restores finer details and extends the dynamic range of the original shot by denoising dark areas and restoring clipped signals. More precisely: 
\begin{itemize}
    \item Figures~\ref{fig:lowlighqualitative1} and~\ref{fig:lowlighqualitative2} show night-time photos with large dynamics, similar to Figure~\ref{fig:teaser}, with both under- and over-exposed areas in the low-resolution central frame. Both the dynamics and the resolution are significantly improved by our algorithm.  
    \item An outdoor day-time photograph is shown in Figure~\ref{fig:dayqualitative2}, with a particularly large dynamic range. The scene contains both under-exposed, noisy areas in the shadows and large bright saturated areas. \red{Note also that the scene contains patterns which are smaller than the resolution of the native image which is a particularly hard setting for demosaicking. Our approach handles such situations well.}
    \item A night scene with both very dark building parts and light bulbs, resulting in a very large dynamic range (Figure~6). Our approach, unlike our
    competitors, can recover details in both the dark and saturated
    areas.
\end{itemize}

\begin{figure*}[ht]
    \caption{Visual comparison for super-resolution only on real same-exposure raw bursts, of respectively $K=20$ and $K=30$
    frames, with state-of-the-art competitors. 
    We {\em do not} present HDR results in this figure. Our approach limits Moir\'e artefacts in the 
    first row and reveals in general more high frequency image details in both rows. The last row shows an example requiring deghosting. The ghosted LR image on the left 
    is obtained by averaging the whole burst to show the pedestrian's motion. \citet{bhat21deep} and our method effectively handle small object motions. The reader is invited to zoom in.
    }
    \setlength\tabcolsep{0.5pt}
    \renewcommand{\arraystretch}{0.5}
    \centering
    \scalebox{1.02}{
    \begin{tabular}{ccccc}
    \includegraphics[trim=0 0 0 0,clip,width=0.19\textwidth]{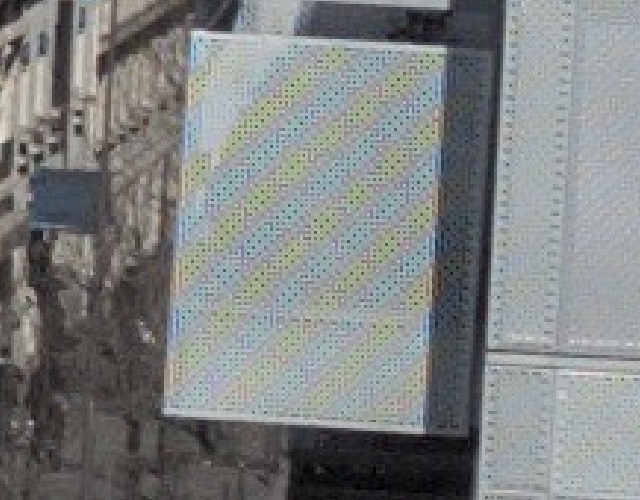}        &  
    \includegraphics[trim=0 0 0 0,clip,width=0.19\textwidth]{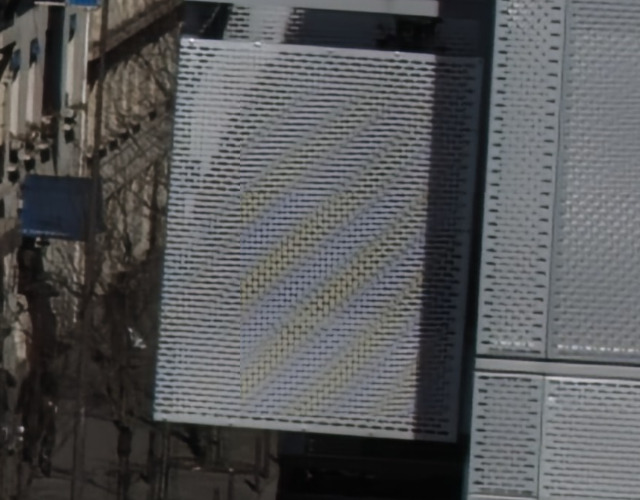}        &  
    \includegraphics[trim=0 13 0 0,clip,width=0.19 \textwidth]{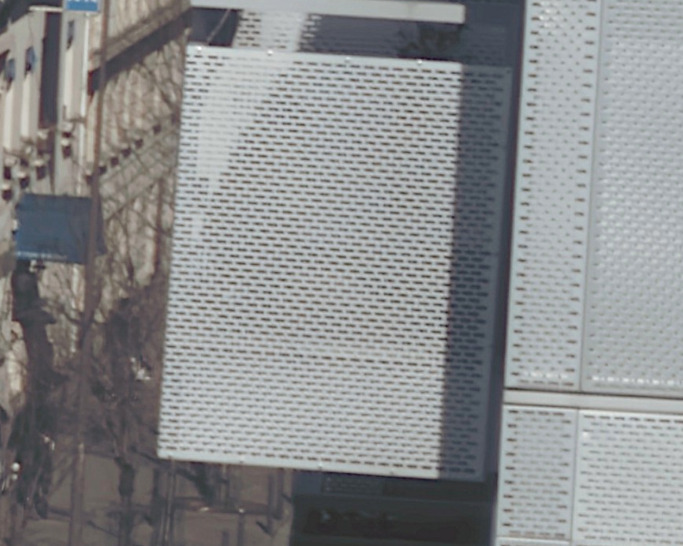}        &  
    \includegraphics[trim=0 0 0 0,clip,width=0.19\textwidth]{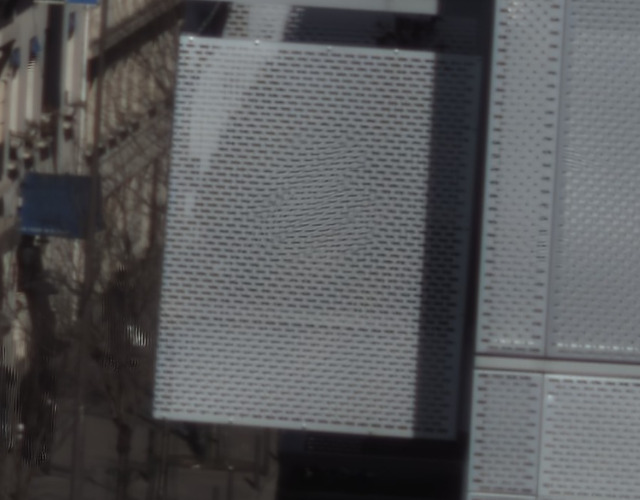}        &  
    \includegraphics[trim=0 0 0 0,clip,width=0.19\textwidth]{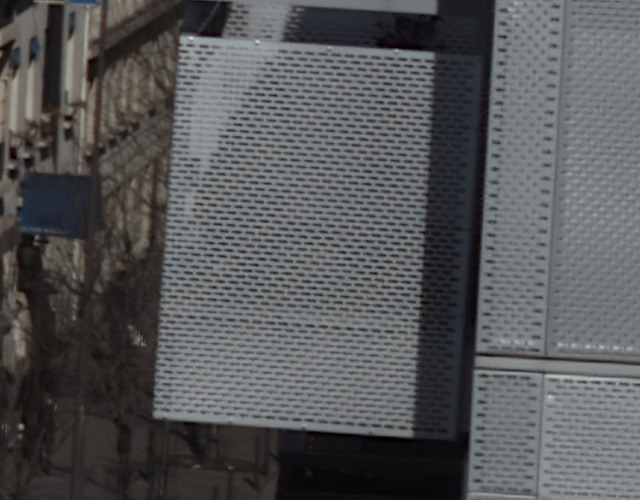}        \\ 
    \includegraphics[trim=0 0 0 0,clip,width=0.19\textwidth]{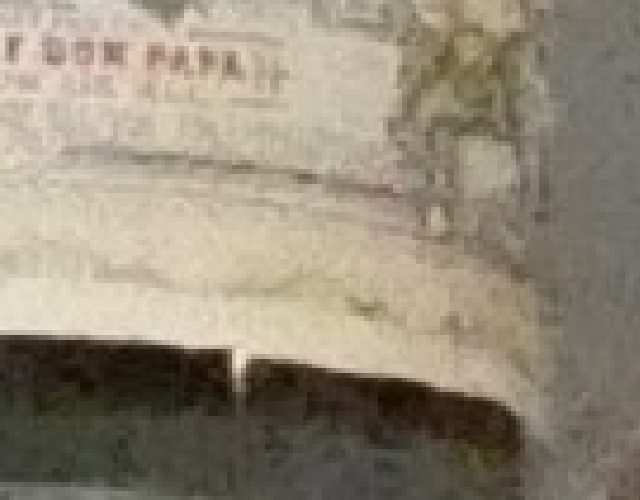}        &  
    \includegraphics[trim=0 0 0 0,clip,width=0.19\textwidth]{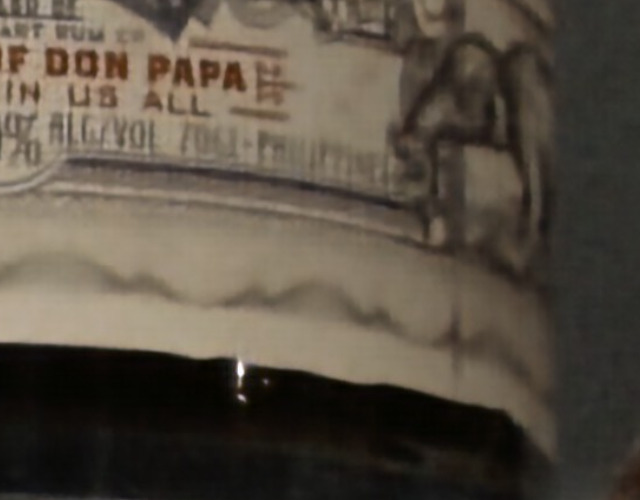}        &  
    \includegraphics[trim=0 25 0 0,clip,width=0.19\textwidth]{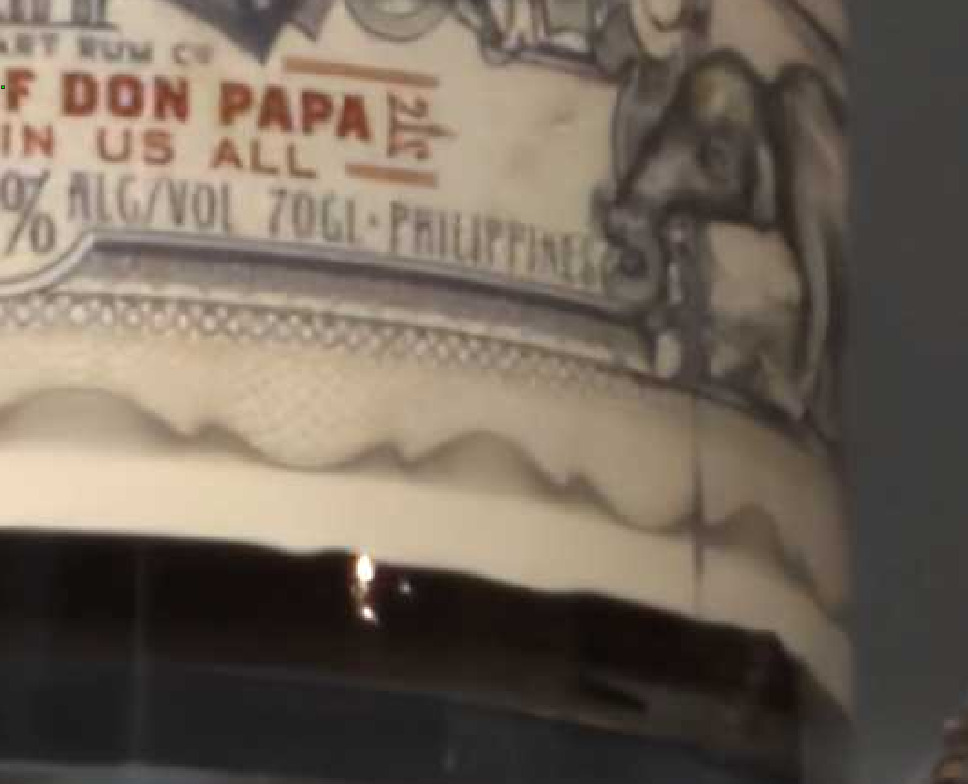}        &  
    \includegraphics[trim=0 0 0 0,clip,width=0.19\textwidth]{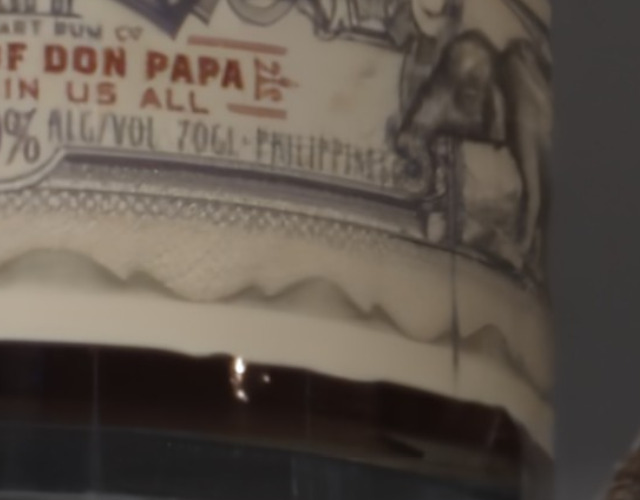}        &  
    \includegraphics[trim=0 0 0 0,clip,width=0.19\textwidth]{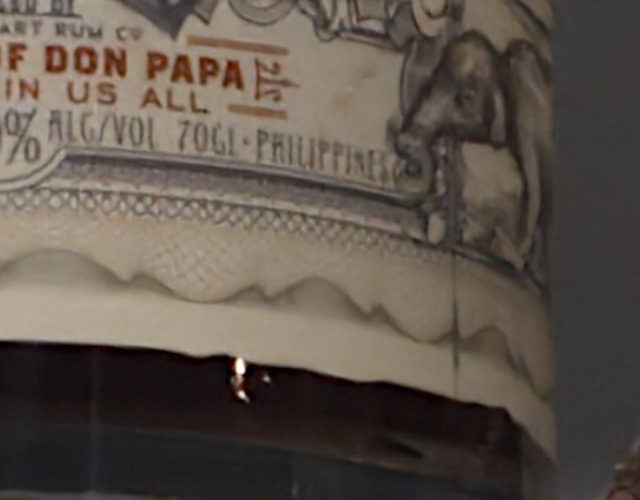}        \\ 

    \includegraphics[trim=0 0 20 0,clip,width=0.19\textwidth]{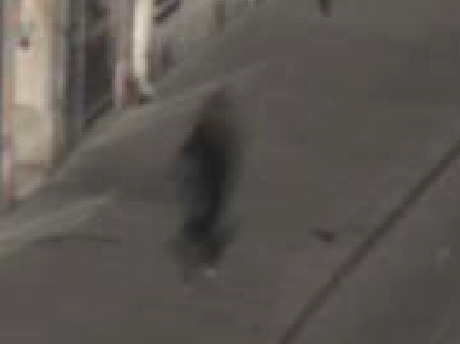}        &  
    \includegraphics[trim=0 0 0 0,clip,width=0.19\textwidth]{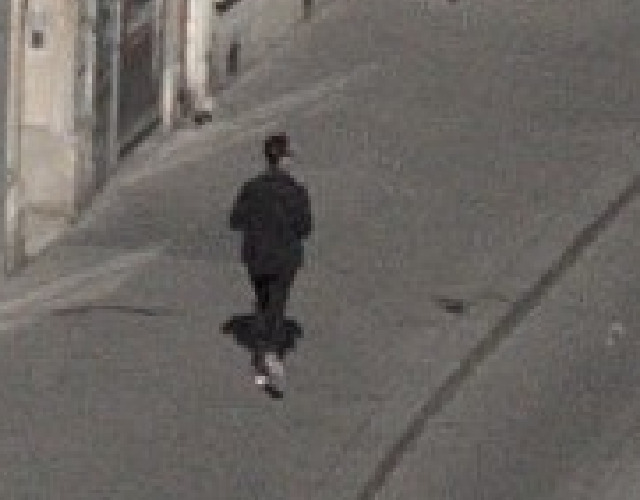}        &  
    \includegraphics[trim=0 500 0 20,clip,width=0.19\textwidth]{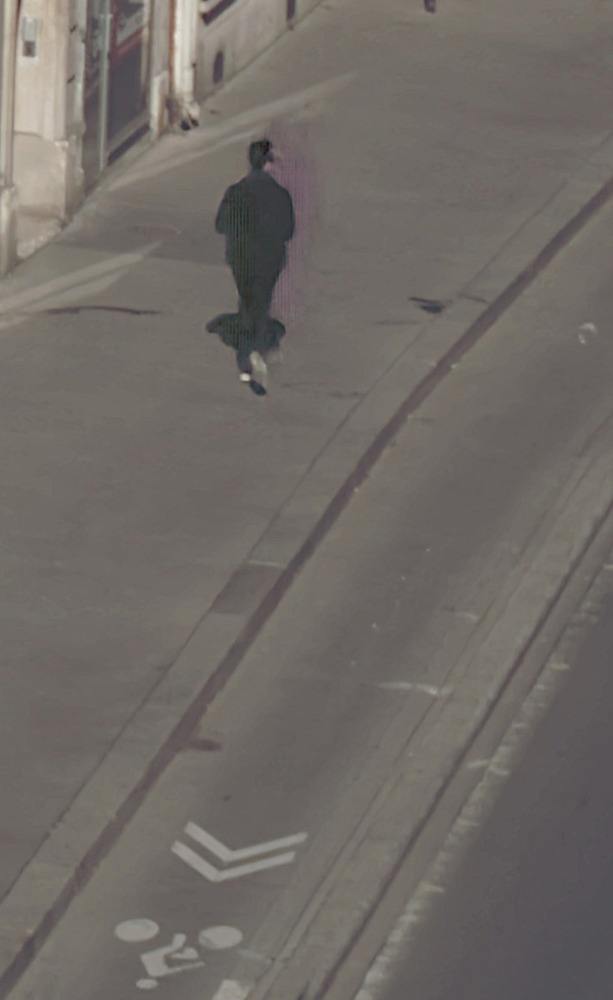}        &  
    \includegraphics[trim=0 0 0 0,clip,width=0.19\textwidth]{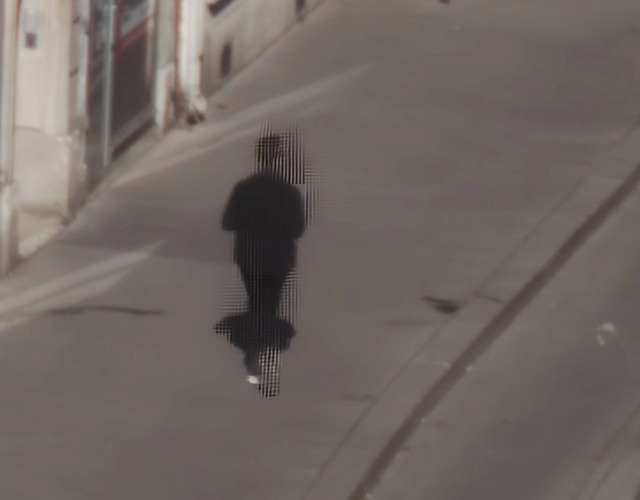}        &  
    \includegraphics[trim=0 0 0 0,clip,width=0.19\textwidth]{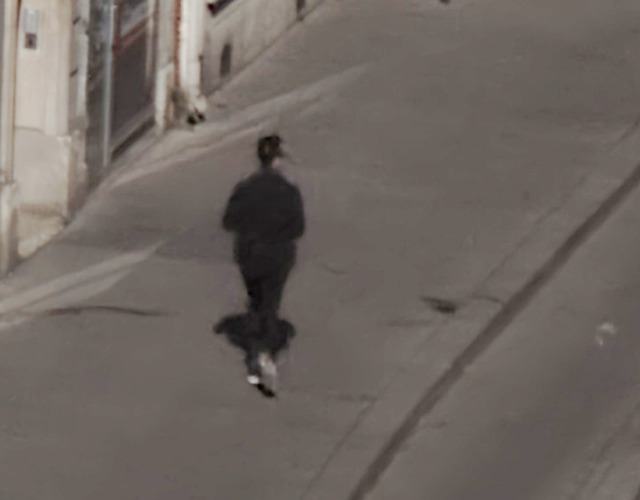}        \\
    
     Low resolution & \cite{bhat21deep} & \cite{lecouat21aliasing}& \cite{luo2021ebsr} & Ours
    \end{tabular}
    }

    \label{fig:SRrealburst}
\end{figure*}

\subsection{Pure super-resolution}\label{subsec:sr}
We now move to pure \red{\red{super-resolution} from raw image bursts, and compare our approach with~\citet{lecouat21aliasing},} using examples from their paper.
The bursts in this section all have the same exposure, 
making the alignment simpler compared to
the previous section.
We first perform a quantitative evaluation on the semi-synthetic benchmark of~\citet{bhat21deep}, following their experimental setup and using their dataset.
Table~\ref{tab:SR4synthetic} presents a comparison of our approach for SR, which can be seen as an improved variant of \citet{lecouat21aliasing}. All methods in the comparison are designed to process raw image bursts.
We first note that our improvements in the image registration module 
yields +1dB over \cite{lecouat21aliasing} for similar network capacities.
The geometric error, measuring alignment discrepancies, is also four times smaller than that of
\cite{lecouat21aliasing}, which further suggests the usefulness of our modified Lucas-Kanade
module.
Since the other methods of the panel do not explicitly predict any motion vector, we cannot
compute the \red{corresponding} geometric errors.
We also have a PSNR gain of about 0.5 to 1dB over three of the recent competitors and fall only behind
\citet{luo2021ebsr} by less than 1dB but with 13 times \red{fewer} learnable parameters.
\red{Therefore}, the proposed approach is also a compact and competitive algorithm for SR alone. A \red{speed comparison}, presented later in Section~\ref{subsec:speed} also shows that \red{our method is faster at inference time}.

The previous \red{comparison} is conducted on semi-synthetic data, both for training the models and for testing, which makes \red{its} conclusions difficult
to generalize to the real world of raw bursts from handheld cameras. Nevertheless, it remains the best existing \red{quantitative} experimental setup, to our knowledge, since it is not possible to acquire reliable HR ground-truth data along with LR raw bursts. 
Figure~\ref{fig:SRrealburst} shows two challenging real-world examples
on which we compare \red{qualitatively} the approaches of \citet{bhat21deep}, \citet{lecouat21aliasing} and \citet{luo2021ebsr} to ours, \red{for $\times4$ super-resolution factor.}
We display in the first row the results for a burst of $K=20$ raw frames
of a textured surface.
Moir\'e artefacts and aliasing can respectively be noticed in the results from~\citet{bhat21deep} and~\citet{luo2021ebsr}.
\red{Such artefacts are not visible in our reconstruction and that of 
of \citet{lecouat21aliasing}.}
The second row shows the results for a burst of $K=30$ raw images
from \cite{lecouat21aliasing}. Amongst the four methods in the panel,
ours returns the sharpest image, with for instance easier-to-read
characters than competitors.
We point out that we {\em have not} used any sharpening algorithm
on any of these images.

\begin{figure}[ht]
    \setlength\tabcolsep{0.5pt}
    \renewcommand{\arraystretch}{0.5}
    \centering
    \begin{tabular}{ccc}
 \includegraphics[trim=0 50 50 0,clip,width=0.155\textwidth]{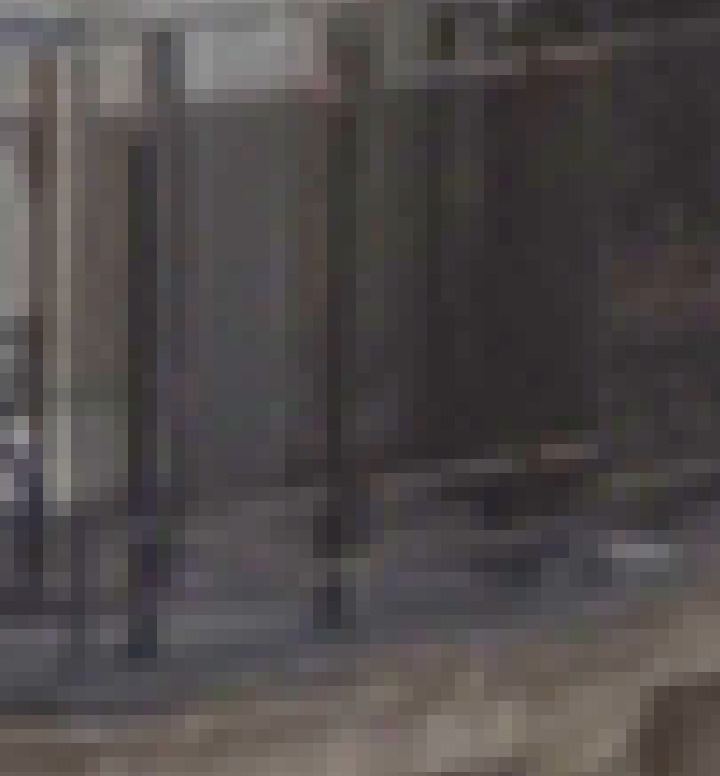}        &
 \includegraphics[trim=0 50 50 0,clip,width=0.155\textwidth]{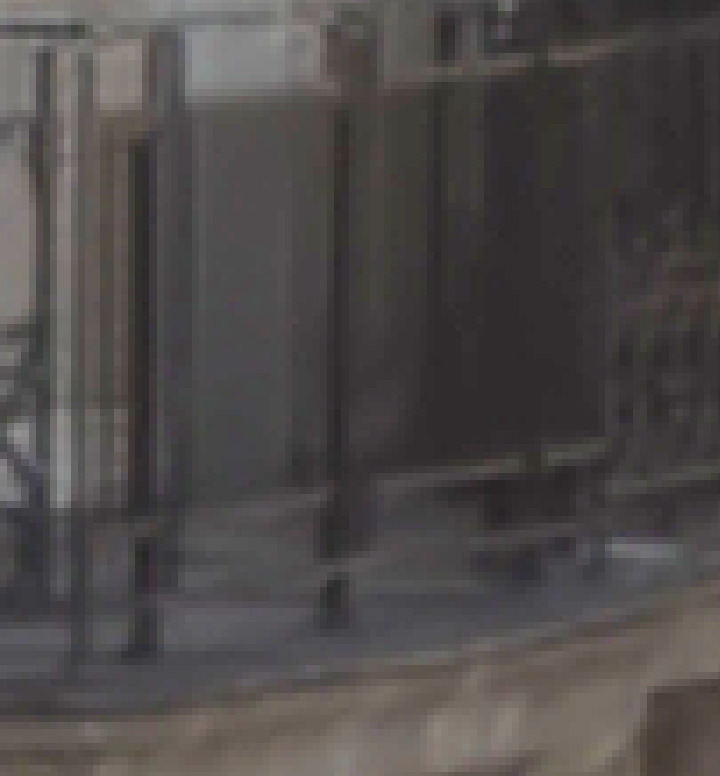}        &
  \includegraphics[trim=0 50 50 0,clip,width=0.155\textwidth]{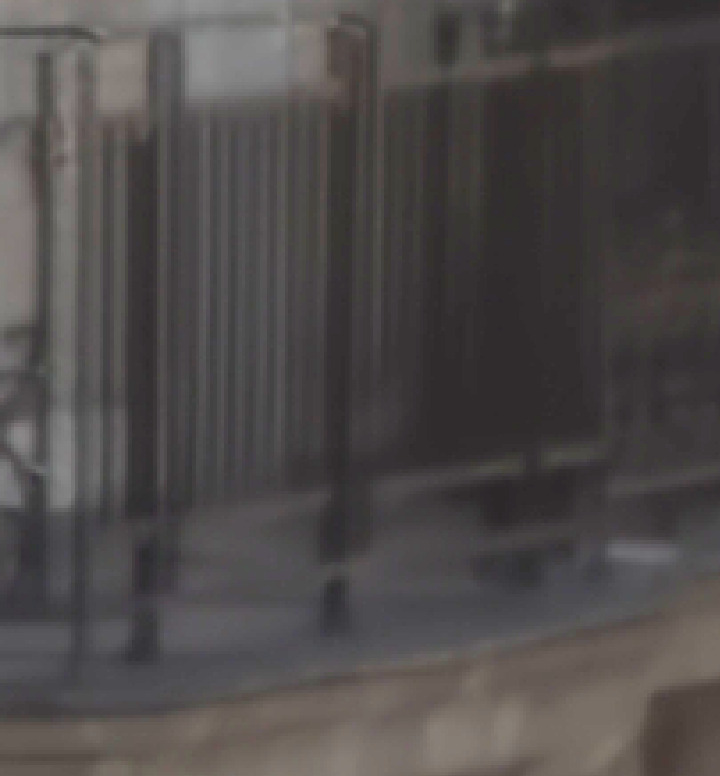}        \\
  \includegraphics[trim=300 530 260 80,clip,width=0.155\textwidth]{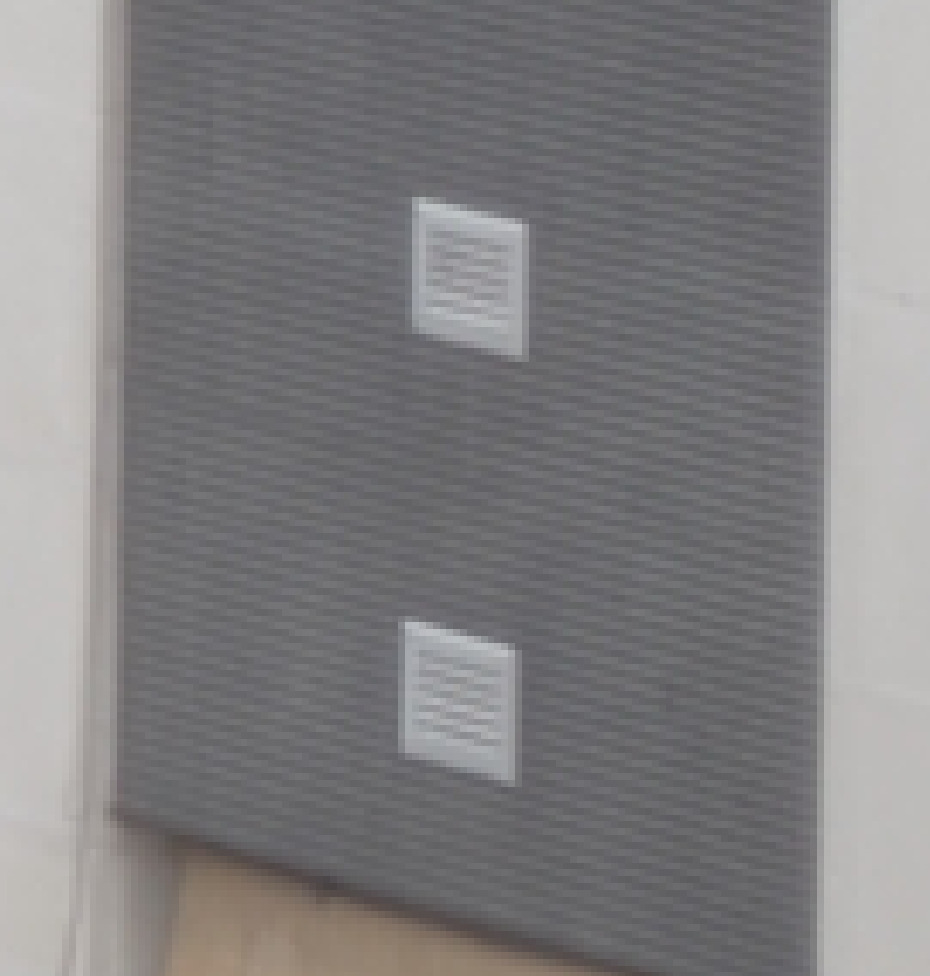}        &
 \includegraphics[trim=300 530 260 80,clip,width=0.155\textwidth]{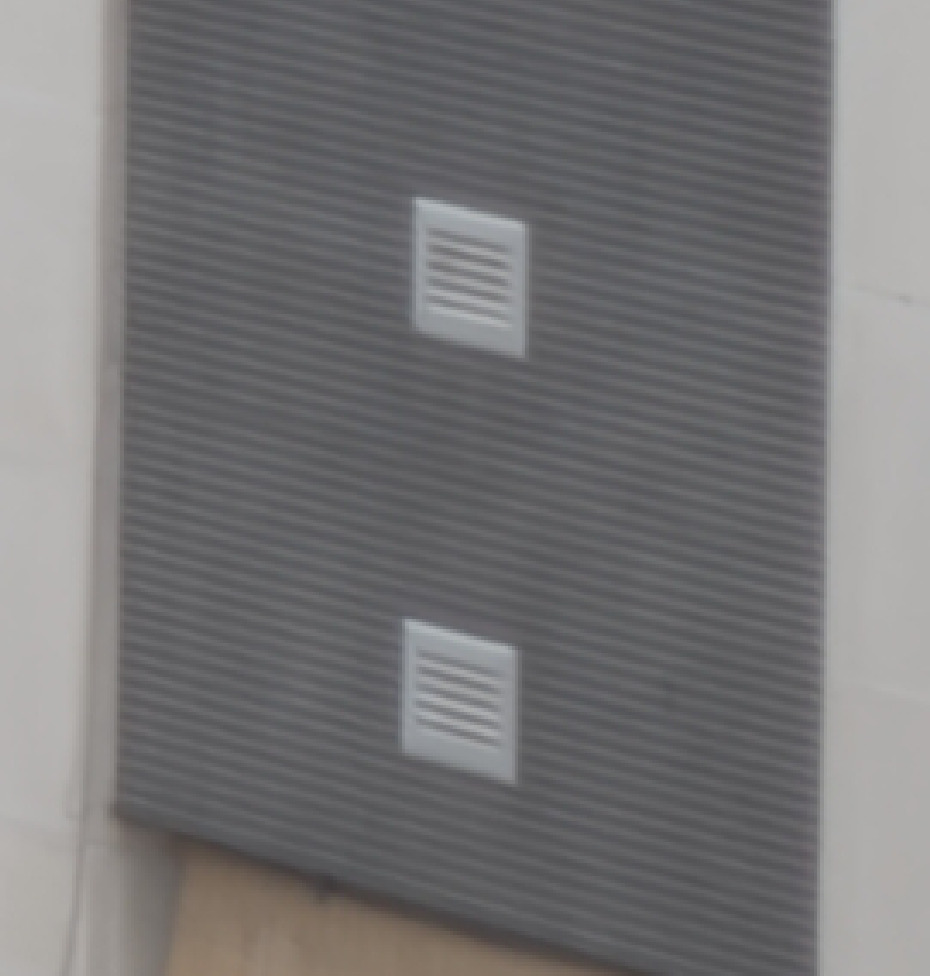}        &
  \includegraphics[trim=300 530 260 80,clip,width=0.155\textwidth]{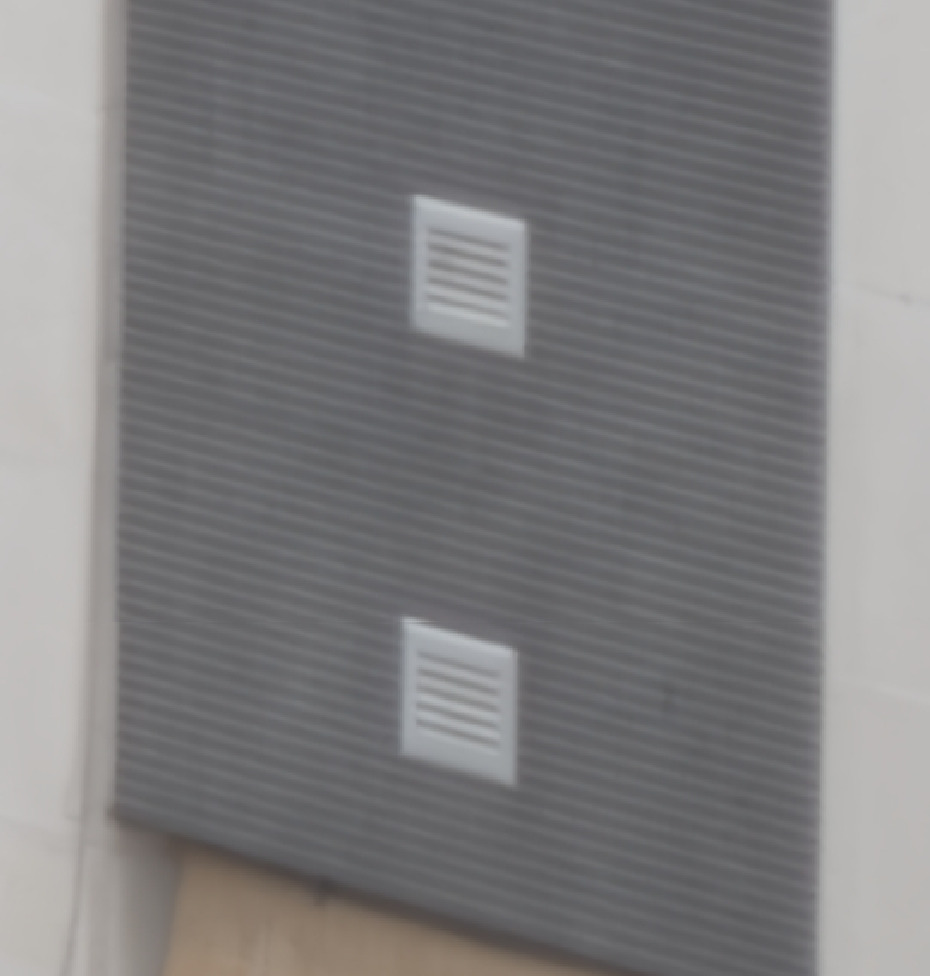}        \\
LR & HR~($\times2$) & HR~($\times4)$ 
    \end{tabular}
    \caption{Visualizing super-resolution limit at resolutions increased by $\times2$ and $\times4$ with our model. The
    first image in the first row benefits from the $\times4$
    improvements whereas the one in the second row (from the same
    photograph) is not further enhanced after $\times2$. See the discussion in the text.}
    \label{fig:SRlimit}
\end{figure}

As remarked by \citet{wronski19handheld}, there is a physical limit to the
maximum frequency one can reconstruct with aliasing, \red{due to} the sensor pitch or the 
lens point-spread function. We verify this property
in Figure~\ref{fig:SRlimit} where we show two crops from the same image,
\red{with $\times2$ and $\times4$ resolution factors}.
The first row shows details of a balcony clearly 
benefiting from a $\times4$ gain in resolution compared to its $\times2$ counterpart.
The second row \red{shows} however that sometimes, 
\red{as predicted  by \citet{wronski19handheld}, $\times4$ upsampling factor may not reveal finer details that its $\times2$ counterpart.}

\begin{table}[t]
    \caption{Super-resolution ($\times 4$) comparison with a selected panel of
recent methods with average PSNR and geometric error when it can be
computed. We {\em do not} perform HDR generation in this experiment. 
Our method falls behind that \citet{luo2021ebsr} within a margin of 
less than 1dB but with 13 times 
fewer parameters. We gain 1dB compared to
\cite{lecouat21aliasing} with 
a similar number
of parameters by upgrading the registration module.}
\begin{tabular}{lrcc}
\toprule
Model &  \# parameters & PSNR  & Geom (avg) \\ \midrule
\cite{bhat21deep} &13M & 40.76 & N/A \\
\cite{lecouat21aliasing} & 3M& 41.45 & \underline{2.56} \\ %
\cite{bhat21reparameterization}& - & 41.56 &  N/A  \\
\cite{dudhane2021burst} &  6.6M & 41.93 &  N/A \\
\cite{luo2021ebsr}  &26M & \textbf{43.35} &  N/A \\
Ours & 3M &\underline{42.42} & \textbf{0.80}  \\ %
\bottomrule
\end{tabular}

\label{tab:SR4synthetic}
\end{table}

\begin{figure*}[htb]
    
\setlength\tabcolsep{0.5pt}
\renewcommand{\arraystretch}{0.5}
\centering
\begin{tabular}{ccccc}
\includegraphics[trim=0 0 0 0,clip,height=0.185\textwidth]{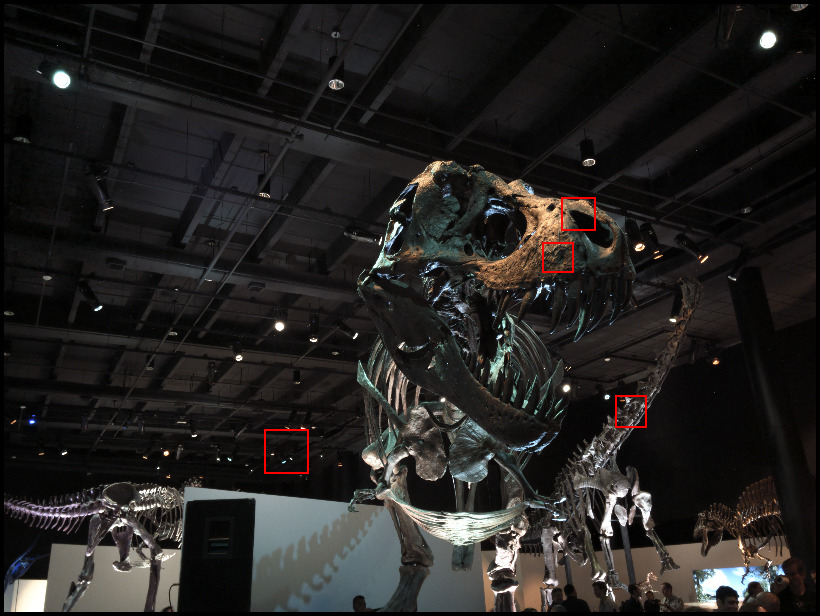} &
\includegraphics[trim=0 0 0 0,clip,height=0.185\textwidth]{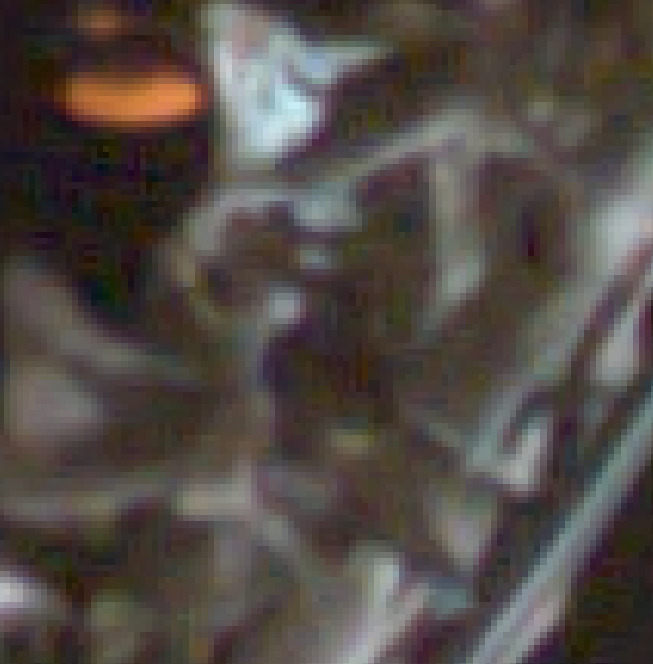} & 
\includegraphics[trim=0 0 0 0,clip,height=0.185\textwidth]{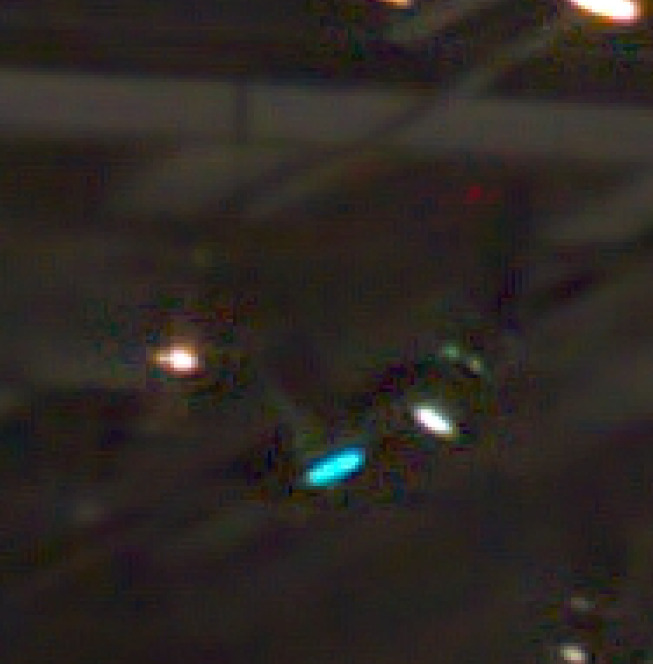} & 
\includegraphics[trim=0 0 0 0,clip,height=0.185\textwidth]{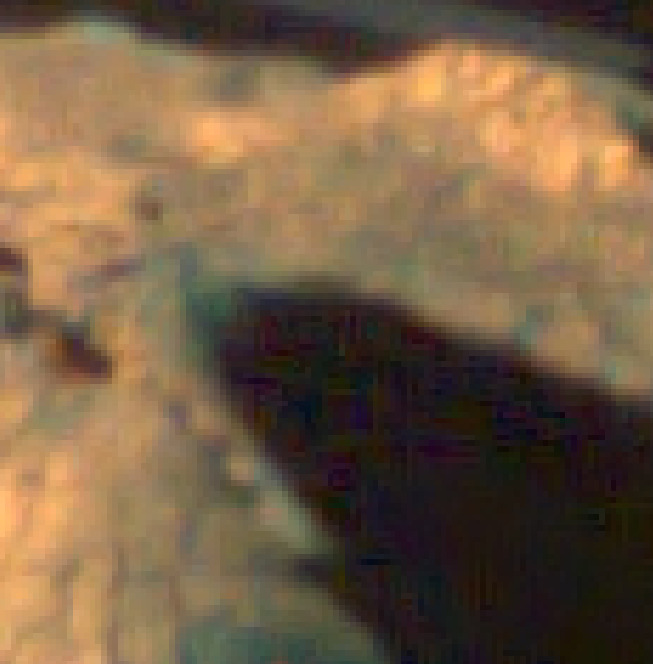} & 
\includegraphics[trim=0 0 0 0,clip,height=0.185\textwidth]{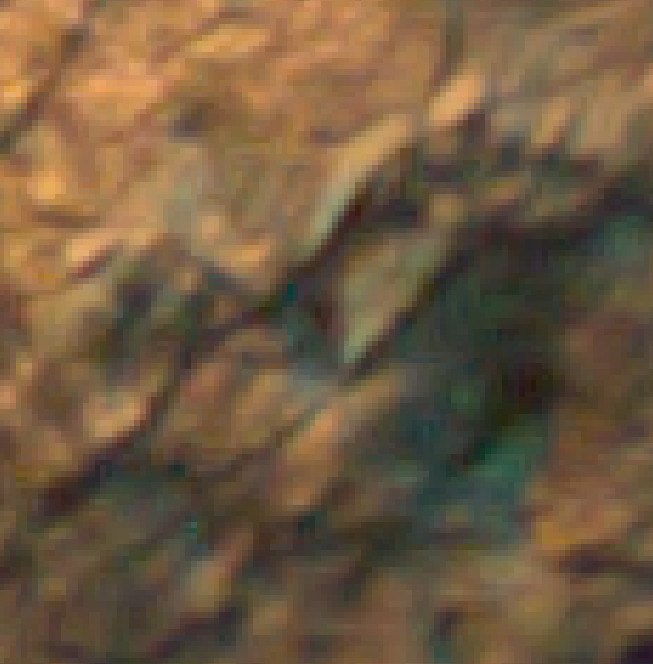}  \\
\includegraphics[trim=0 0 0 0,clip,height=0.185\textwidth]{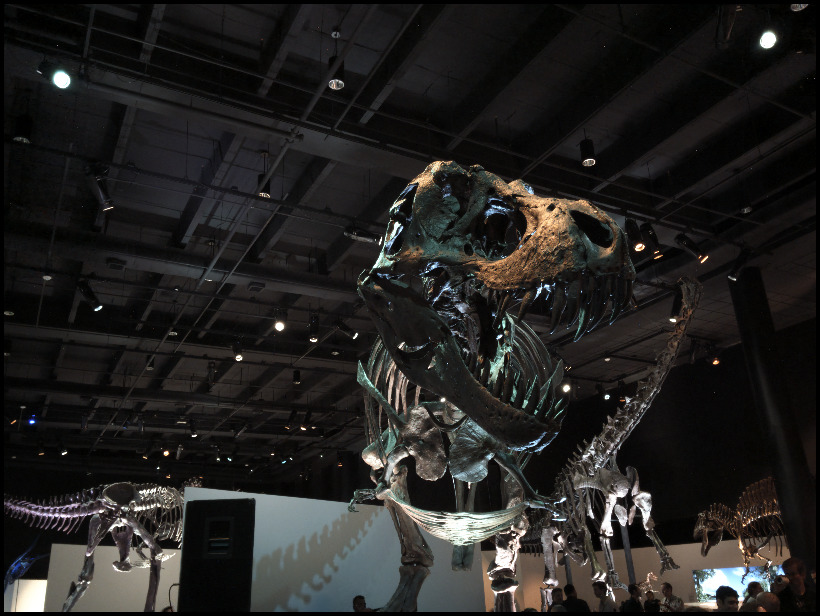} &
\includegraphics[trim=0 0 0 0,clip,height=0.185\textwidth]{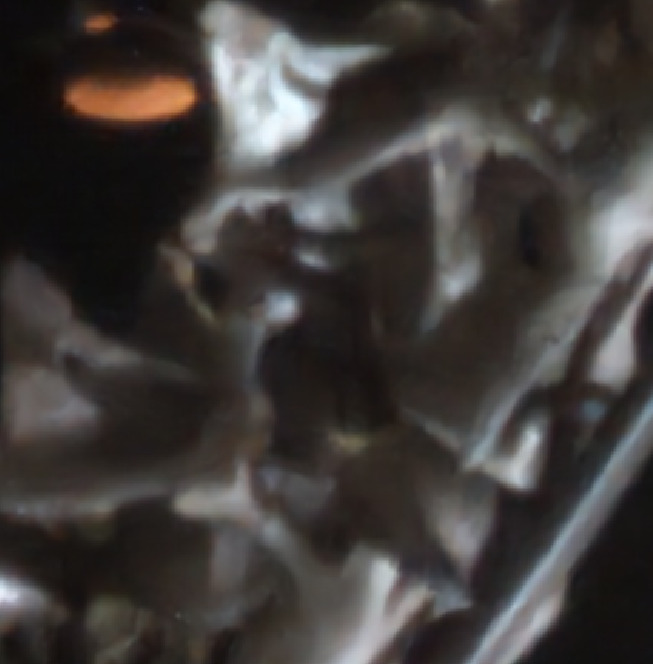} & 
\includegraphics[trim=0 0 0 0,clip,height=0.185\textwidth]{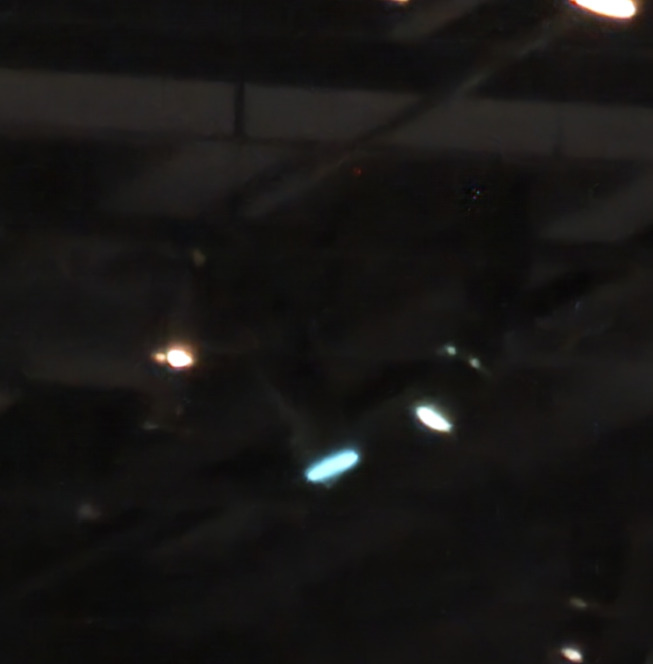} & 
\includegraphics[trim=0 0 0 0,clip,height=0.185\textwidth]{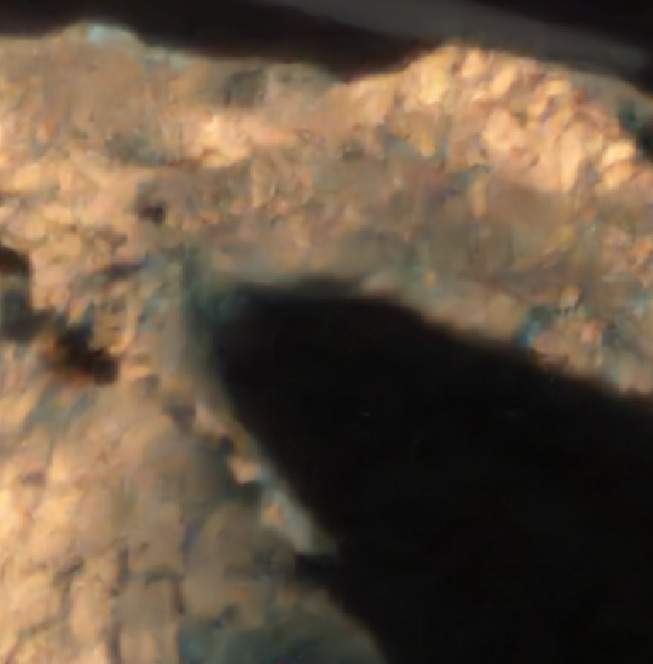} & 
\includegraphics[trim=0 0 0 0,clip,height=0.185\textwidth]{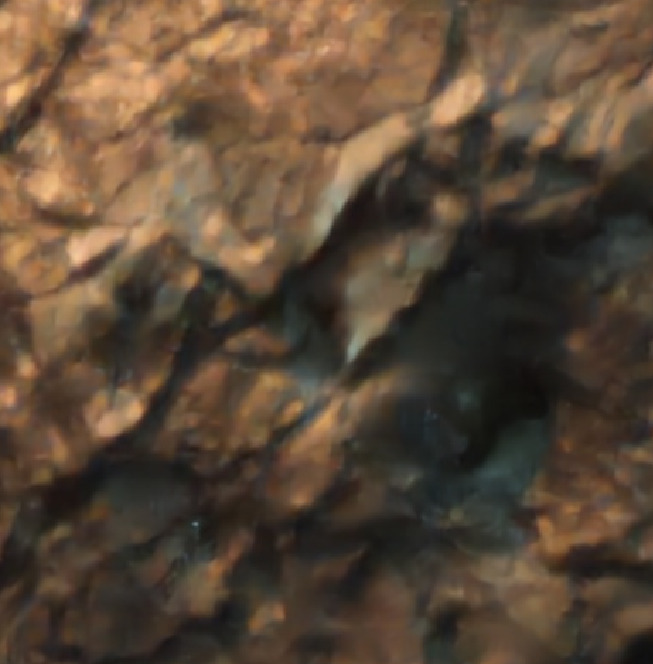}  \\
\end{tabular}

\caption{
\red{
Comparison between the public-domain implementation \cite{monod2021analysis} dubbed here HDR+Ipol of Google’s HDR+ \cite{hasinoff16burst} (top) with our HDR/SR$\times 4$ method. {\bf Left:} The images reconstructed by HDR+Ipol (top) and our method (bottom) from a burst of 8 same-exposure images acquired by a Nexus 5. Note that they are barely distinghuisable at this resolution. {\bf Right:} Crops showing that our algorithm reveals finers details while effectively suppressing noise in dark areas. 
}
}

\label{fig:hdrplus}
\end{figure*}

\begin{figure*}
\centering
\small
\setlength\tabcolsep{0.5pt} 
\renewcommand{\arraystretch}{0.5}
\scalebox{1.12}{
\begin{tabular}{cc}

\begin{tabular}{ccc}
\includegraphics[height=0.044\textwidth,trim=0 0 0 250]{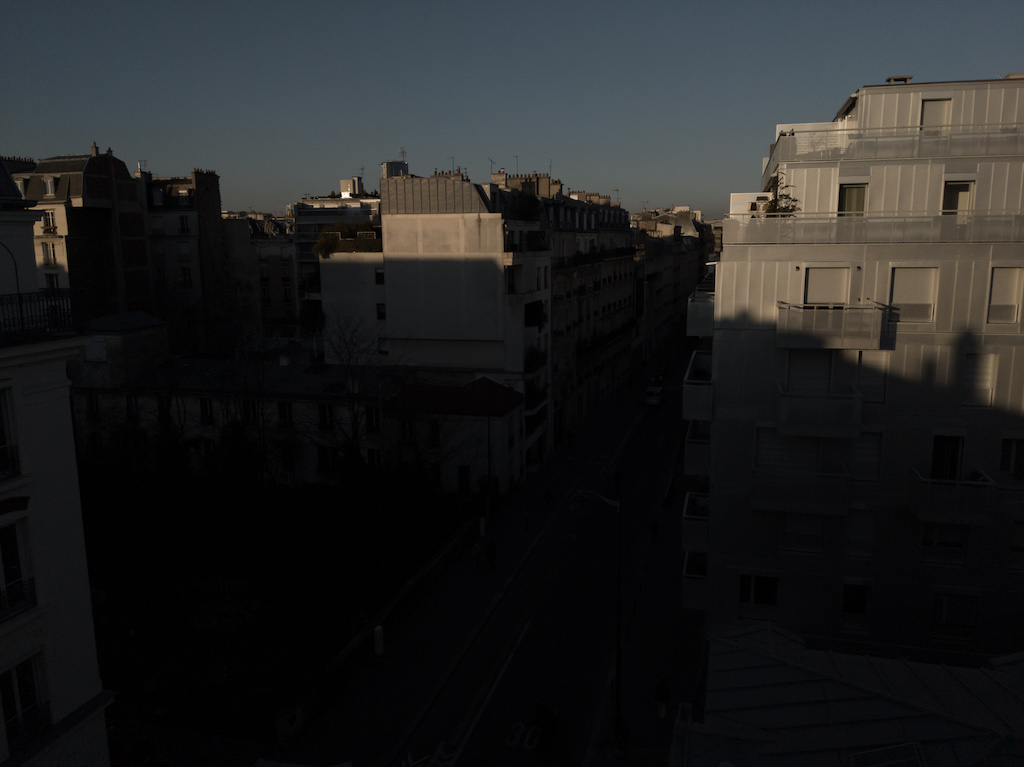} & 
\includegraphics[height=0.044\textwidth,trim=0 0 0 250]{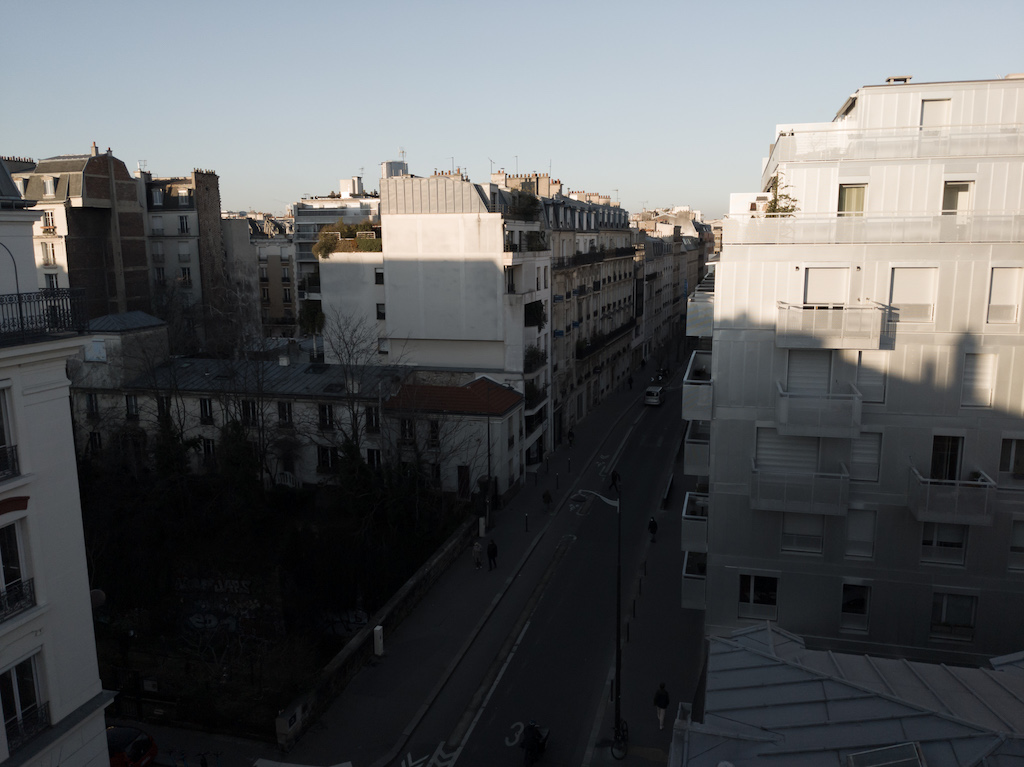} &
\includegraphics[height=0.044\textwidth,trim=0 0 0 250]{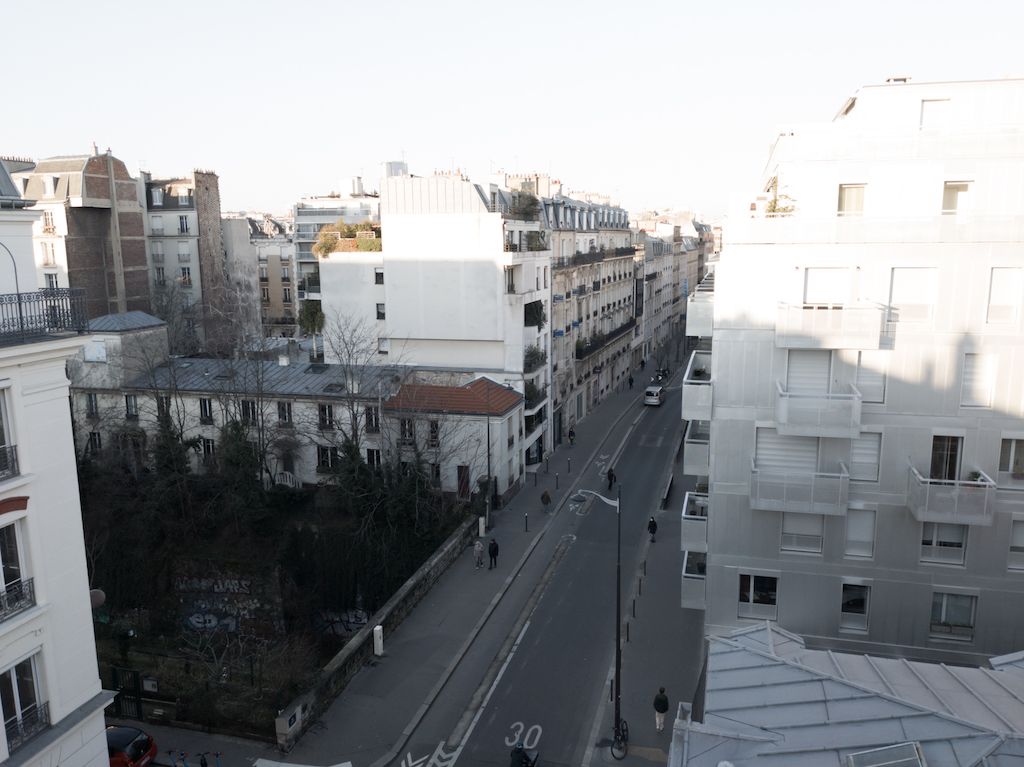} \\ 
\multicolumn{3}{c}{\includegraphics[height=0.2\textwidth]{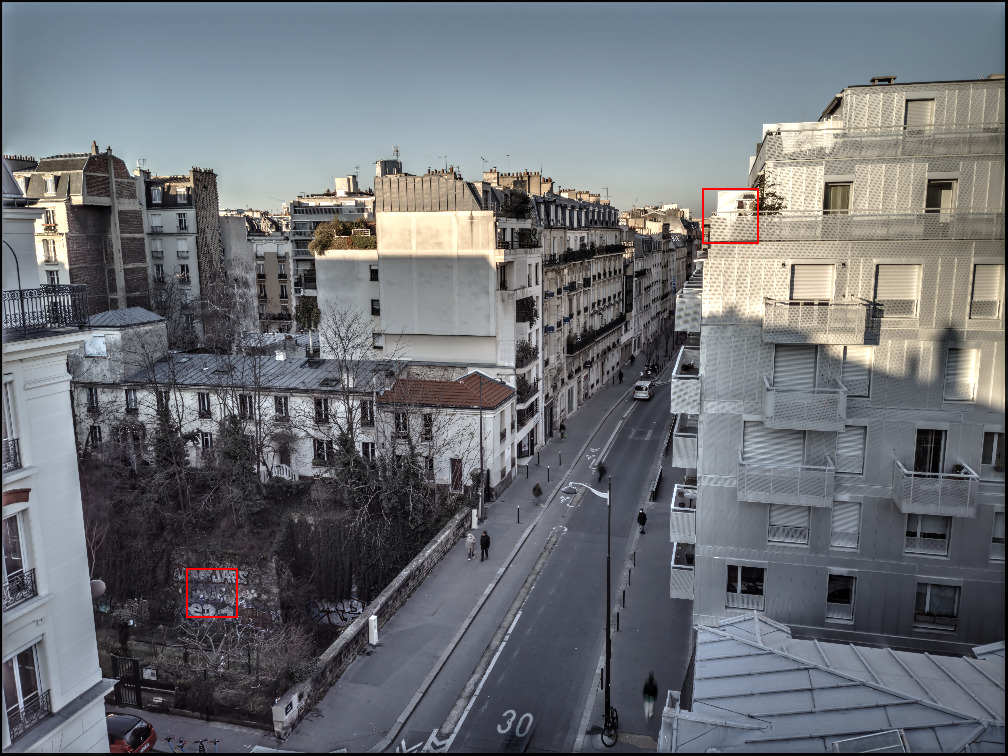}}
\end{tabular}

\begin{tabular}{ccccc}
\includegraphics[trim=0 0 0 0,clip,width=0.12\textwidth]{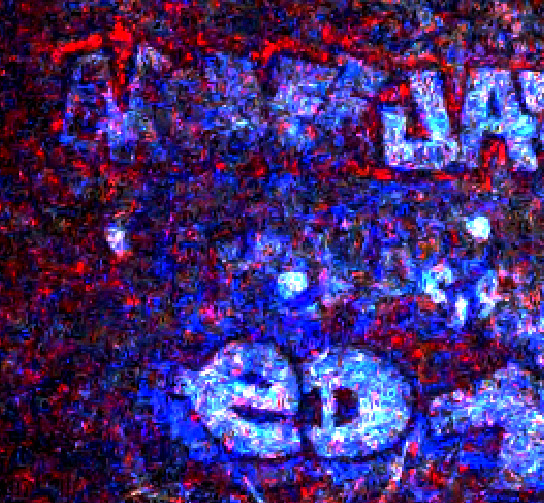}&
\includegraphics[trim=0 0 0 0,clip,width=0.12\textwidth]{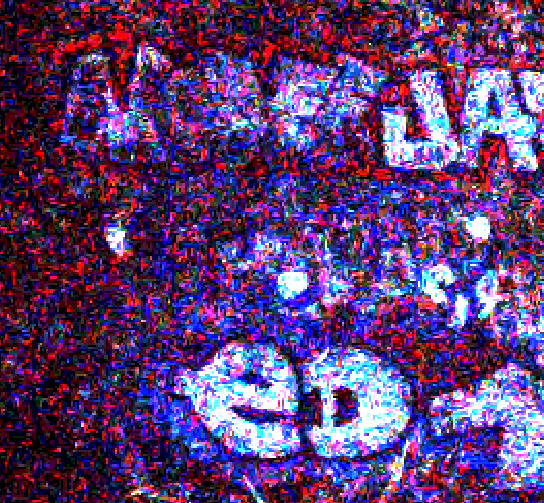}&
\includegraphics[trim=0 0 0 0,clip,width=0.12\textwidth]{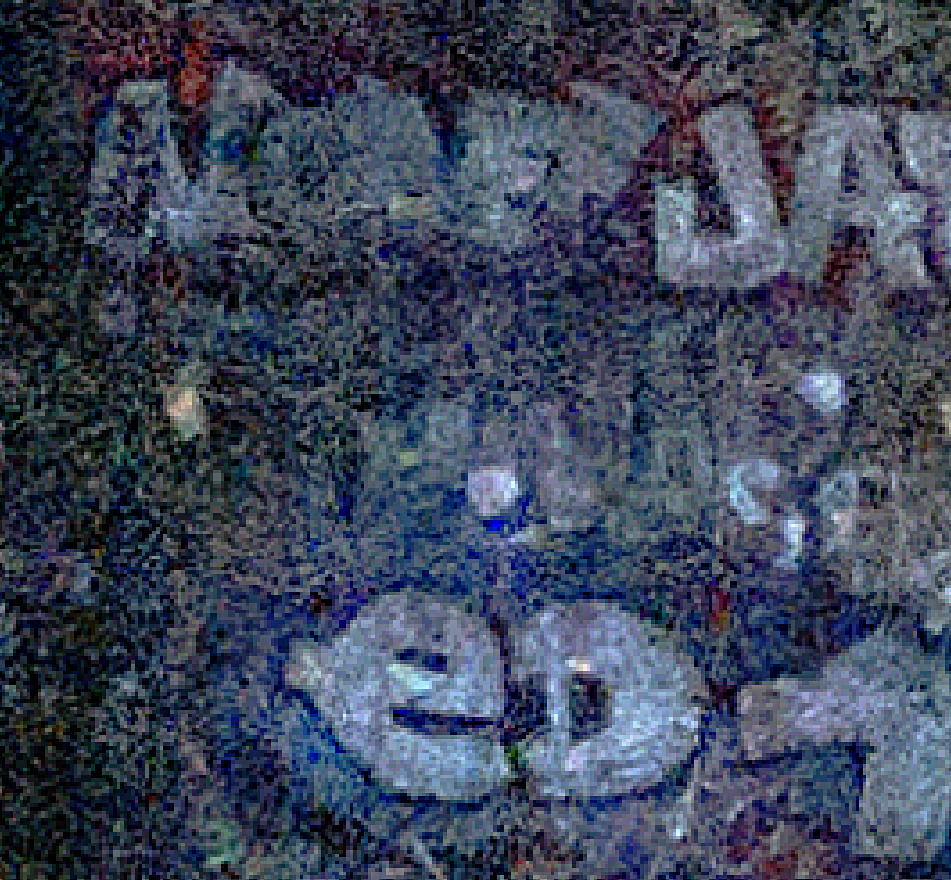}&
\includegraphics[trim=0 0 0 0,clip,width=0.12\textwidth]{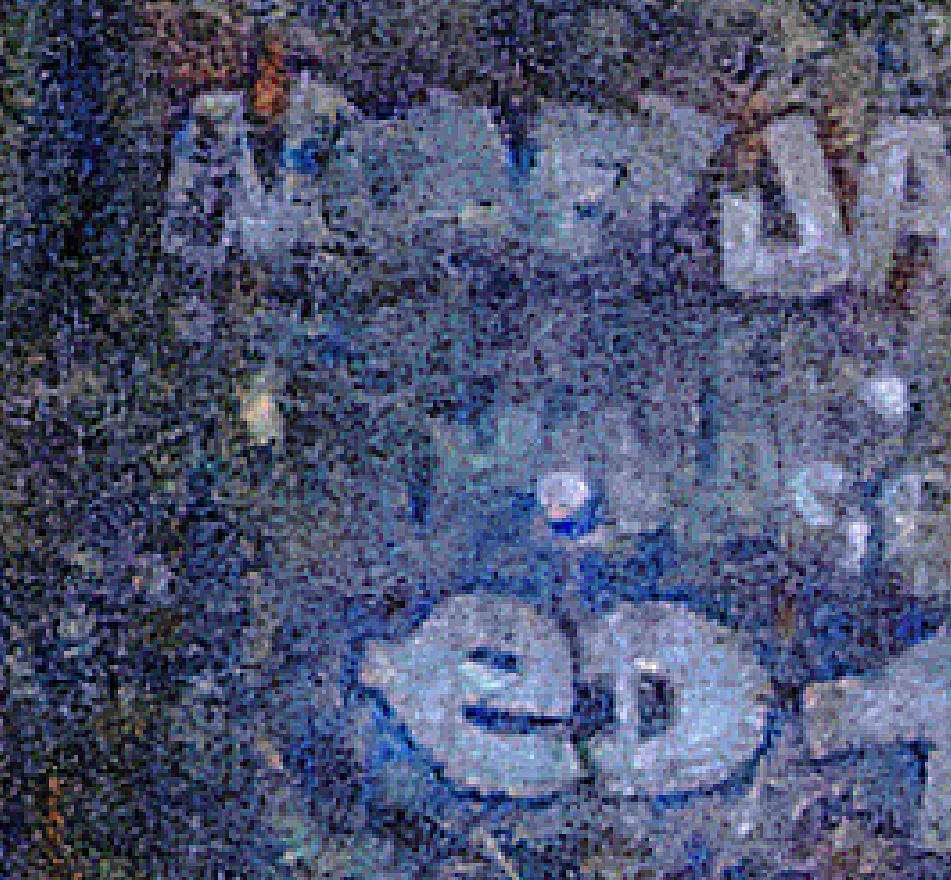}&
\includegraphics[trim=0 0 0 0,clip,width=0.12\textwidth]{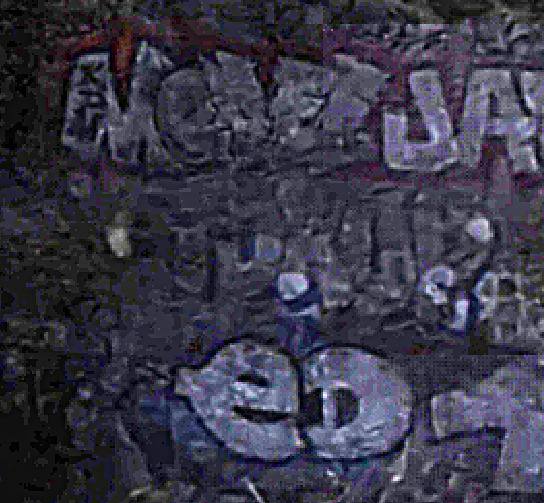}\\
\includegraphics[trim=0 0 0 0,clip,width=0.12\textwidth]{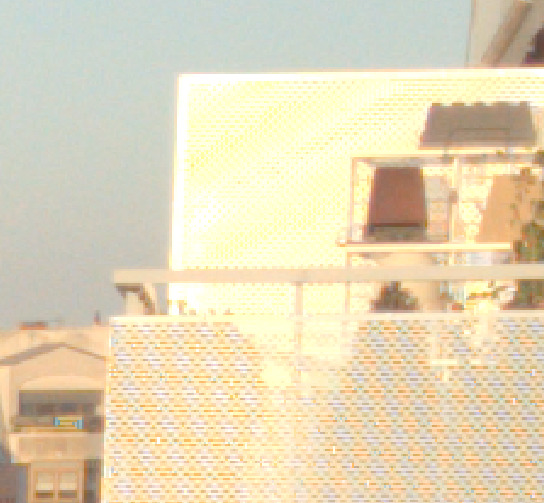}&
\includegraphics[trim=0 0 0 0,clip,width=0.12\textwidth]{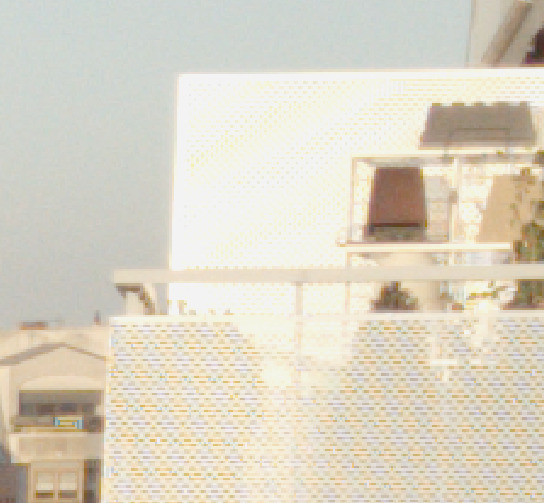}&
\includegraphics[trim=0 0 0 0,clip,width=0.12\textwidth]{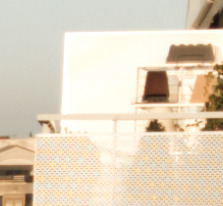}&
\includegraphics[trim=0 0 0 0,clip,width=0.12\textwidth]{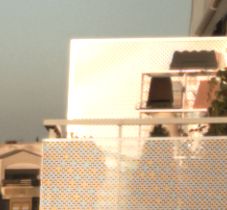}&
\includegraphics[trim=0 0 0 0,clip,width=0.12\textwidth]{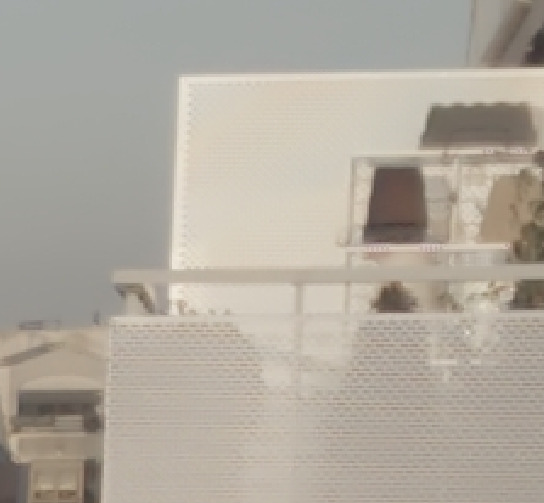}\\

\begin{tabular}{c} \cite{yan19attention} \\ (3 frames) \end{tabular} &\begin{tabular}{c}  \cite{wu18deep}\\ (3 frames) \end{tabular} &  \begin{tabular}{c}  \cite{santos20single}  \\ (1 frame) \end{tabular} &
\begin{tabular}{c}  \cite{liu20single} \\ (1 frame) \end{tabular}& 
\begin{tabular}{c}   Ours\\ (3 frames, no SR) \end{tabular}
\vspace{0.5cm}

\end{tabular}

\\

\begin{tabular}{ccc}
\includegraphics[height=0.044\textwidth,trim=0 0 0 30]{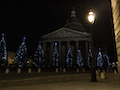} & 
\includegraphics[height=0.044\textwidth,trim=0 0 0 30]{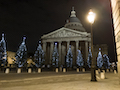} &
\includegraphics[height=0.044\textwidth,trim=0 0 0 30]{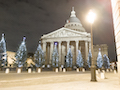} \\ 
\multicolumn{3}{c}{\includegraphics[height=0.2\textwidth]{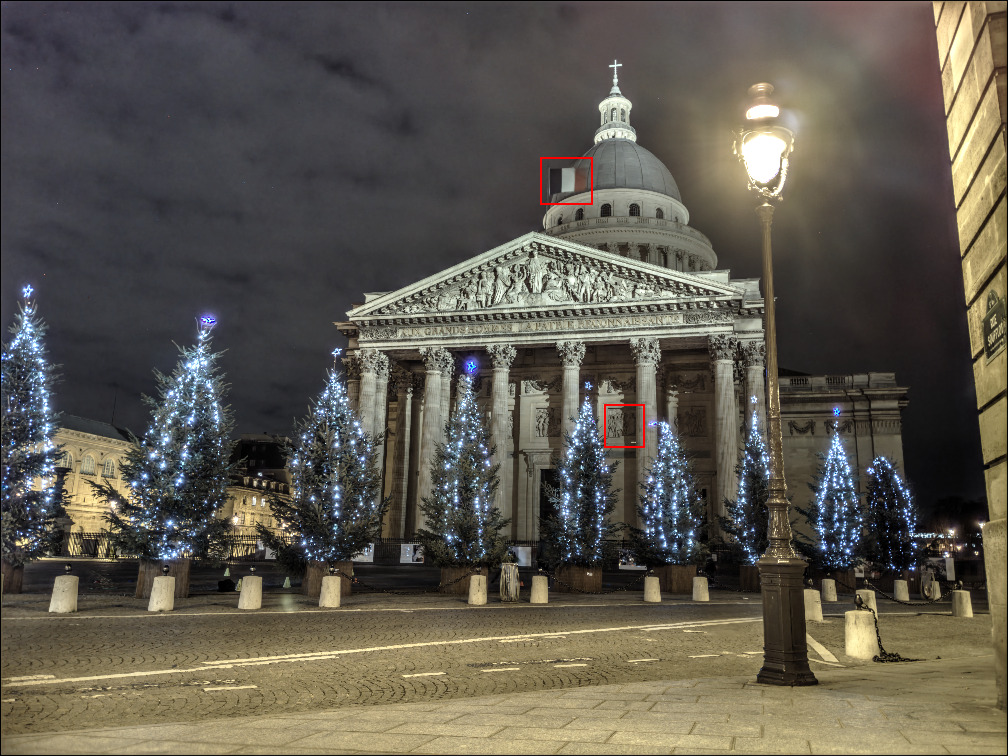}}
\end{tabular}

\begin{tabular}{ccccc}
\includegraphics[trim=0 50 0 25,clip,width=0.12\textwidth]{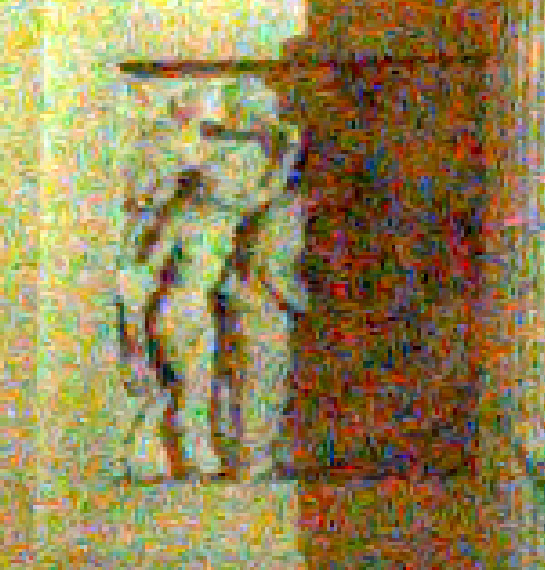}&
\includegraphics[trim=0 50 0 25,clip,width=0.12\textwidth]{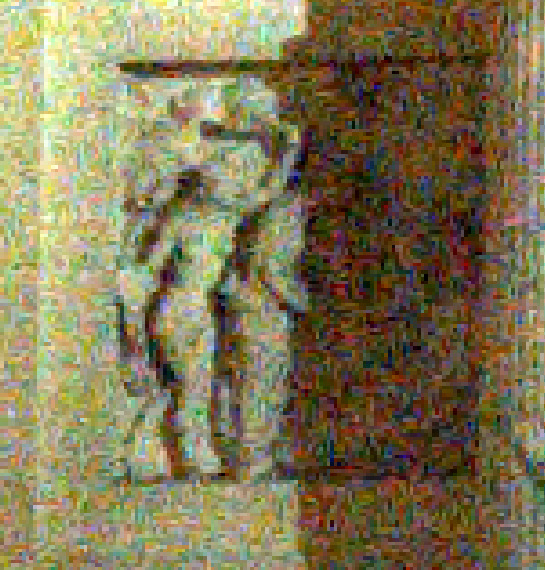}&
\includegraphics[trim=0 13 0 7,clip,width=0.12\textwidth]{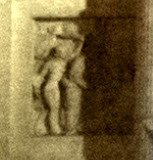}&
\includegraphics[trim=0 0 0 3,clip,width=0.12\textwidth]{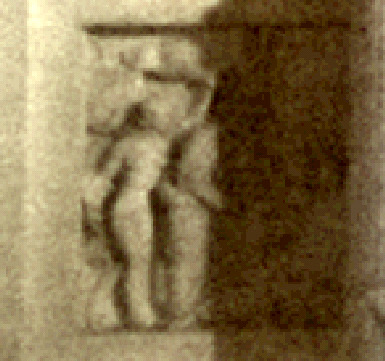}&
\includegraphics[trim=0 50 0 25,clip,width=0.12\textwidth]{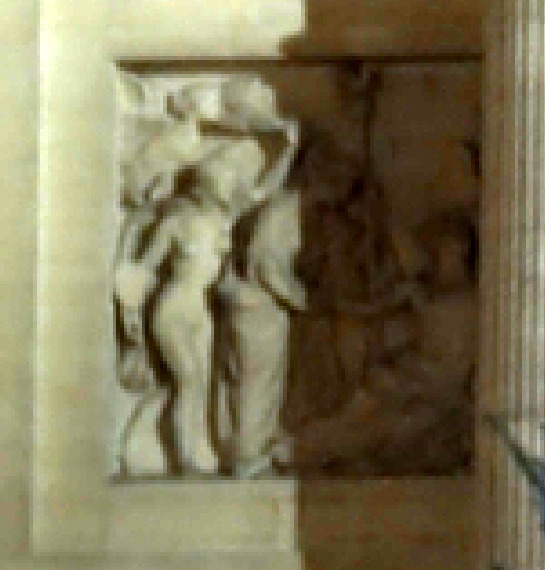}\\
\includegraphics[trim=0 50 0 25,clip,width=0.12\textwidth]{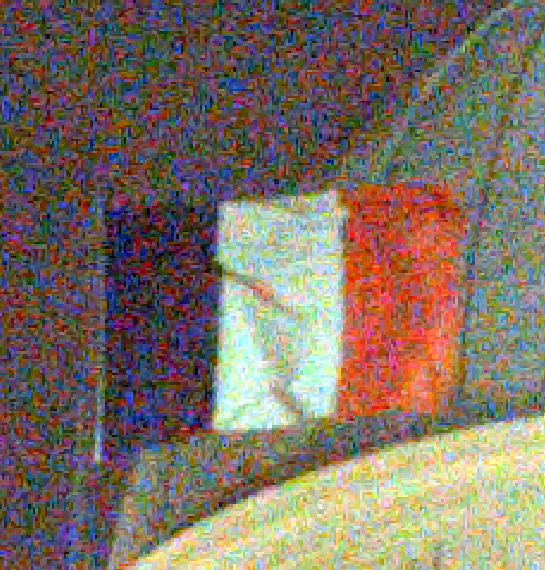}&
\includegraphics[trim=0 50 0 25,clip,width=0.12\textwidth]{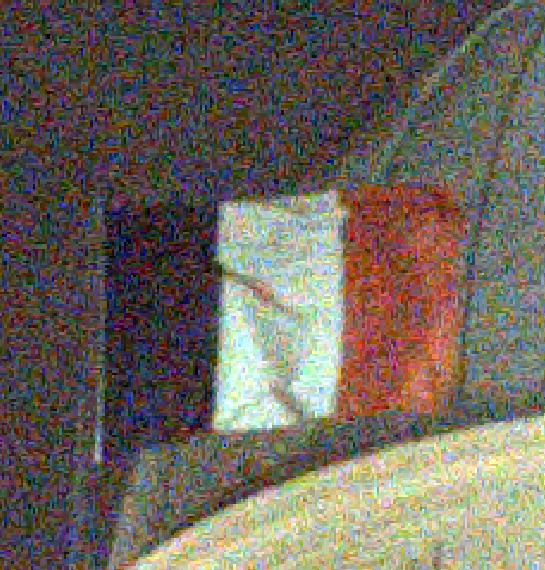}&
\includegraphics[trim=0 18 0 10,clip,width=0.12\textwidth]{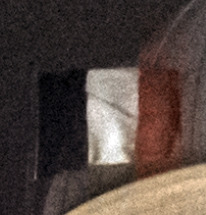}&
\includegraphics[trim=0 0 0 0,clip,width=0.12\textwidth]{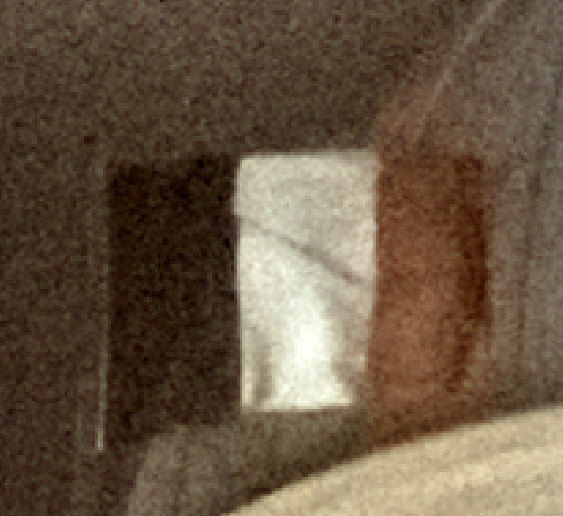}&
\includegraphics[trim=0 50 0 25,clip,width=0.12\textwidth]{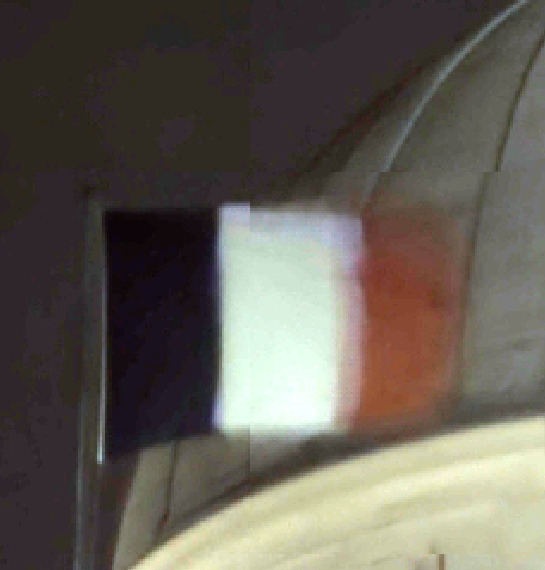}\\
\begin{tabular}{c} \cite{yan19attention} \\ (3 frames) \end{tabular} &\begin{tabular}{c}  \cite{wu18deep}\\ (3 frames) \end{tabular} &  \begin{tabular}{c}  \cite{santos20single}  \\ (1 frame) \end{tabular} &
\begin{tabular}{c}  \cite{liu20single} \\ (1 frame) \end{tabular}&
\begin{tabular}{c}   Ours\\ (3 frames, no SR) \end{tabular}
\end{tabular}

\end{tabular}
}
\caption{Comparison with CNN-based HDR methods processing one to three input frames. {\bfseries Left:} A sequence of three input frames, followed by our result after tone mapping, for two scenes. {\bfseries Right:} Small crops from the scenes obtained by various methods. To be fair, we compare them with a version of our model that does not perform super-resolution ($\times 1$ upscaling factor) and only processes a burst of 3 images, in the EV range [-2.4,0,2.4]. We observe that, in well-exposed regions, the reconstruction performances of the three methods are similar. Our method appears to be more robust to noise, but more sensitive to non-rigid motion as shown in the case of the flag.
}
\label{fig:cnncomparisonnight}
\end{figure*}

\subsection{\red{Pure HDR imaging}}\label{subsec:hdr}
We evaluate the ability of \red{our approach to align and merge
raw images into HDR image at the same resolution as the input.}

We compare our approach with a bracketing technique, implemented
with the weights of~\cite{hasinoff10noiseoptimal}, two
state-of-the-art CNNs~\cite{wu18deep, yan19attention} trained
to predict a 32-bit image from only three LDR images with
-2, 0 and +2EV or -3, 0 and +3EV, and recent single-image HDR
CNNs~\cite{santos20single, liu20single}.
We generate 266 raw bursts with 32-bit ground-truth images, 
each burst containing 11 synthetic raw images
with small random
shifts and rotations and Poissonian-Gaussian noise with
parameters $\alpha$ and $\beta$ 
selected according the distribution in 
Figure~\ref{fig:isoplot}.  More details about data generation can be in found in Section \ref{sec:implem} of the appendix.
To evaluate the CNNs trained on RGB images, 
we first
pick the three raw frames corresponding
to \{-2.4,0,+2.4\} EV in the burst and demosaick 
them with the approach of \citet{malvar04high}.
We also demosaick the frames before merging the HDR images with 
the bracketing technique. If the raw frames are not aligned, 
after demosaicking, we align the frames either with the phase correlation algorithm \red{\cite{tursun16deghosting}
on the MTB features \citet{ward03fast}}
or with our Lucas-Kanade-based registration technique.
For fairness with the CNNs, we compare our approach when there are only 
three frames in the bracket (the same as for the CNNs) and with the whole
burst.

We present in Table~\ref{tab:raw_hdr_fusion} the results of our comparison.
We evaluate the PSNR and 
the SSIM metrics on both the output of each algorithm and after
evaluating the irradiance maps with $\mu$-law, 
playing the role of a tone mapping algorithm~\cite{kalantari17deep}.
However these typical image processing
metrics may not be
adapted to HDR imaging~\cite{aydin08metrics, eilertsen21how} we thus also
report the HDRVDP2 perceptual quality score
of \citet{narwaria15hdrvdp} (version 2.2.2).
Note that \citet{wu18deep} and \citet{yan19attention} use RGB images for training, while our method leverages more information by directly processing raw frames.
We also compare our method to
the single-image methods of \citet{santos20single}
and \citet{liu20single} running on the central
frame of the burst.

Our algorithm using 11 frames achieves the best results as
expected, with HDRVDP2 margins ranging from +4 to +9 over
recent CNN-based methods and of +4 over the bracketing technique of 
\citet{hasinoff10noiseoptimal} using 11 frames too.
The gap with CNNs comes from our ability to restore
the darker areas in raw photographs 
containing large read noise whereas these networks
are trained on RGB images only.
Figure~\ref{fig:cnncomparisonnight}
shows qualitative comparisons with the baselines in Table~\ref{tab:raw_hdr_fusion} for bursts
of 21 images taking during day time and night time.
Our method achieves the best visual results in
both dark and saturated areas. Note that the CNN baselines considered here have been designed to
handle 1 or 3 images only, which is not sufficient to achieve effective denoising through image fusion in challenging settings.

We also compare our approach with a public-domain  implementation~\cite{monod2021analysis}  of
Google's HDR+~\cite{hasinoff16burst} that addresses HDR imaging by
fusing images with the same exposure. In this setting, HDR essentially boils
down to burst denoising, which is effectively handled by our approach.
Figure~\ref{fig:hdrplus} shows
a qualitative comparison of HDR+ with
our technique. We achieve better denoising, especially in the darkest areas, while also increasing 
spatial resolution.

\begin{table*}[t]
    \caption{Quantitative comparison of various algorithms for HDR imaging -- we do not perform super-resolution in this experiment -- on a synthetic dataset consisting of bracketed raw bursts simulated with our pipeline. Our method directly takes raw frames as an input. The other methods process RGB frames obtained here with VNG demosaicking.
    Our algorithm quantitatively outperforms the other HDR methods on this dataset, which is not surprising as it is trained leverage the information lost in the raw to rgb conversion. 
    }
\small
    \centering
    \begin{tabular}{lcccccc} \toprule
        Method & PSNR~(dB) & $\mu$-PSNR~(dB) & SSIM~(\%)  &$\mu$-SSIM~(\%)& HDR-VDP2~(Q)   \\

        \midrule
        \quad \texttt{K=1 frames}\\ \midrule
        \cite{liu20single} & \underline{20.11} & \underline{24.42} &  \underline{0.611}& \underline{0.690} & \underline{57.32}\\
        \cite{santos20single} & \textbf{22.14} & \textbf{ 25.85} &\textbf{ 0.641} &\textbf{ 0.702 }& \textbf{62.94} \\ 

        \midrule
        \quad \texttt{K=3 frames}\\ \midrule
        \cite{hasinoff10noiseoptimal}~+~MTB & \underline{28.08} & \underline{29.46} & \underline{0.819} & \underline{0.847} & 61.13 \\
        \cite{hasinoff10noiseoptimal}~+~PLK & 27.25 & 28.69  & 0.814& 0.836 &  60.82  \\
        \cite{wu18deep} & 26.47 & 27.61 & 0.771 & 0.782 & \underline{61.80 } \\ 
        \cite{yan19attention} & 26.31 & 27.11 & 0.761 & 0.774 & 61.14  \\

        \textbf{Ours} &\textbf{ 33.75} &\textbf{ 34.39 }& \textbf{0.942} &\textbf{ 0.943} & \textbf{63.24 } \\ 
        \midrule
        \quad \texttt{K=11 frames}\\ \midrule
        \cite{hasinoff10noiseoptimal}~+~MTB & \underline{ 29.54} &   \underline{30.96}  &  \underline{0.862} & \underline{0.892} &\underline{62.07}\\
        \cite{hasinoff10noiseoptimal}~+~PLK & 28.80& 30.21 &  \underline{0.862}  & 0.888 &61.95 \\
        \textbf{Ours} &\textbf{37.83} & \textbf{39.22} &\textbf{ 0.964} &\textbf{0.971} & \textbf{65.44} \\ 
        \bottomrule
    \end{tabular}

    \label{tab:raw_hdr_fusion}
\end{table*}

\subsection{Multi-exposure registration}\label{subsec:registr}
We evaluate the performance of our registration module based on learnable
features. 
We measure the \red{ geometric alignment error} between the ground-truth motion
and predicted one~\cite{sanchez2016inverse} computing the Euclidean distance
between the aligned image corners with that of the ground-truth ones
and is counted in number of pixels in the HR image.

We report the mean and the median
over 266 validation bursts (containing 11 images per burst) 
synthesized with the same protocol as for generating the training data.
We compare a typical multi-exposure registration scheme consisting
in combining MTB features and phase correlation~\cite{tursun16deghosting}
(used for prealigning the images in our model),
with the 3 iterations of the pyramid Lucas-Kanade (PLK) algorithm over plain pixels and deep features 
learnt in an end-to-end manner.
The three methods are run on the mosaicked and possibly noisy 
images, prior to any ISP processing.
We evaluate this panel over three scenarios: (i) HDR generation without 
SR from noise-free raw bursts, (ii) HDR generation without SR from raw bursts
with noise and (iii) joint HDR and SR with factor $\times4$ from raw bursts 
with noise.

Table~\ref{tab:multi_exposure_registration} shows that, in all cases, 
our approach achieves the best quantitative
results, with a margin ranging from 0.5px for the (unrealistic) noise-free
benchmark to more than 1px for the more challenging ones featuring noise.
Interestingly, using more iterations does not always mean a better alignment. 
A plausible explanation is that our model is trained for using three iterations of the LK algorithm, and may be sub-optimal for more iterations. 

We have also empirically observed that the errors in this table
are always greater than that reported by \citet{lecouat21aliasing}
in their work for aligning frames with the same exposure.
This gap is caused in practice by the darkest and brightest
frames, much harder to align because of the 
noise \red{in dark regions} and large saturated areas.

\begin{figure}[t] 
    \setlength\tabcolsep{2pt}
    \renewcommand{\arraystretch}{0.5}
    \centering
    \begin{tabular}{cc}
 \includegraphics[trim=40 150 0 100,clip,width=0.22\textwidth]{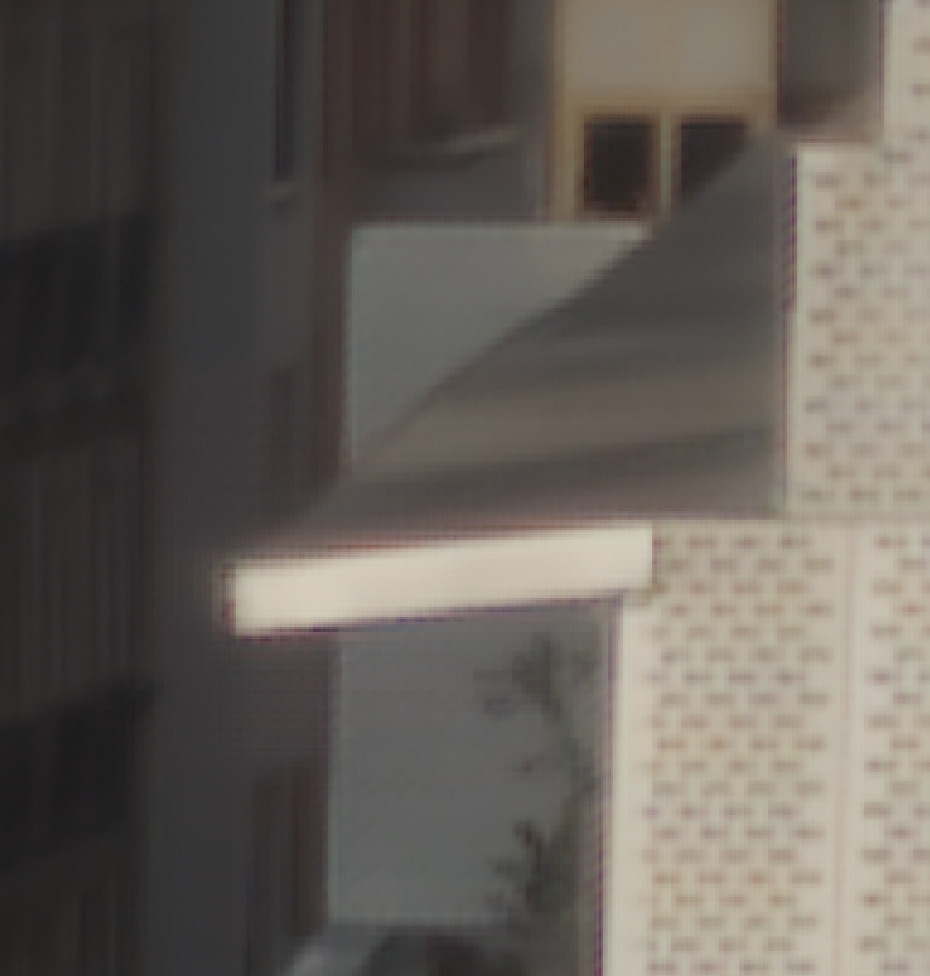}        &
 \includegraphics[trim=40 150 0 100,clip,width=0.22\textwidth]{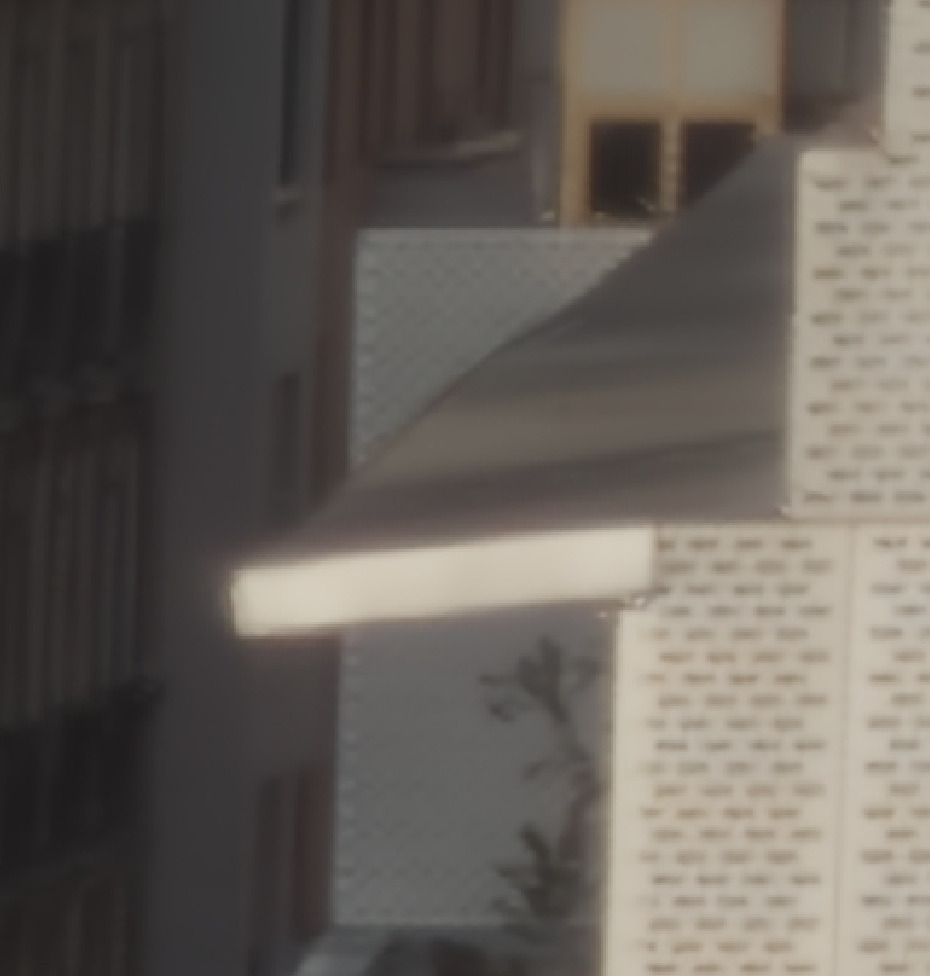}  \\
 LK & LK, with deep features 
    \end{tabular}
    \caption{Qualitative comparison of the reconstructed
    image with a pyramid of Lucas-Kanade run on plain pixels or deep features.
    Note the zipping artefacts along the edges of the large white rectangle. Our learnable variant is faster and
    leads to more accurate results. The reader is invited to zoom in.}
    \label{figure:deepfeatures}
\end{figure}

Figure~\ref{figure:deepfeatures} \red{compares} the advantage of running
the Lucas-Kanade \red{algorithm with deep features and plain pixels in a real
situation}. Note the purple zipping artefacts caused 
by faulty alignment before image fusion in the left
image obtained with the plain-pixel Lucas-Kanade algorithm.
These artefacts vanish in the image on the right
using deep features.

\begin{table}[]
    \caption{Quantitative comparison of registration methods on synthetic data
    with average and median geometric errors~\cite{sanchez2016inverse}.
    We compare MTB features combined with sub-pixelic 
    phase correlation~\cite{tursun16deghosting},
    the pyramid Lucas-Kanade (PLK) algorithm and our variant of PLK using
    deep features. The three algorithms are run on the mosaicked images.
    On each benchmark, we outperform both vanilla PLK and the MTB-based 
    approach.
    }
    \small
    \centering
    \begin{tabular}{lcc} \toprule
        Model  & Geom (avg.) & Geom (med.) \\ \midrule 
        \multicolumn{3}{c}{\texttt{$\times 1$ - No noise - 11 raw frames}}\\\midrule
        MTB + phase correlation  & 2.93 & 2.61   \\ 
        3 PLK iterations & 1.32  & 0.97  \\
        3 PLK iteration +deep features (ours) & \underline{0.91} & \textbf{0.60} \\ 
        5 PLK iterations &  1.47 & 1.10   \\ 
        5 PLK iterations + deep features (ours) & \textbf{0.88} & \underline{0.61}  \\ 
        \midrule 
        \multicolumn{3}{c}{\texttt{$\times 1$ - Noise - 11 raw frames }}\\         \midrule 
        MTB + Phase correlation  & 3.58 & 2.99  \\ 
        3 PLK iterations & 2.77 & 2.40  \\
        3 PLK iteration +deep features (ours) & \textbf{1.25} & \textbf{0.95} \\ 
        5 PLK iterations & 2.76 & 2.10  \\ 
        5 PLK iterations + deep features (ours)& \underline{1.40} & \underline{1.00}  \\\midrule 

        \multicolumn{3}{c}{\texttt{$\times 4$ (aliasing) - Noise - 11 raw frames }}\\         \midrule 
        MTB + Phase correlation  & 5.93 & 4.67 \\
        3 PLK iterations & 3.82 & 3.58  \\ 
        3 PLK iteration +deep features (ours) & \textbf{2.04} & \textbf{2.03} \\ 
        5 PLK iterations &  3.87 & 3.50  \\
        5 PLK iterations + deep features (ours) & \underline{2.62} & \underline{2.17}   \\ 
        \bottomrule
    \end{tabular}

    \label{tab:multi_exposure_registration}
\end{table}

\subsection{Discussion}\label{subsec:discussions}

\begin{figure}[ht]
    \setlength\tabcolsep{2pt}
    \renewcommand{\arraystretch}{0.5}
    \centering
    \begin{tabular}{cc}
 \includegraphics[trim=250 340 250 340,clip,width=0.22\textwidth]{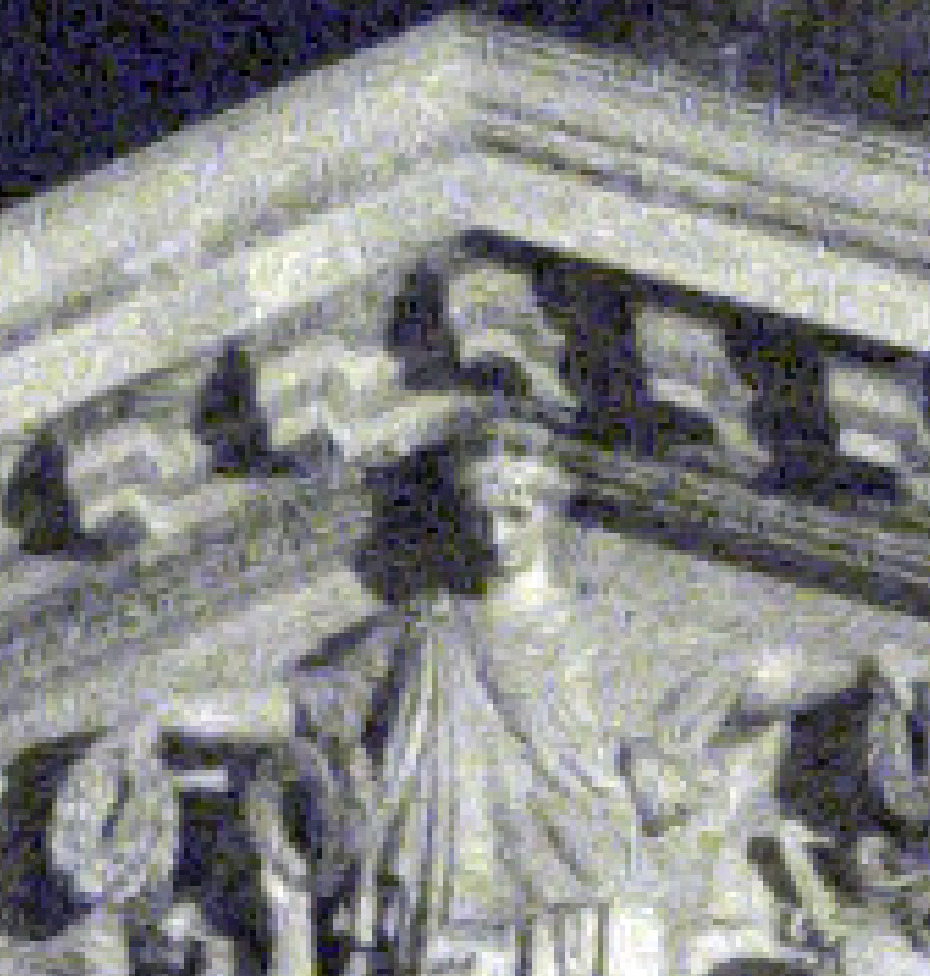}        & 
 \includegraphics[trim=250 340 250 340,clip,width=0.22\textwidth]{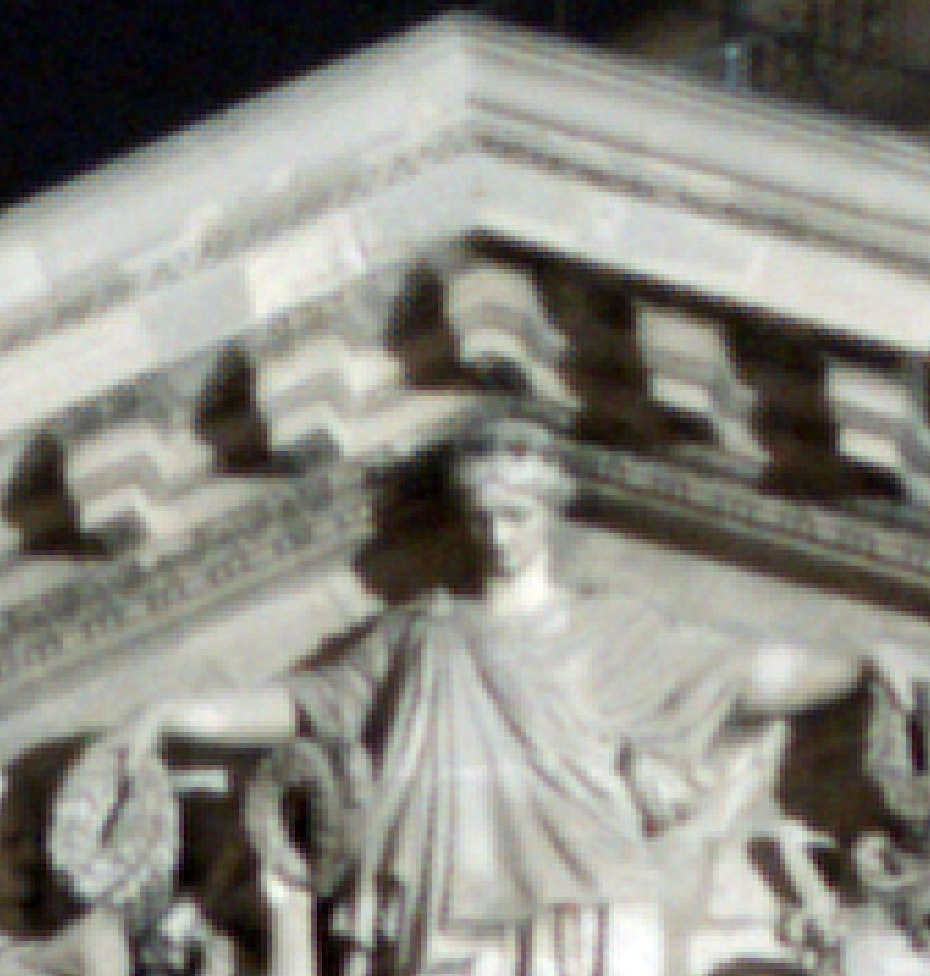}        \\
 Low-resolution & No prior \\ 
 \includegraphics[trim=250 340 250 340,clip,width=0.22\textwidth]{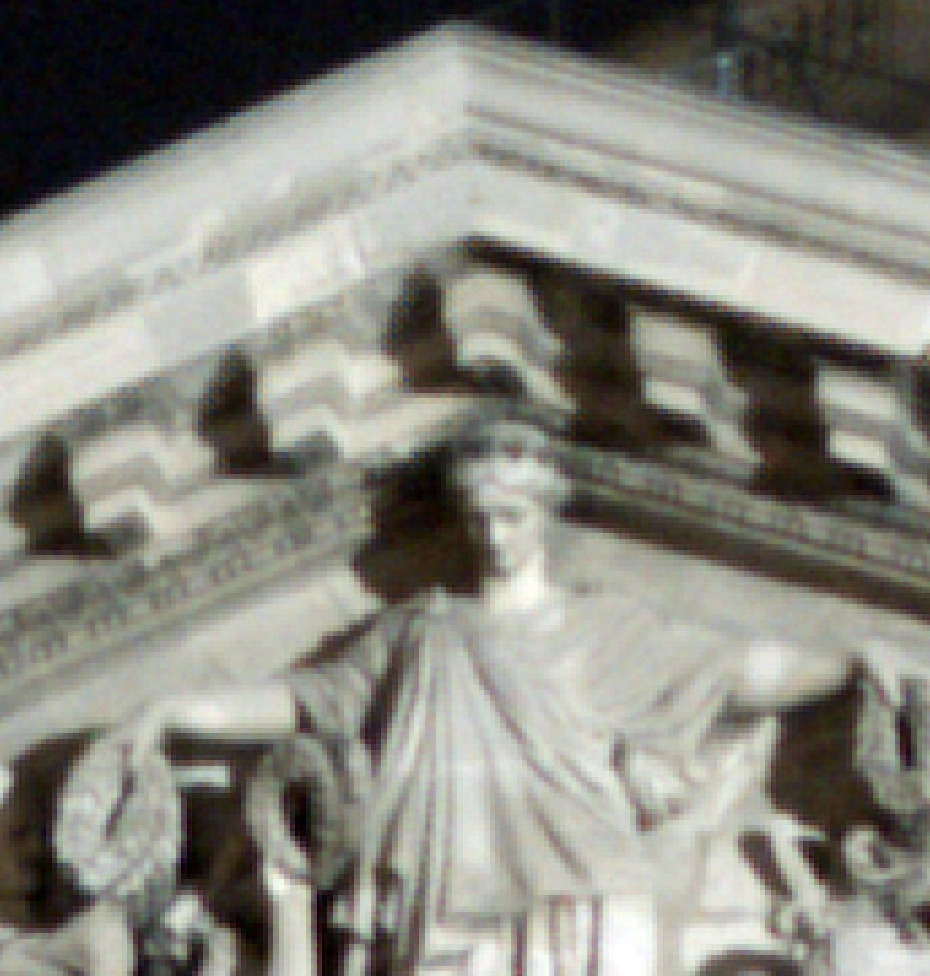}        & 
 \includegraphics[trim=250 340 250 340,clip,width=0.22\textwidth]{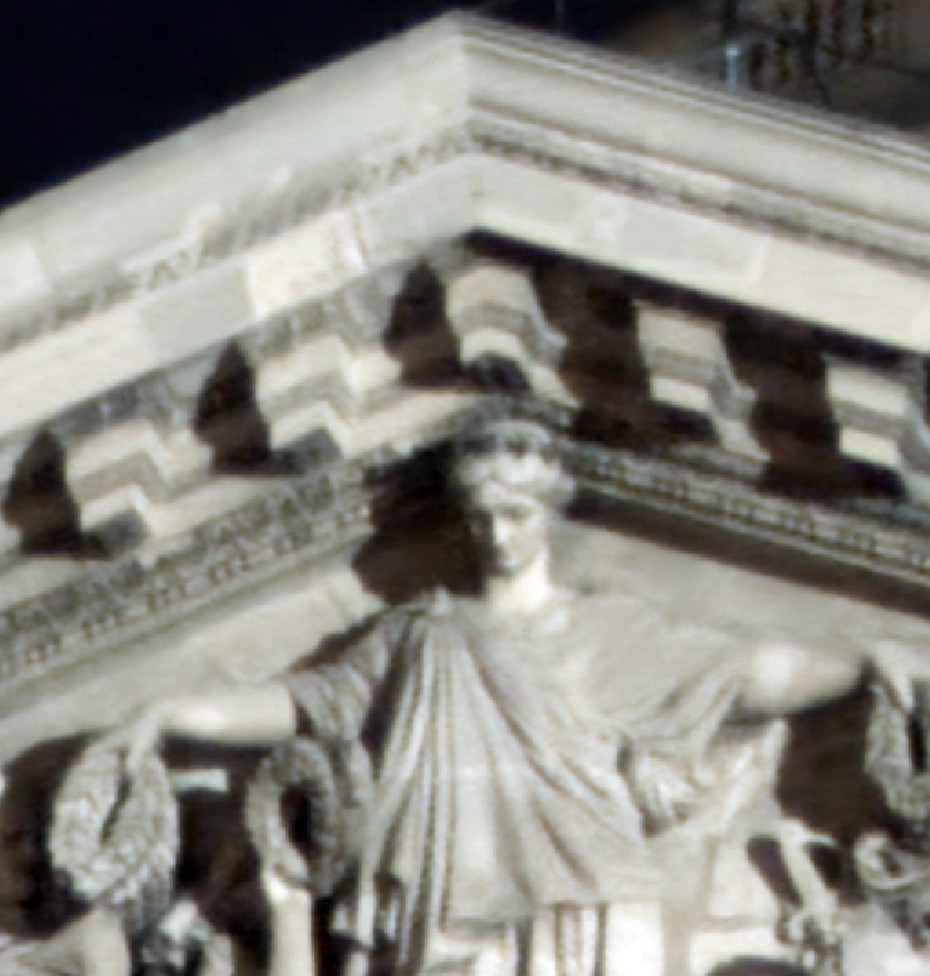}        \\
TV-$\ell_1$ & Ours \\
    \end{tabular}
    \caption{Visual comparison of the impact of the prior for joint HDR
    and $\times 4$ SR. We fuse $K=20$ images in this example.
    We note for the three methods effectively suppress the noise present in the original LR frames. However, our learnable prior (here with 300k parameters)
    yields a higher quality image. The reader is invited to zoom in.}
    \label{fig:priorcomparison}
\end{figure}

\paragraph{Choice of the prior function.}
An important component of our approach is the image 
proximal operator $G_\omega$.
Figure~\ref{fig:priorcomparison} shows a qualitative comparison of 
a prior-free
version, solely aligning and merging the frames, using
the image gradients soft-thresholding 
function derived from the classical TV-$\ell_1$ prior, 
and our approach with a learnable module.
The TV-based version is \red{significantly} sharper than
that the one without prior. The parametric prior returns a 
better zoomed-in image, \eg next to the head and the dress of the statue.

\paragraph{Robustness on real images.} 
A key \red{advantage} of our approach is
the accuracy of its registration module, as detailed on Table~\ref{tab:multi_exposure_registration} and illustrated in
Figure~\ref{figure:deepfeatures}.
We have remarked that this module is particularly
efficient for aligning raw frames with the same exposure, as illustrated
by Figure~\ref{fig:SRrealburst} in the context of SR with factor
$\times 4$. Given a burst of raw photographs including
moving objects with reasonable motion during exposure, \eg
the pedestrian in the figure, we can predict \red{high-quality} HR image 
well-aligned with the reference frame whereas the
competitors
may introduce ghosting or colored artefacts.
Notwithstanding, we have also noted that non-rigid motions in the raw frame burst may
lead to blur in the final predicted image. For instance, 
Figure~\ref{fig:cnncomparisonnight} compares the restoration results
from \citet{wu18deep}, \citet{yan19attention} and our model for a 
crop \red{featuring} a waving flag, \ie a non-rigid motion.
We select $K=3$ images with EV values of \{-2.4,0,2.4\} EV
for the CNNs and for our model.
The CNNs trained to remove ghosting artefacts accurately align
the flag with the reference frame 
whereas our prediction is blurry in the red section of the flag.
This \red{may stem from} the fact that multi-exposure image
registration is a very challenging problem and that 
we have not such non-rigid motions in our training data.
\red{
For a better deghosting, the injection of non-rigid motion in the training data, similarly to the dataset introduced in \cite{kalantari17deep}, is an interesting future research direction.}

\section{Conclusion}
We have introduced an effective algorithm for the reconstruction of high-resolution, high-dynamic range color images from raw photographic bursts with exposure bracketing. We have demonstrated its excellent performance with super-resolution factors of up to $\times 4$ on real photographs taken in the wild with hand-held cameras, and high robustness to low-light conditions, noise, camera shake, and moderate object motion. We have also shown that it compares favorably to the state of the art in both the HDR imaging and super-resolution domains. The keys to the success of this algorithm are the powerful underlying image formation model, a hybrid approach for solving the corresponding inverse problem that combines the interpretability and generalization power of model-based techniques with the flexibility and robustness of (deep) machine learning
technology, and new twists on classical alignment techniques that make them
robust by design. The proposed approach also has limitations: In particular its
performance in saturated regions is far from perfect, and it is only robust to relatively
small scene motions. Future work will include improving these aspects,
but also adapting the method to large camera displacements (wide-baseline setting)
and scene motions, which in turn will lead to 3D capture applications. A missing
piece is of course handling motion and defocus blur, and we also plan to adapt
our method to a new range of applications such as focus stacking, where image
alignment requires comparing images  \red{with different focuses} that may be even more different than
those obtained under changing exposures. Finally, we are also interested in
scientific applications in astronomy, microscopy, and remote sensing.

\begin{acks}

This work was funded in part by the French government under management of Agence Nationale
de la Recherche as part of the ``Investissements d'avenir'' program, reference ANR-19-P3IA-0001
(PRAIRIE 3IA Institute). JM and BL were supported by the ERC grant number 714381 (SOLARIS
project) and by ANR 3IA MIAI@Grenoble Alpes (ANR-19-P3IA-0003). JP was supported in part by
the Louis Vuitton/ENS chair in artificial intelligence and the Inria/NYU collaboration.
This work was granted access to the HPC resources of IDRIS under the allocation 2022-AD011011252R2 made by GENCI. This work was done while TE was a PhD student at Inria.
\end{acks}

\bibliographystyle{ACM-Reference-Format}
\bibliography{bibliography.bib}

\clearpage
\appendix
\onecolumn

\section{Ablation studies}

In this section, we provide additional experiments to better understand the impact of our method's components.

\subsection{Learning the image prior}
We quantitatively validate the choice of a parametric proximal function in Eq.~\eqref{eq:learntproximal}
to address joint HDR and SR by a factor of 4.
We compare the proposed CNN-based implementation with the 
the classical total-variation $\ell_1$ 
(TV-$\ell_1$), for instance used by~\citet{choi09high}, and a simple
weighted least-squares problem, \ie without prior.
We generate 266 bursts of $K=9$ images with exposure values
in $[-3,3]$EV.
Table~\ref{tab:priorcomparison} shows average PSNRs for methods embedding no
prior (simple weighted least-squares), TV-$\ell_1$ prior and three variants
of the CNN $G$ with parameter $\omega$ of sizes 30K, 300K and 3M.
The variants with the parametric priors are trained according to the protocol
described in the previous section.
This table shows, as expected, a clear advantage of learnable penalty
functions over handcrafted ones, with a margin of more than 4dB for the 
shallowest network and more than 7dB for the deepest one over the TV-$\ell_1$
variant.
Note that the prior-free version is only 0.4dB below its TV-$\ell_1$-based
counterpart, suggesting that machine 
learning is important to design an efficient
prior function to address joint HDR and SR restoration.

\begin{table}[h!]
    \caption{
Quantitative comparison of the choice of the prior over the total performance of the method.
Average PSNR on predicted linear HDR images jointly super-resolved by a factor of 4
for typical handcrafted image priors and variants of the proposed parametric one
with several parameter sizes.
The learnable ones achieve the best scores overall by an important margin.
The more parameters yields the best PSNRs.}
\begin{tabular}{@{}lcc@{}}
\toprule
Prior                    & PSNR  \\ \midrule
No prior &   26.15 \\
TV-$\ell_1$ & 26.51         \\
Tiny (30k) & 30.71 \\
Small Prior (300k) & \underline{32.56} \\
Large Prior (3M) &  \textbf{34.18} \\
\bottomrule
\end{tabular}

\label{tab:priorcomparison}
\end{table}

\subsection{Alignement sub-components evaluation}
We quantify the impact of each components of the alignment module in Table~\ref{tab:jointHDRSRsynthetic} by measuring
the mean PSNR of predicted HDR images with resolution enhanced by $\times4$. 
We generate 266 raw bursts with 11 frames for each burst with
the same protocol than the other experiments.
We decompose it into three bricks: using bracketed images, the confidence
function $g$ and running the pyramid Lucas-Kanade (PLK) algorithm on
deep features.
Adding each component one-by-one gradually increases the mean PSNR,
the maximum value being naturally reached when the three components
are gathered. Note that the PLK algorithm run on deep
features brings an improvement of about +2dB, which alone
is a better contribution than the total of +1.3dB by combining
bracketed images and the confidence function $g$.
We also give an upper-bound to this performance by running
a version of our model where we give the ground-truth motion
to align the images. Such an oracle model achieves an average
PSNR of 34.18dB compared to the 31.42dB of the best setting
where the motion is estimated instead.
It suggests that there is room to improvement but each sub-components
in the alignment actually helps to further narrow the gap with the 
oracle model.

\begin{table}[]
    \caption{
Ablation study for the alignment module. We report mean PSNR on HDR images with $\times$4 super-resolution. The first configuration (\#1) uses a burst but no bracketing (constant
exposure). The fourth
configuration (\#4) is the setting we use in practice, with bracketed exposures, the confidence function $g$ and the pyramid
Lucas-Kanade algorithm run on learnt features.
Adding these three components one by one  gradually improves the mean PSNR, showcasing the 
importance of each module. The fifth (\#5) configuration is an upper bound where we use the ground-truth
motion (and thus do not need LK with deep
deatures). 
}
\begin{tabular}{lccccc}
\toprule
Settings & \#1 & \#2 & \#3 & \#4 & \#5 \\
\midrule
Bracketing                   &        & \checkmark & \checkmark & \checkmark & \checkmark   \\
Confidence function $g$      &        &            & \checkmark & \checkmark & \checkmark   \\
LK with deep features        &        &            &            & \checkmark &              \\
Oracle motion                &        &            &            &            & \checkmark   \\
\midrule

PSNR  &                    28.50  & 29.28      & 29.77      & 31.42 & 34.18    \\

\bottomrule
\end{tabular}

\label{tab:jointHDRSRsynthetic}
\end{table}

\subsection{Performance with the number of frames}
\begin{figure}
    \centering
    \includegraphics[width=0.4\textwidth]{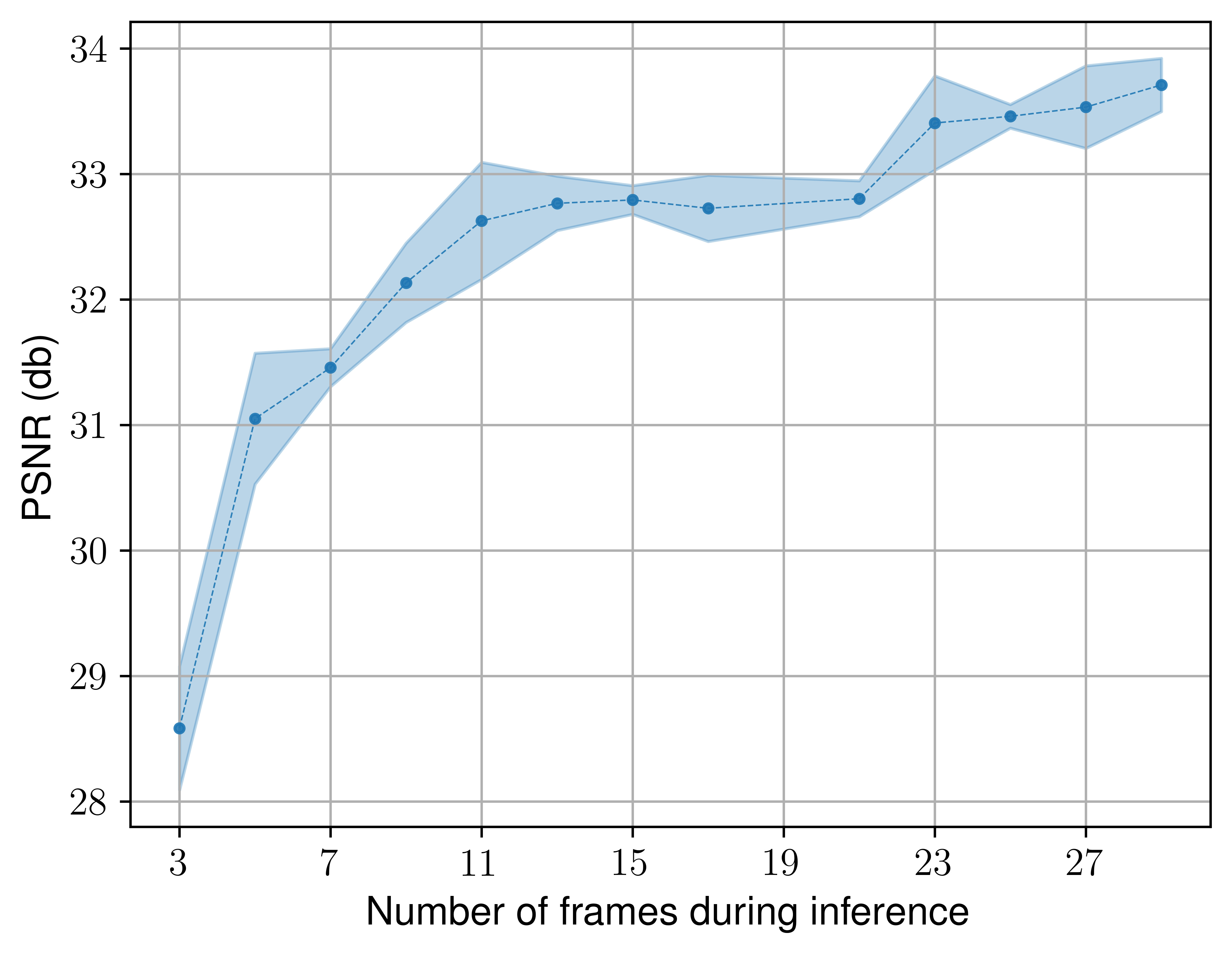}
    \caption{Average PSNR on predicted HDR images with spatial resolution increased by ($\times 4$),
    from a varying number of frames in the burst (from 3 to 30). Our approach
    benefits from any additional input frame, especially for less than $K=11$ images. 
    The average PSNR is evaluated from 3 seeds.}
    \label{fig:burstlength}
\end{figure}

We compare the performance of our approach with respect
to the number of frames in the burst for joint SR and HDR on
synthetic data.
Figure~\ref{fig:burstlength} shows the mean PSNR taken over 3
seeds for bursts of length ranging from 3 to 30.
Our model greatly benefits from additional frames for burst sizes
smaller than 11; Starting from 3 images and a PSNR score of 
28.6dB, we gain up to 4dB when accumulating 11 frames. 
Beyond this number, we gain an extra decibel by accumulating more 
than 20 frames.
It is consistent with typical bracketing techniques, \eg \cite{granados10optimal, hasinoff10noiseoptimal}, for which
more images means better noise removal in the dark regions.
Thanks to the learnt robust registration algorithm and prior, the
performance of our approach hardly falls down when accumulating
more and more images, unlike typical multi-image algorithms that
may accumulate registration, \eg as noted by~\citet{wronski19handheld}.

\subsection{Computational speed}\label{subsec:speed}
Our algorithm leverages optimization and machine learning techniques,
which leads to a dramatically smaller number of parameters than state-of-the-art
CNNs for tasks such as super-resolution. 
We evaluate the computational speed and memory consumption 
of our model embedding three variants of the learnable operator $G$ with 
varying number of parameters.
We compare our three versions of the proposed network with that of 
\cite{bhat21deep} and \cite{luo2021ebsr}, the best performers for
$\times 4$ SR in Table~\ref{tab:SR4synthetic}.
We run this five-way comparison within the same python environment, \eg same
version of Pytorch, and on
the same GPU (Nvidia Titan RTX) for fairness.
We show in Table~\ref{tab:computationalefficiency} that
our hybrid method exceeding the SR state of the art in the previous 
paragraphs, is also the lightest in the panel.
The method of \cite{luo2021ebsr} has 26 million parameters and that
of \cite{bhat21deep} about 13 million parameters whereas our deepest model
has 3 million of them, \ie four time less than \cite{bhat21deep}.
This gap in size of parameter is due to the building blocks in the
competitors' architectures. Indeed, they heavily rely on memory-greedy
attention modules, whereas our implementation of $G$ is based on the 
fully convolutional U-net architecture of~\cite{zhang2020deep, lecouat21aliasing}.
This table also shows that, for resolution factor of $\times 4$, 
our approach is much faster than the
state of the art, while coping with them according to
Table~\ref{tab:SR4synthetic}. In this table, our ``Small'' model
is less than a decibel below \citet{luo2021ebsr}'s model but with an
inference time forty times smaller on the same GPU.
We are also four times faster than \citet{bhat21deep}.
Likewise, our models require three to four times less GPU memory than our
selected competitors, which is an important designing point for deploying
such a technology in commercial software running on consumer-grade devices.
We also report in Table~\ref{tab:computationalefficiency} information about the $\times2$ case
since in many situations pushing the resolution further brings little improvement, 
\eg Figure~\ref{fig:SRlimit} and the analysis of \citet{wronski19handheld}.
In this configuration, our model requires even less memory to 
process $400\times400$ tiles and may run on modest GPUs.

\begin{table}[]
    \caption{Comparison of inference speed for different models for burst super-resolution. 
We have benchmarked the inference speed of different models for processing a 
burst of 14 12Mpixel raw images (Pixel 4a) on a single Titan RTX GPU. We have used the 
official implementations released by the authors without any modification. \red{We have not optimized inference speed (yet) using with standard tools such as mixed precision 
and/or model compression.}}
\small

\begin{tabular}{lrrrr}
\toprule
Model               & 
\#~parameters & 
Runtime & 
\begin{tabular}[c]{@{}l@{}}Memory\\ (200x200)\end{tabular} & 
\begin{tabular}[c]{@{}l@{}}Memory\\ (400x400)\end{tabular} \\ \bottomrule
Competitors~($\times4$) & & & & \\
\cite{bhat21deep}  & 13,000k        & 40.0sec   & 3.5Gb  & 11.5Gb \\
\cite{luo2021ebsr} & 26,000k        & 9.5min    & 3.5Gb  & 12Gb\\ 
\midrule
Ours~($\times4$) &  &  &  & \\
Very Small  & 60k     & 13.4sec & 1.2Gb & 2.8Gb \\
Small       & 250k    & 20.0sec & 1.2Gb &2.8Gb  \\
Large       & 3,000k  & 38.2sec & 1.3Gb & 3.1Gb \\
\midrule
Ours~($\times2$) & &  &  & \\
Very Small & 60k      & 4.7sec  & 800Mb & 1.2Gb  \\
Small      & 250k     & 6.2sec  & 820Mb & 1.2Gb \\
Large     & 3,000k    & 10.7sec & 860Mb & 1.3Gb \\
\bottomrule
\end{tabular}

\label{tab:computationalefficiency}
\end{table}

\subsection{Limitations}\label{subsec:limitations}
Albeit our approach favorably addresses HDR, SR and joint HDR and SR against the state of the art,
we have noted throughout our experiments a few limitations in certain cases that may degrade
the performance of our model.

\paragraph{Lack of robustness to non-rigide motion for joint HDR and SR.} We have observed that our model which performs joint HDR and super-resolution are less robust to non-rigid motion than our models performing only burst super-resolution. An example of artefacts that we typically get is shown in Figure~\ref{fig:cnncomparisonnight} in the case of the moving flag. 

\begin{figure}[t]
\small
    \setlength\tabcolsep{1.0pt}
    \renewcommand{\arraystretch}{0.5}
    \centering
    \begin{tabular}{cccc}
  \includegraphics[trim=0 40 0 0,clip,width=0.23\textwidth]{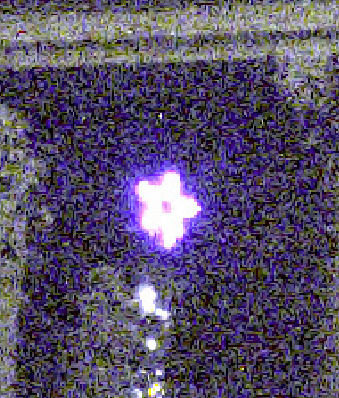}        &
  \includegraphics[trim=0 40 0 0,clip,width=0.23\textwidth]{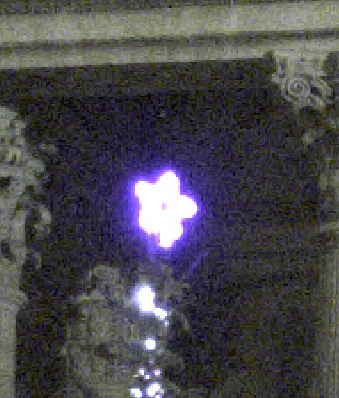}        &
  \includegraphics[trim=0 40 0 0,clip,width=0.23\textwidth]{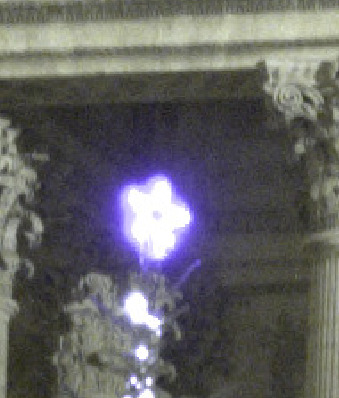}        &
    \includegraphics[trim=0 0 0 0,clip,width=0.23\textwidth]{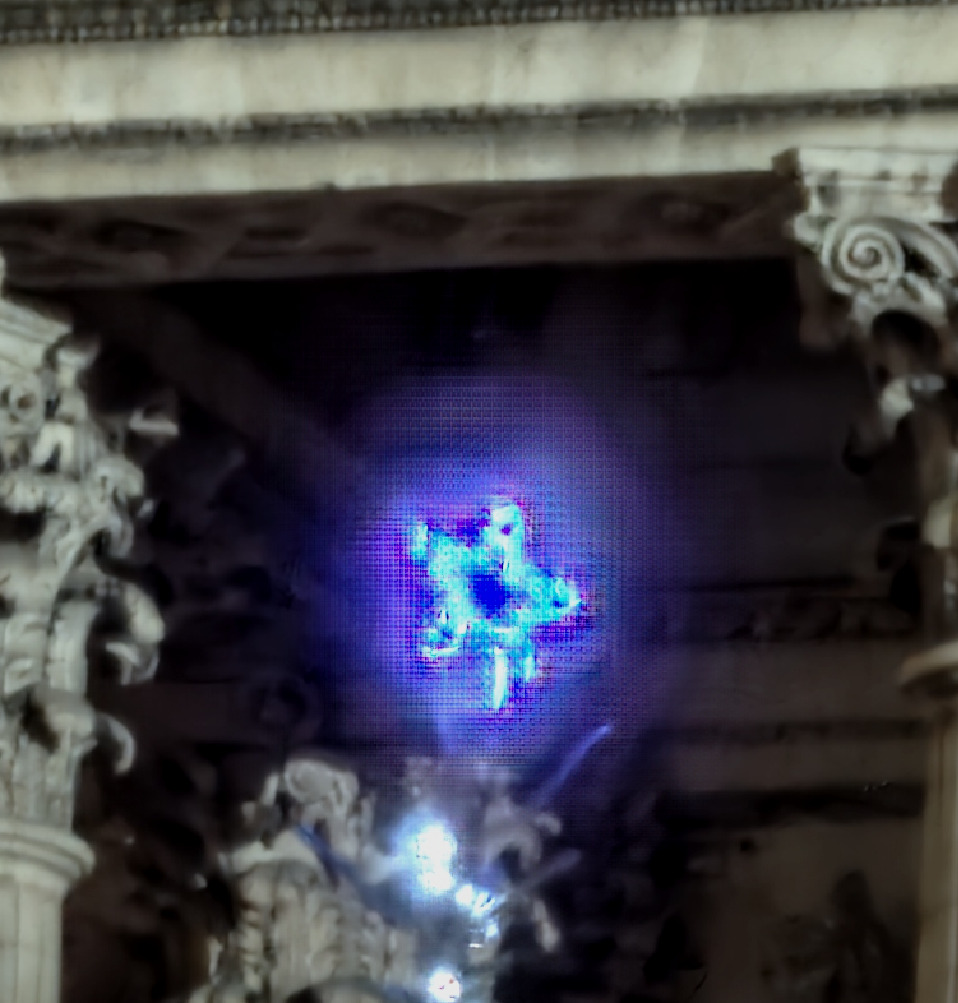}        \\
 Darkest frame & Medium frame & Lightest frame & Our reconstruction
    \end{tabular}
    \caption{Color halos in very bright areas and saturated regions. \red{Because all input views are saturated in the neighborhood of the star, key info required to correctly reconstruct the image there is missing, leading to severe artefacts there.}}
    \label{fig:blooming}
\end{figure}

\paragraph{Saturated areas.} In the pictures shot with a smartphone, we have sometimes noticed color halos next to saturated areas (Figure~\ref{fig:blooming}). They may be caused by the fact that the corresponding very high
level of contrast is hard to simulate  in our synthetic data.

\paragraph{Hot pixels.} The method is not trained 
to correct hot pixels, that may locally 
alter HDR imaging techniques. We assume that these pixels
are corrected upstream in the camera pipeline, which is a 
classical assumption in the field.

\subsection{Supervision with various loss functions}

\red{
In this subsection, we propose an ablation study to assess the effectivness of different supervision loss instead of the basic L1 loss. We use the same loss function as the one described in \cite{mildenhall2021nerf}, which gives a tone curve $\psi(x)=\log(x+\epsilon)$ which more strongly penalizes errors in dark regions. Results of the ablation study are presented in Table~\ref{tab:my_label}. Our experiment that the log loss gives better result in term of $\mu$-psnr. The selection of the right supervision loss for the training of our model is an interesting direction for future research.
}

\begin{table}[h!]
    \centering
    \begin{tabular}{lccc}
    \toprule
       loss  & psnr (dB) & $\mu$-psnr (\%) & $\mu$-ssim (\%)  \\
       \midrule
        $L_1$ &36.52 & 38.08 & 0.9682\\
        log($\epsilon=10^{-1}$) &37.04 & 37.86 & \textbf{0.964} \\
        log($\epsilon=10^{-2}$) & \textbf{37.71} &\textbf{38.70} & 0.967 \\
        log ($\epsilon=10^{-3}$) &13.74 &8.07 & 0.199\\
        \bottomrule
    \end{tabular}
    \caption{Super resolution factor x1, ablation study with different training loss.}
    \label{tab:my_label}
\end{table}

\subsection{Ablation HDR with no motion}

\begin{table*}[htb]
    \caption{\red{
        Quantitative comparison of various algorithms for HDR imaging – we do not perform super-resolution in this experiment – on a synthetic dataset consisting of bracketed raw bursts simulated with our pipeline. Our method directly takes raw frames as an input. The other methods process RGB frames obtained here with VNG demosaicking. Our algorithm quantitatively outperforms the other HDR methods on this dataset, which is not surprising as it is
        trained leverage the information lost in the raw to rgb conversion
        }}
\small
    \centering
    \begin{tabular}{lcccccc} \toprule
        Method & psnr (dB) & $\mu$-psnr (dB) & ssim  &$\mu$-ssim (dB)&HDR-VDP2(Q)   \\
        \midrule
        \quad \texttt{K=1 frame}\\ \midrule
        Liu {\em et al.}~\cite{liu20single} & 19.98 & 24.25 & 0.608 & 0.687 & 56.51 \\
        Santos {\em et al.}~\cite{santos20single} & 22.05 & 25.80 & 0.635 & 0.699 & {61.69} \\ 
        \midrule
        \quad \texttt{K=3 frames}\\ \midrule
        Wu {\em et al.}~\cite{wu18deep} & 26.42 & 27.51 & 0.765 & 0.774 & 61.04 \\ 
        Yan {\em et al.}~\cite{yan19attention} & 26.22 & 27.01 & 0.752 & 0.768 & {60.37} \\ 
        Hassinof {\em et al.}~\cite{hasinoff10noiseoptimal}& 30.55 & 31.26 & 0.874 & 0.878 &\underline{67.77} \\
        \textbf{Ours} & \textbf{34.29 }& \textbf{34.31} & \textbf{0.945} & \textbf{0.934} & \textbf{68.63} \\
        \midrule
        \quad \texttt{K=11 frames}\\ \midrule
        Hassinof {\em et al.}~\cite{hasinoff10noiseoptimal}& 33.80 & 33.43 & 0.917 & 0.927 &\underline{68.86} \\
        \textbf{Ours} &\textbf{ 38.73 }&\textbf{ 38.56 }&\textbf{ 0.973} & \textbf{0.969 }&\textbf{70.05}\\ 
        \bottomrule
    \end{tabular}

\label{tab:raw_hdr_fusion_nomotion}
\end{table*}

\red{
In addition to Table~\ref{tab:raw_hdr_fusion}, we provide
in Table~\ref{tab:raw_hdr_fusion_nomotion} below a HDR fusion evaluation of the same methods, but on a variant of the 
synthetic test set where we have not simulated
motion between frames. In this new table, our method still
achieves the best HDR-VDP scores for brackets of both $K=3$ and
$K=11$ images, but with a margin of only 1 point over
our implementation of \cite{hasinoff10noiseoptimal}. We 
can conclude that both approaches achieve similar results. 
However, when we compare these margins that of 
3 points in Table~\ref{tab:raw_hdr_fusion}
between our approach over \cite{hasinoff10noiseoptimal}, it suggests
that our technique is more robust to alignment failures than the
pure bracketing technique of \citet{hasinoff10noiseoptimal}.
}

\section{Implementation Details}\label{sec:implem}
\red{
We include below details about our datasets and implementation for reproducibility purposes. See also Table \ref{tab:computationalefficiency} for the number of parameters used in different variants of our method.
}

\red{
\paragraph{Data Generation.}
Given a collection of sRGB images, we construct bursts of LDR low-resolution raw images and HDR/high-resolutions RGB targets.
For the generation of realistic raw data from sRGB images, we follow the approach described in \cite{bhat21deep}, using the author’s publicly available code on the training split of the Zurich raw to RGB dataset \cite{ignatov2020replacing}. The approach consists of applying the inverse RGB to raw pipeline introduced in \cite{brooks19unprocessing}.
For the training of our model, we generate bursts of 11 frames of size 256x256 with random motions. 
Displacements are randomly generated, applying random translations of $\pm6$ pixels and random rotations of $\pm1 ^{\circ}$. 
Frames are downscaled with bilinear interpolation in order to simulate LR frames containing aliasing. 
}

\red{
We then apply two different random gains in order to simulate frames with varying exposure. First, a random gain in the range [-5ev,5ev] is  applied to the ground-truth image, in order to simulate 32-bit ground-trugh images with a large dynamic.
We then apply a different gains in the range [-3ev, 3ev] for each image of the burst in order to simulate images with different exposure times. This results in synthetic bursts with different saturated areas and signal to noise ratios. 
Synthetic noise is added to the frames. The noise levels are sampled following the empirical model of Figure \ref{fig:isoplot}.
Finally, color values are discarded according to the Bayer pattern.
}

\red{
\paragraph{Validation Split}
In order to perform further comparison and conduct the ablation study, we build a  validation set by randomly extracting 266 images from the Zurich raw to RGB dataset \cite{ignatov2020replacing}.
}

\begin{figure}
    \centering
\includegraphics[width=0.5\textwidth]{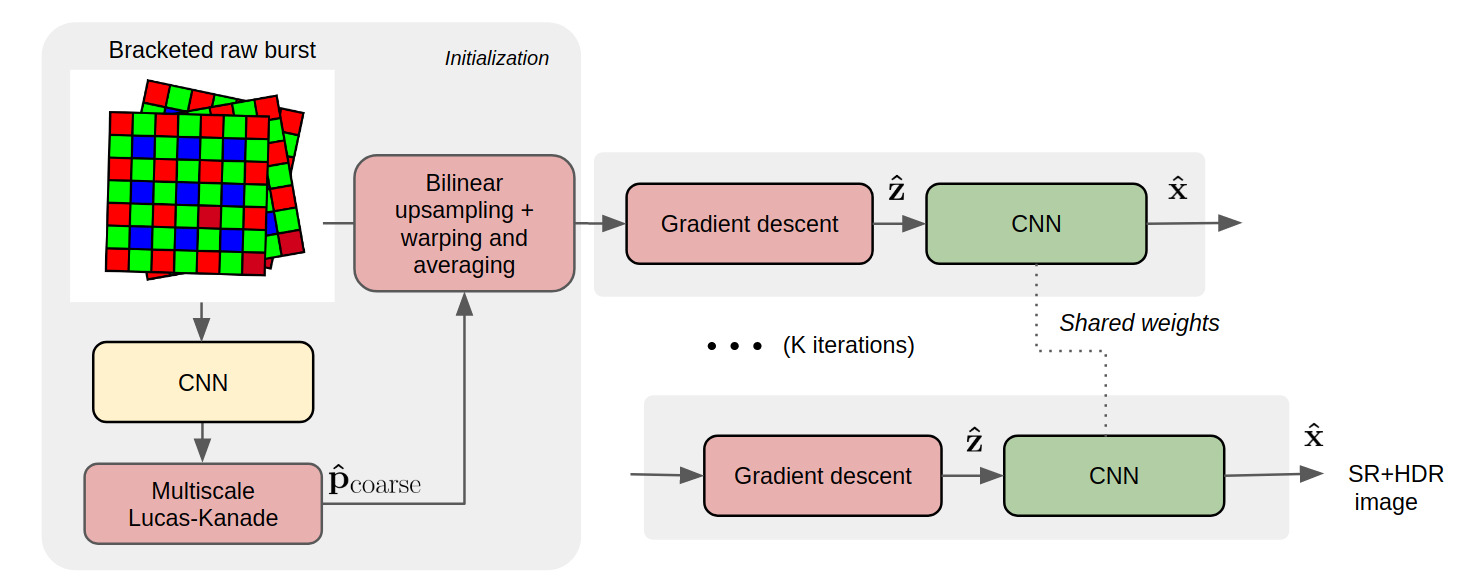}
    \caption{\red{A diagramatic view of our model.}}
    \label{fig:diaghdr}
\end{figure}

\red{
\paragraph{Model.}
We summarize in Figure \ref{fig:diaghdr} our proposed pipeline. In all our experiments, we unroll 3 iterations of the HQS algorithm.
}

\red{
\paragraph{Deep Prior.}
We give more details about the architecture of the deep prior used in our experiments.
For all our experiments, we use a smaller variant of the ResUNet architecture introduced in \cite{zhang2020deep} for single-image super resolution.
This architecture involves four scales, each of which has an identity skip connection between downscaling and upscaling operations. Downscaling operations are implemented using 2x2 strided convolution while upscaling are implemented with pixel-shuffling.  Each residual block is made of two 3x3 convolution layers and ReLU activation combined with an identity skip. For each scale we apply a cascade of 2 residual blocks. The network has respectively 32,64,128,128 channels for each convolution per scale.
}

\red{
\paragraph{Model variants.}
We also run experiments with an even smaller version of the network 
with 32 features per channel (dubbed small) and 16 features per
channel (dubbed tiny).
}

\red{
\paragraph{Training procedure}
We minimize Eq. \ref{eq:trainingloss} using Adam optimizer with learning rate set to $10^-5$ for 400k iterations. We decrease the learning rate by a factor 2 every 100k iterations. The weights of the CNNs are randomly initialized with the default setting of the PyTorch library. Our approach is implemented in Pytorch and takes approximately 2 days to train on a Nvidia Titan RTX GPU. 
}

\end{document}